%% file: main.tex
\pdfoutput=1
\documentclass[10pt,twocolumn,letterpaper]{article}
\input{release_style}
\input{preamble}

\submissiontrue
\openextratrue
\title{Token2Tile: Token-Wise Sparse Attention with Tile Execution\protect\\ through a Low-Rank Space for Video DiTs}
\input{authors}
\date{}
\begin{document}
\maketitle
\input{sections/abstract}
\input{sections/introduction}
\input{sections/preliminaries}
\input{sections/method}
\input{sections/experiments_brief}
\input{sections/related_work}
\input{sections/conclusion}
\FloatBarrier
\clearpage
\makeatletter
\immediate\write\@auxout{\string\newlabel{main:end}{{}{\number\numexpr\value{page}-1\relax}{}{}{}}}
\makeatother
\phantomsection\label{references:start}
{\small
\bibliographystyle{ieeenat_fullname}
\bibliography{references}
\label{references:end}
}

\clearpage
\onecolumn
\appendix
\setcounter{table}{0}
\setcounter{figure}{0}
\setcounter{equation}{0}
\setcounter{algorithm}{0}
\renewcommand{\thetable}{A\arabic{table}}
\renewcommand{\thefigure}{A\arabic{figure}}
\renewcommand{\theequation}{A\arabic{equation}}
\renewcommand{\thealgorithm}{A\arabic{algorithm}}
\renewcommand{\theHtable}{appendix.\arabic{table}}
\renewcommand{\theHfigure}{appendix.\arabic{figure}}
\renewcommand{\theHequation}{appendix.\arabic{equation}}
\providecommand{\theHalgorithm}{}
\renewcommand{\theHalgorithm}{appendix.\arabic{algorithm}}
\section*{Appendix}
\phantomsection\label{appendix:start}
The appendix gives derivations, the experimental protocol, and the full numbers behind the figures. Tables and figures prefixed A refer to this appendix.
\input{sections/appendix_arxiv}

\FloatBarrier
\label{appendix:end}
\end{document}

%% file: release_style.tex
\usepackage[letterpaper,textwidth=6.875in,textheight=9in,top=0.8in]{geometry}
\usepackage{xcolor}
\definecolor{releaseink}{HTML}{202020}
\definecolor{releasemuted}{HTML}{606469}
\definecolor{releaserule}{HTML}{404040}
\usepackage[numbers,sort&compress]{natbib}
\usepackage[font=small,labelfont=bf,labelsep=period]{caption}
\newcommand{\affmark}[1]{\textsuperscript{\normalfont\scriptsize #1}}
\makeatletter
\def\@affiliations{}
\newcommand{\affiliations}[1]{\gdef\@affiliations{#1}}
\newcommand{\authornote}[1]{\thanks{\textcolor{releasemuted}{#1}}}
\def\@maketitle{%
  \begingroup
  \parindent=0pt\centering
  \vspace*{4pt}%
  {\color{releaserule}\hrule height 0.7pt}\vspace{9pt}%
  {\color{releaseink}\fontsize{17}{20.5}\selectfont\bfseries\@title\par}%
  \vspace{8pt}{\color{releaserule}\hrule height 0.4pt}\vspace{11pt}%
  {\color{releaseink}\fontsize{10.5}{14}\selectfont\bfseries\@author\par}%
  \vspace{5pt}%
  {\fontsize{9.5}{12}\selectfont\@affiliations\par}%
  \vspace{14pt}%
  \endgroup
}
\renewcommand\section{\@startsection{section}{1}{\z@}%
  {-2.8ex plus -0.8ex minus -0.2ex}{1.2ex plus 0.2ex}%
  {\normalfont\fontsize{12}{14.4}\selectfont\bfseries}}
\renewcommand\subsection{\@startsection{subsection}{2}{\z@}%
  {-2.4ex plus -0.7ex minus -0.2ex}{0.9ex plus 0.2ex}%
  {\normalfont\fontsize{11}{13.2}\selectfont\bfseries}}
\renewcommand\subsubsection{\@startsection{subsubsection}{3}{\z@}%
  {-2.2ex plus -0.6ex minus -0.2ex}{0.7ex plus 0.2ex}%
  {\normalfont\normalsize\bfseries}}
\renewenvironment{abstract}{%
  \par\begingroup\normalsize
  {\centering\fontsize{11}{13.2}\selectfont\bfseries\abstractname\par}%
  \vspace{5pt}\noindent\ignorespaces
}{\par\endgroup\vspace{8pt}}
\def\ps@plain{%
  \let\@oddhead\@empty\let\@evenhead\@empty
  \def\@oddfoot{\hfil{\normalfont\fontsize{9}{11}\selectfont\color{releasemuted}\thepage}\hfil}%
  \let\@evenfoot\@oddfoot
}
\makeatother
\AtBeginDocument{%
  \crefname{section}{Sec.}{Secs.}%
  \Crefname{section}{Section}{Sections}%
  \crefname{table}{Tab.}{Tabs.}%
  \Crefname{table}{Table}{Tables}%
  \crefname{figure}{Fig.}{Figs.}%
  \Crefname{figure}{Figure}{Figures}%
  \crefname{equation}{Eq.}{Eqs.}%
  \Crefname{equation}{Equation}{Equations}%
  \crefname{appendix}{Sec.}{Secs.}%
  \Crefname{appendix}{Section}{Sections}%
}

%% file: preamble.tex
\usepackage{times}
\usepackage{graphicx}
\usepackage{amsmath,amssymb}
\newtheorem{theorem}{Theorem}
\usepackage{booktabs,multirow}
\usepackage{algorithm}
\usepackage{algpseudocode}
\usepackage{microtype}
\usepackage{xspace}
\usepackage{array}
\usepackage{tabularx}
\newcolumntype{Y}{>{\raggedleft\arraybackslash}X}
\usepackage{placeins}
\usepackage{enumitem}
\newif\ifsubmission
\newif\ifopenextra %
\usepackage{colortbl}
\definecolor{tablehead}{HTML}{FFFFFF}
\definecolor{modelrow}{HTML}{FFFFFF}
\definecolor{denserow}{HTML}{FFFFFF}
\definecolor{oursrow}{HTML}{D3E9AD}
\definecolor{cvprblue}{rgb}{0.21,0.49,0.74}
\usepackage[pagebackref,breaklinks,colorlinks,allcolors=cvprblue]{hyperref}
\usepackage[capitalize,noabbrev]{cleveref}
\newcommand{\method}{Token2Tile\xspace}
\newcommand{\R}{\mathbb{R}}

\newcommand{\up}{\ensuremath{\uparrow}}
\newcommand{\down}{\ensuremath{\downarrow}}

%% file: authors.tex
\author{%
  Xuecheng Wu\affmark{1,2}\quad
  Aichen Feng\affmark{2}\quad
  Jinyan Chen\affmark{2}\quad
  Qingjie Zhao\affmark{1}\quad
  Julien Lai\affmark{2}\authornote{Corresponding author.}
}
\affiliations{%
  \affmark{1}\,Beijing Institute of Technology\qquad
  \affmark{2}\,NVIDIA
}

%% file: sections/abstract.tex
\begin{abstract}
Attention is essential for modeling spatio-temporal dependencies in video diffusion transformers, but its quadratic cost makes it a major inference bottleneck. Its high sparsity offers an opportunity to accelerate generation by focusing computation on important interactions. However, accuracy and execution favor different granularities: each query attends to its own key tokens, whereas GPU kernels work on tiles. We present \method, a training-free sparse attention operator that combines per-query scoring and a per-query residual with tile execution. An online low-rank space, computed directly from queries and keys to preserve attention logits, makes clustering, scoring, and residual approximation efficient. \method unites query-specific selections to allocate exact work across query clusters, then packs the selected keys into GPU tiles. On MiniMax-H3 and Wan2.1-14B, \method delivers $1.48$--$1.54\times$ end-to-end speedup over dense FlashAttention-3 with VBench scores within 0.5 points of dense. At comparable speeds on H3, it improves PSNR over SOL and SubBlock by 5.5--6.8\,dB; on Wan, it exceeds SOL by 4.6--4.8\,dB while running faster.
\end{abstract}

%% file: sections/introduction.tex
\section{Introduction}
\input{figures/pareto}

Attention in video diffusion transformers grows quadratically with the number of latent tokens~\cite{dit,wan,minimaxh3}; on MiniMax-H3, it consumes roughly 64\% of each dense denoising step. Efficient kernels such as FlashAttention-3 (FA3)~\cite{flashattention3} raise GPU utilization but still compute every query--key interaction. As a drop-in replacement in a pretrained generator, sparse attention can shorten generation while keeping the output close to what the dense model would produce, without retraining.

Accuracy and execution favor different units of work. Each query---the token whose output is being computed---attends to its own set of keys, but GPU kernels are efficient on tiles: contiguous blocks of queries against contiguous blocks of keys. Many sparse attention methods therefore decide per tile or per group: block-sparse routing scores one pooled representative per tile~\cite{xattention,spargeattn,solattn}, while semantic permutation groups similar tokens and decides per cluster~\cite{svg2}. A pooled representative averages away the differences among the queries it stands for; on captured video attention it retains far less of each query's attention than the query itself would choose (\cref{fig:motivation}). Even accurate selection leaves attention mass outside the selected set, whose contribution must be approximated to preserve the output, motivating residual branches~\cite{sla,svgear,pisa}.

\method resolves this tension by retaining per-query scores and residuals while organizing exact attention into contiguous tiles. An online low-rank space, computed jointly from the current queries and keys to preserve attention logits, supports clustering, scoring, and residual approximation. Each query scores key clusters in this space and selects its own set. Within each query cluster, we unite the members' selections to allocate exact work according to their preferences, then pack the selected keys into contiguous tiles. Each query uses centroid scores to approximate the attention contributions outside the union, forming a \emph{per-query residual}. Clusters and selections are reused across denoising steps, so building them costs little.

\method delivers up to $1.54\times$ end-to-end speedup over dense FA3 on MiniMax-H3 and Wan2.1-14B while maintaining high video quality. At comparable generation speeds, it improves PSNR by up to 6.8\,dB over block-sparse attention (\cref{fig:pareto,tab:main}). Its attention kernels reach up to $7.9\times$ speedup over FA3 (\cref{fig:speed-accuracy-all}). Our contributions are:
\begin{itemize}[leftmargin=*,topsep=1pt,itemsep=0pt]
    \item \method, a training-free sparse attention operator that combines per-query scoring, a per-query residual, and adaptive allocation through unions with efficient tile execution;
    \item an online low-rank space, computed in closed form from queries and keys to preserve the attention logits, in which clustering, scoring, and the residual are all carried out accurately and efficiently within a single representation;
    \item a statistical model linking query-cluster disagreement to union size, together with controlled studies that separate the effects of scoring, decisions, and budget allocation.
\end{itemize}

%% file: figures/pareto.tex
\begin{figure*}[t]
\centering
\includegraphics[width=\textwidth]{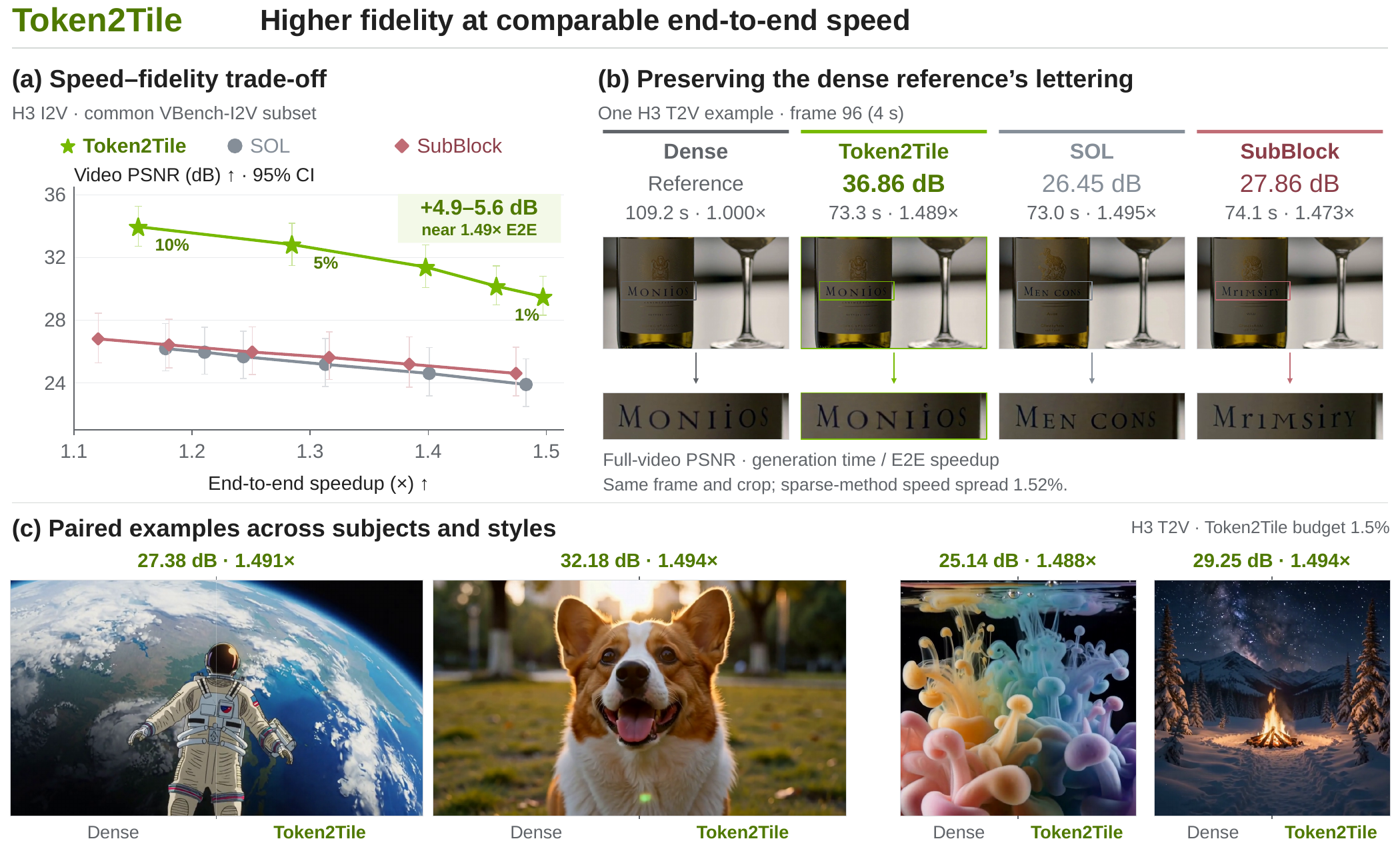}
\caption{\textbf{Higher dense-reference fidelity at comparable generation speed.}
(a) H3 I2V sweep on a common VBench-I2V subset with 95\% bootstrap PSNR intervals; green labels are nominal budgets. Near $1.49\times$ on this subset, \method improves PSNR by 4.9--5.6\,dB.
(b) Crops of lettering from the dense video and three sparse methods at $1.47$--$1.50\times$ (whole-video PSNR, request time).
(c) Dense-left/\method-right splits of four T2V examples at 1.5\%.}
\label{fig:pareto}
\label{fig:frames}
\end{figure*}

%% file: sections/preliminaries.tex
\section{Preliminaries}
\label{sec:preliminaries}
\input{figures/motivation_fig}
\paragraph{Dense attention and truncation.}
For one attention head, let $Q,K,V\in\R^{N\times d}$ and $\ell_{ij}=q_i k_j^\top/\sqrt d$. Dense attention computes
\begin{equation}
 O_i=\frac{\sum_j e^{\ell_{ij}}v_j}{\sum_j e^{\ell_{ij}}}.
 \label{eq:dense}
\end{equation}
Let $\mathcal E_i$ be the keys computed exactly for query $i$, $\alpha_i$ the dense attention mass outside $\mathcal E_i$, and $O_i^{E}$, $O_i^{R}$ the separately normalized outputs over $\mathcal E_i$ and its complement. Dropping the complement then changes the output by
\begin{equation}
 O_i-O_i^{E}=\alpha_i(O_i^{R}-O_i^{E}).
 \label{eq:truncation}
\end{equation}
The error depends on both the omitted mass and the omitted values; selecting the most prominent interactions alone need not preserve the output.

\paragraph{Tile-sparse execution and routing.}
FlashAttention~\cite{flashattention,flashattention3} evaluates \cref{eq:dense} tile by tile with an online softmax, so a key tile can be skipped only if every query in the tile skips it. Block-sparse methods therefore route with pooled proxies and keep blocks by top-$k$, top-$p$, or a row threshold~\cite{xattention,spargeattn,svg2,solattn}: all queries of a tile share one decision, and the keys of a block are kept or dropped together.

\paragraph{Why score per query, select key clusters, and keep a per-query residual.}\Cref{fig:motivation} tests each choice on attention captured from four video generators at a 1\% per-query key budget. (a) Per-query scoring: a query that selects key clusters with its own scores captures 59--68\% of its dense attention mass on H3, Wan, and HunyuanVideo and 32\% on the diffuse LTX-2.3; sharing one selection across a contiguous tile of 64 queries, or selecting contiguous key blocks from pooled scores, captures 10--62\%. (b) Clustering keys: the same query selecting contiguous 64-key blocks instead of key clusters captures 1.5--3.2$\times$ less of the mass. (c) Clustering queries: uniting the same per-query selections over query clusters reduces the executed pairs 9--16$\times$, over contiguous query tiles of the same size only 5--7$\times$. (d) The residual: even per-query selection leaves 32--68\% of the mass in the tail, and dropping it instead of compensating it with the per-query residual raises the output error 1.4--27$\times$.

%% file: figures/motivation_fig.tex
\begin{figure*}[t]
\centering
\includegraphics[width=\textwidth]{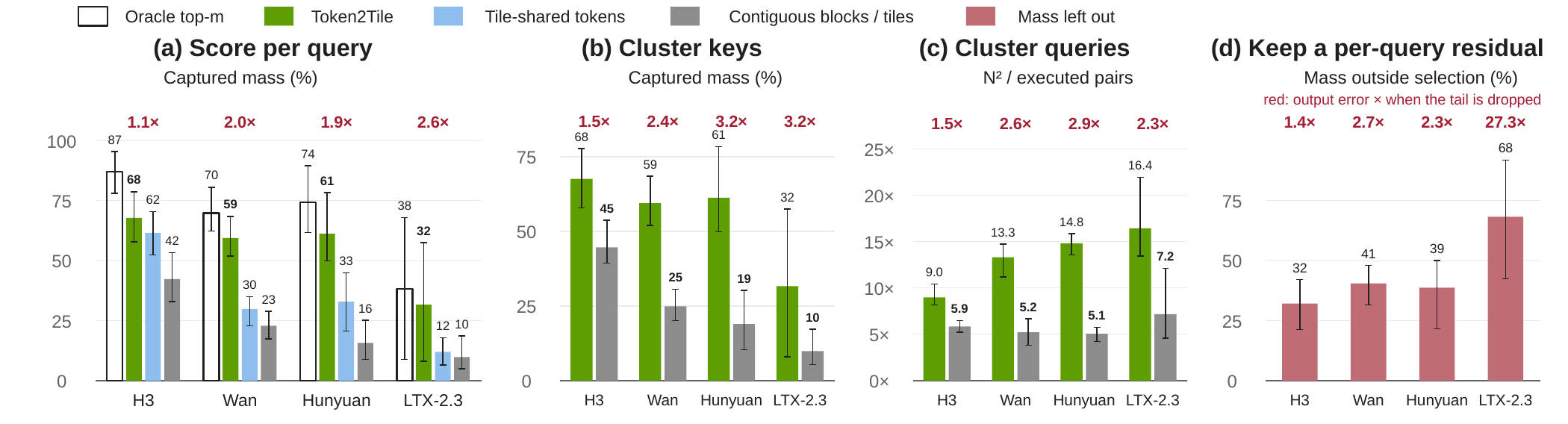}
\caption{\textbf{Why score per query, select key clusters, and keep a per-query residual.} Attention captured from four generators at a 1\% per-query key budget; whiskers span the inputs, red labels give the ratio between the green bar and the bar it is compared with. (a) Dense attention mass captured by per-query selection of key clusters and other selection methods. (b) Per-query selection of key clusters versus contiguous 64-key blocks. (c) Reduction in executed pairs when per-query selections are united over query clusters versus contiguous query tiles of the same size. (d) Mass left in the tail, and the output-error increase when the tail is dropped instead of compensated by the per-query residual.}
\label{fig:motivation}
\end{figure*}

%% file: sections/method.tex
\section{Token2Tile}
\label{sec:method}
\input{figures/method} 
\method partitions keys into clusters and uses each cluster in one of two ways for each query: as its original key tokens in the exact branch, or as a centroid in the per-query residual (\cref{fig:method}). Both branches keep query-specific weights, so the operator approximates the whole attention output while evaluating only part of the query--key pairs.

\subsection{A logit-preserving low-rank space}
\label{sec:geometry}
We use one low-rank space for clustering, scoring, and the per-query residual. Since all three depend on query--key interactions, the space should preserve $QK^\top$ rather than reconstruct $Q$ and $K$ separately. We seek matched projections $P_q,P_k\in\R^{d\times r}$, with $\widetilde Q=QP_q$ and $\widetilde K=KP_k$, that minimize the logit error
\begin{equation}
 \mathcal L(P_q,P_k)=\frac{1}{N^2}
 \big\|QK^\top-\widetilde Q\widetilde K^\top\big\|_F^2.
 \label{eq:logit-mse}
\end{equation}
This objective has a closed-form solution from the $d\times d$ second moments $M_q=Q^\top Q/N$ and $M_k=K^\top K/N$.
\begin{theorem}
\raggedright
\label{thm:projection}
For positive-definite $M_q,M_k$, let $M_q^{1/2}M_k^{1/2}=U\Sigma Z^\top$. A pair of rank-$r$ projections minimizing \cref{eq:logit-mse} is
\begin{equation}
 P_q=M_q^{-1/2}U_r\Sigma_r^{1/2},\qquad
 P_k=M_k^{-1/2}Z_r\Sigma_r^{1/2},
 \label{eq:projection}
\end{equation}
with minimum error $\sum_{j>r}\sigma_j^2$.
\end{theorem}
The solution follows from Eckart--Young~\cite{eckartyoung}; the derivation and clustering geometry are given in \cref{sec:full-derivations}. For the original 128-dimensional queries and keys ($d=128$), we use a 32-dimensional low-rank space ($r=32$). Its projections require one $128\times128$ decomposition per head, computed online from the request's own queries and keys without training.

We then apply $k$-means~\cite{lloyd} in this shared space: $C_K=2048$ key clusters serve as the units of selection and residual approximation, while $C_Q\approx N/128$ query clusters group similar-scoring queries for tile execution. Each query cluster holds about 128 queries, matching the kernel's tile height.

\subsection{From per-query selection to a tile-sparse mask}
\label{sec:selection}
Given a token budget $b$, query $i$ scores each key cluster $c$ against its centroid $\mu_c$ in the low-rank space, $g_{ic}=\widetilde q_i\mu_c^\top/\sqrt d$, orders the clusters by $g_{ic}$, and keeps the shortest prefix $\mathcal S_i$ whose sizes sum to at least $bN$, the last cluster whole. Queries of one cluster $a$ then execute the union of their selections,
\begin{equation}
 \mathcal U_a=\bigcup_{i:a_i=a}\mathcal S_i,\qquad
 \mathcal E_i=\{j:z_j\in\mathcal U_{a_i}\},
 \label{eq:union}
\end{equation}
plus the clusters of pinned context keys (text or audio tokens), which stay exact. Each query keeps its own exact logits and residual weights. Members of a query cluster are close in the space in which they score, so they rank the key clusters alike and the union stays compact (\cref{fig:motivation}c, \cref{tab:overlap}). The union raises density above the budget, but we treat its varying size as adaptive allocation: clusters whose members select different keys receive more exact work (\cref{sec:ablation}). A statistical model connects the members' score disagreement to expected union size; its estimates rank observed union sizes with correlation 0.95--0.99 (\cref{sec:union-model}). We measure the executed work as the exact-pair density
\begin{equation}
 \rho=\frac{\sum_a |\mathcal Q_a|\sum_{c\in\mathcal U_a}n_c}{N^2},
 \label{eq:density}
\end{equation}
with $\mathcal Q_a$ the queries of cluster $a$. Whole clusters and the union both separate $\rho$ from $b$ (the 2.5\% H3 T2V schedule executes $\rho\approx16.5\%$ averaged over the denoising trajectory), so we compare methods at equal $\rho$.

\subsection{Per-query residual}
Each query's output combines exact attention over $\mathcal E_i$ with a residual over the key clusters outside its union, each represented by its centroid score $g_{ic}$, size $n_c$, and value sum $s_c=\sum_{z_j=c}v_j$:
\begin{equation}
 \widehat O_i=\frac{\sum_{j\in\mathcal E_i}e^{\ell_{ij}}v_j+\sum_{c\notin\mathcal U_{a_i}}e^{g_{ic}}s_c}
 {\sum_{j\in\mathcal E_i}e^{\ell_{ij}}+\sum_{c\notin\mathcal U_{a_i}}e^{g_{ic}}n_c}.
 \label{eq:output}
\end{equation}
The residual runs over the complement of the executed union, not of the query's own selection, so a key added by another member is never counted twice. Equivalently, with unnormalized weights $W_{ij}=e^{\ell_{ij}}$, their cluster approximation $\widetilde W_{ij}=e^{g_{i,z_j}}$, and the executed mask $M$,
\begin{equation}
 \widehat W=\widetilde W+M\odot(W-\widetilde W):
 \label{eq:correction}
\end{equation}
unselected keys keep an approximate contribution instead of being pruned; the $N\times N$ matrices are never materialized. Proxy logit error combines within-cluster variation and low-rank approximation. When its maximum over omitted keys is below $\log 2$, the residual has a tighter worst-case output-error bound than truncation (\cref{sec:operator-details}).

\subsection{Execution and temporal reuse}
\label{sec:execution}
Keys and values are permuted into contiguous cluster segments, and a sparse kernel traverses the selected segments of each query cluster, packing compute tiles across cluster boundaries instead of rounding every cluster pair up to a fixed block. Across denoising calls, labels, centroids, and unions are refreshed every $R=4$ calls, and the projection is fixed within a request; reuse calls apply the cached structure to the current $Q,K,V$, so we cache structure, not attention outputs. Requests keep separate states; under classifier-free guidance, the unconditional branch reuses the structure built by the conditional branch.
\ifopenextra
The first sparse call of a layer runs four $k$-means iterations; each later refresh warm-starts from the centroids of the previous refresh and updates the key clusters once and the query clusters twice. A reuse call keeps the decisions of the last refresh, namely which key clusters each query cluster executes exactly and which it leaves to the residual, and recomputes their values: exact attention and the residual weights use the current queries and keys, and the value sums use the current values. The centroid scores $g_{ic}$ that rank key clusters for selection are also the residual logits in \cref{eq:output}, so one scoring serves both.
\fi
A call costs
\begin{equation}
 O\!\left(\rho N^2d+NC_Kr+NC_Kd+\frac{C_{\mathrm{refresh}}}{R}\right),
\end{equation}
where the $NC_K$ terms are the residual scores and value aggregation and $C_{\mathrm{refresh}}$ is clustering and selection. On long video sequences the first term dominates (\cref{sec:kernel}).

\subsection{Choosing a denoising-time budget}
\label{sec:budget-calibration}
After a dense prefix, we allocate attention budgets across four denoising windows using offline error-injection calibration. A calibration input provides the window-wise error curves $e_w(b)$: the denoiser is replayed at one probe step per window for each budget on a grid, measuring SSE over dense-output energy. With window lengths $h_w$, dynamic programming solves
\begin{equation}
 \min_{b_w\in\mathcal B}\sum_w h_we_w(b_w)
 \quad\text{s.t.}\quad
 \sum_w h_wb_w\le B\sum_w h_w
 \label{eq:dp}
\end{equation}
for each target $B$, treating window errors as additive. The schedule is normalized to $B$ and reused across generation prompts without updating model weights. Multiple calibration inputs are evaluated in \cref{sec:app-budget}.

%% file: figures/method.tex
\begin{figure*}[t]
\centering
\includegraphics[width=\textwidth]{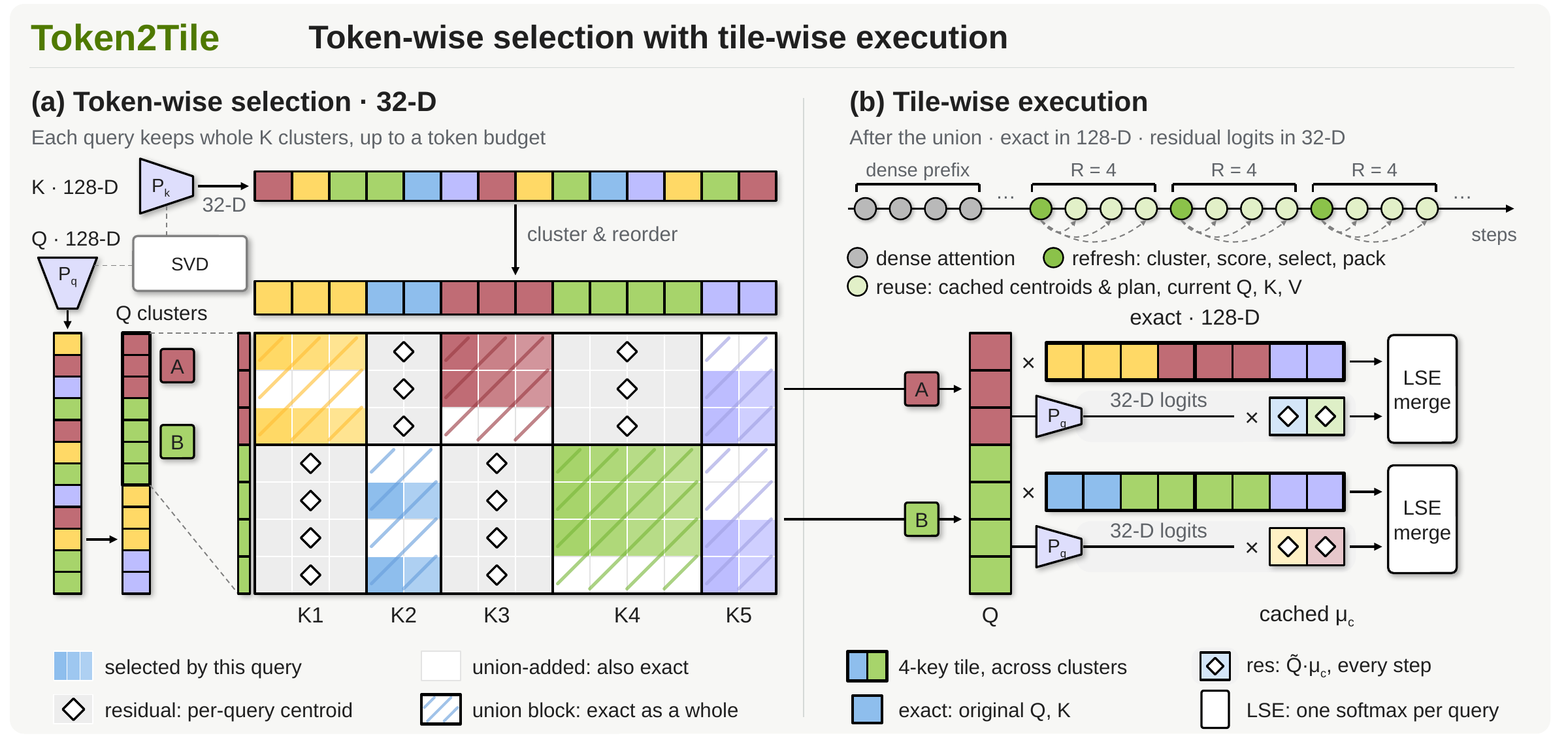}
\caption{\textbf{Token-wise selection with tile-wise execution.}
(a) An SVD of the second-moment square-root product gives matched projections $P_q,P_k$ from 128 to 32 dimensions, in which queries and keys are clustered per token. Each query keeps the shortest prefix of whole key clusters that meets its token budget (colored); a query cluster executes the union of its members' selections (outlined), and each remaining cluster contributes one centroid term per query (grey, $\diamond$).
(b) The original $Q$ multiplies the union keys, packed into tiles across cluster boundaries; $QP_q$ multiplies the centroids of the unselected clusters, and an LSE merge yields one softmax per query. Clusters, selections and plans are refreshed every $R=4$ calls; reuse calls use the current $Q$, $K$, $V$. No $N\times N$ tensor is materialized.}
\label{fig:method}
\end{figure*}

%% file: sections/experiments_brief.tex
\section{Experiments}
\label{sec:experiments}
\raggedbottom
\setlength{\textfloatsep}{12pt plus 1pt minus 2pt}
\subsection{Setup}
We evaluate MiniMax-H3~\cite{minimaxh3} (243 frames, $1344\times768$) and Wan2.1-14B~\cite{wan} (81 frames, $1280\times720$) on T2V with VBench prompts and on I2V with VBench-I2V images~\cite{vbench,vbenchpp}, with multi-seed VBench subsets for sweeps and ablations, on eight H200 GPUs with Ulysses-8~\cite{ulysses} in SGLang-Diffusion~\cite{sglangdiffusion}. PSNR, SSIM~\cite{ssim}, and LPIPS~\cite{lpips} measure fidelity to the paired dense video, and speedup is the paired request-time ratio against dense FA3. Generation comparisons retain each implementation's default dense-prefix and dense-layer policies: SOL~\cite{solattn} and SubBlock~\cite{subblock} in SGLang on both models, and official SVG2/SVG-EAR~\cite{svg2,svgear} on Wan. SubBlock is not tuned separately for Wan.

A budget $b$ lets each query keep about $bN$ of the $N$ key tokens; after unions, 1--5\% budgets execute $\rho=$ 12--32\% (H3) and 8--26\% (Wan) of the query--key pairs (\cref{tab:union}). Operator and kernel studies use \emph{captured attention}: the $Q,K,V$ inputs of one rank (65K--144K tokens), recorded during generation at three denoising steps from H3, Wan, HunyuanVideo~\cite{hunyuanvideo}, and LTX-2.3~\cite{ltxvideo}. Attention SNR is $-20\log_{10}$ of relative $\ell_2$ error to FP32 dense attention (higher is better). Operator comparisons use real outputs for \method, SVG-EAR, and PISA, and FP32 implementations for other rules; kernel comparisons always use real outputs. Kernel times are CUPTI duration sums on one H200 SXM. Details are in \cref{sec:protocol}.

\subsection{Generation quality and speed}
\input{tables/main_results}
\paragraph{Matched speed.} At similar speed, \method is 5.5--6.8\,dB closer to the dense video than block-sparse attention (SOL, SubBlock) on H3 (27.83\,dB at $1.493\times$ on T2V, 30.37\,dB at $1.476\times$ on I2V) and 4.6--4.8\,dB closer than SOL on Wan while also running faster ($1.52$--$1.54\times$ against $1.41$--$1.42\times$; \cref{tab:main}); SSIM and LPIPS agree, and the VBench dimensions stay within 0.5 points of dense.

\paragraph{The whole trade-off.} \Cref{fig:pareto}a sweeps every method on H3 I2V. From a 10\% to a 1\% budget, \method moves from $1.15\times$ at 34.0\,dB to $1.50\times$ at 29.5\,dB; SOL ($\tau=0.1$--2) and SubBlock (sparsity 0.5--0.9) span $1.12$--$1.80\times$ at 20.9--26.8\,dB. At every speed the curves share, \method is 4.9--7.8\,dB higher: 29.5 against 23.9--24.6\,dB near $1.48\times$, and 34.0 against 26.2--26.4\,dB near $1.16\times$.

\paragraph{Blind study.} Raters chose which of two accelerated videos is closer to the dense one; counting ties as half, \method (at the budgets of \cref{tab:main} on H3 and 5\% on Wan) scores 81\% against block-sparse attention (95\% interval 77--85\%) and 56\% against SVG2/SVG-EAR (52--61\%); in the rated Wan videos \method runs at $1.38$--$1.40\times$ and SVG2/SVG-EAR at $0.96$--$1.07\times$ the speed of dense FA3 (\cref{sec:human-study}, \cref{tab:blind}).

\input{tables/distilled_best}
\paragraph{Fewer GPUs and fewer steps.} On one GPU, \method at 5\% reaches $1.48\times$ on Wan T2V, on par with block-sparse attention. On the eight-step distilled H3 variants, LightX2V and HyperFlow~\cite{lightx2v,hyperflow}, \method adds $1.13$--$1.17\times$ denoiser speedup while retaining 30.5--35.6\,dB PSNR (\cref{tab:distilled-best}). The reuse interval $R$ is a hyperparameter: $R=4$ by default, and $R=1$ for these eight-step models. End-to-end speedup is $1.03$--$1.11\times$: with fewer denoising calls, VAE decoding and other work outside the denoiser account for a larger share of request time.

\subsection{Accuracy at equal density}
\label{sec:methods-equal}
\input{tables/equal_density}
To compare selection rules rather than budgets, \cref{tab:equal-density} matches each rule to \method{}\textquotesingle s executed pair density, head by head. \method is the most accurate rule on all four generators and at every matched budget (\cref{tab:operator-compare}), 3.8--6.8\,dB above the next best, SVG-EAR, at 3.5\% and 13--30\,dB above the block rules; giving the block rules \method{}\textquotesingle s granularity changes them by at most 1.4\,dB (\cref{tab:same-granularity}). At that budget it reproduces dense attention to within 1.9--2.9\% relative error while computing 15--26\% of the pairs. All budgets and controls are in \cref{sec:app-equal}.

\subsection{Kernel performance}
\label{sec:kernel}
\input{figures/kernel}
\input{figures/speed_accuracy_fig}
\paragraph{The online structure costs about a tenth of a call.} \Cref{fig:kernel} breaks down one H3 attention call (step 32, layer 24) against 27.20\,ms for dense FA3. Its $R=4$ mean is $2.71\times$ faster than FA3. Exact attention takes 80\% of the time and the residual 9.5\%; projection, clustering, scoring, selection, and planning together take about a tenth. Across the four generators, \method reaches 2.5--3.8$\times$ at a 3.5\% budget and 4.40--7.92$\times$ at 1\% (\cref{sec:app-kernel}).

\paragraph{Speed and accuracy together.} The budget moves the kernel along one speed--accuracy curve (\cref{fig:speed-accuracy}): from 10\% to 1\%, the H3 kernel goes from $1.5\times$ to $4.4\times$ faster than FA3 at 46 to 27\,dB SNR, and within this range every other real kernel we ran falls below the curve; the other three generators behave the same (\cref{sec:app-kernel}).

\input{figures/tile_execution_fig}
\paragraph{Tile execution.} On the same cluster masks, our exact kernel is $1.61$--$1.64\times$ faster than FlashInfer variable-block attention (\cref{fig:tile-execution}). Fixed $64\times64$ or $128\times128$ covers execute $1.66$--$3.0\times$ as many pairs; packing across cluster boundaries avoids this extra work. Variable-length tiles cost no speed: at equal density and 3.5--10\% budgets, the full call is $1.03$--$1.14\times$ as fast as a routed fixed-block kernel on H3, Wan, and HunyuanVideo (\cref{tab:same-density-kernel}). The processor keeps 1.7--3.4\,GiB resident per rank, mainly in shared workspace (\cref{tab:attn-memory}).

\subsection{Ablations}
\label{sec:ablation}
\input{tables/ablation_main}
\Cref{tab:ablation-main} isolates each component in generated videos and captured attention; further variants are in \cref{sec:ablation-details}.

\paragraph{Per-query scores and residuals.} The per-query residual contributes the largest fidelity gain: removing it lowers video PSNR by 1.80/3.25\,dB on T2V/I2V and attention SNR by up to 26.5\,dB on LTX-2.3; sharing it within a query cluster recovers only part of this gain. Shared selection from the query-cluster mean ties on T2V but loses 0.99\,dB on I2V and 0.9--2.3\,dB on captured attention. Per-query scoring and selection cost 5.5\% of the deployed call. Summing members' estimated attention mass for one decision per key cluster retains similar fidelity when each query cluster keeps its union-derived key budget: within about 0.5\,dB on captured attention and $+0.16$/$-0.08$\,dB video PSNR (s.e.\ 0.14; \cref{tab:tile-video}). Without the residual, per-query selection loses 0.4--0.5\,dB to shared selection, showing the joint role of query-specific scores and the residual.

\paragraph{Union execution and adaptive allocation.} Clustering raises selection overlap (Jaccard 0.30--0.45 versus 0.16--0.20 within contiguous tiles), so the union executes 1.3--3.1$\times$ fewer pairs than over contiguous tiles of the same size (\cref{tab:overlap,tab:union}). Unions expand per-query prefixes by 3.0--14$\times$; at 3.5\%, the added exact pairs raise SNR by 5.6--11.0\,dB. With the same summed-mass decision rule, union-derived budgets improve SNR by 0.35--0.95\,dB over uniform budgets at matched total density (\cref{tab:ablation-main}c). Uniform budgets for union selection cost 0.3--0.9\,dB at 3.5\% (0.14--0.9\,dB over both budgets). Global greedy reallocation improves only H3 at 3.5\% (0.07\,dB in the reference initialization; \cref{tab:e1-aggregation,sec:noise}).

\input{tables/union_growth_main}
\paragraph{Query-cluster size balances density and execution.} Roughly 128 queries per cluster yield the fastest FlashInfer variable-block execution on three of four captures, and come within 9\% of the best on Wan (\cref{tab:group-size}). Halving cluster size saves 21--24\% of pairs at 1.1--2.1\,dB lower SNR; doubling adds 26--31\% more pairs and 6--11\% more time. The statistical model closely predicts the change in union inflation when cluster size doubles (\cref{tab:union-growth}).

\paragraph{A 32-dimensional low-rank space suffices.}\label{sec:dimred} On H3 videos, the 32-dimensional low-rank space of \cref{sec:geometry} improves on clustering and scoring in the original 128-dimensional space by 0.04/0.66\,dB on T2V/I2V. Its clustering assignments and per-query residual computation together run 1.65--1.88$\times$ faster (\cref{tab:dimred-fused}). It also exceeds a shared 32-dimensional principal component analysis (PCA) projection for $Q$ and $K$ by 0.09/0.57\,dB, with 7--12\% lower within-cluster logit variance than the original space. On captured attention, the original space has 0.1--1.5\,dB higher SNR at 1\%, and using it to score residual centroids gains 0.2--0.7\,dB at 3.5\% (\cref{tab:operator-compare,tab:dimred}). Substituting the 128-dimensional stage timings gives an estimated 6--8\% full-call slowdown at 3.5\% and 11--13\% at 1\% (\cref{sec:app-lowrank}); the H3 video control favors the cheaper shared low-rank space. Reducing it to 16 dimensions costs up to 2.6\,dB at 3.5\% (3.7\,dB at 1\%); increasing it beyond 32 changes results within $k$-means variability (\cref{sec:app-lowrank,sec:noise}).

\paragraph{Spend time on the budget.} At the kernel time of refreshing every other call ($R=2$), keeping $R=4$ and raising the budget gains 1.07/0.46\,dB over the default, versus 0.34/0.16\,dB from refreshing. Rebuilding every call ($R=1$) gains 0.42/0.27\,dB for a 1.24--1.43$\times$ slower call; $R=8$ saves 4--8\% but loses 0.55/1.03\,dB. Recomputing centroids changes video PSNR by at most 0.03\,dB; reselecting recovers 0.17/0.19\,dB at greater cost than $R=2$ (\cref{tab:reuse}).

\input{figures/budget_alloc_main_fig}
\paragraph{Budget allocation by error injection.}\label{sec:budget-alloc} The calibrated schedule from \cref{sec:budget-calibration} gains 0.55--0.80\,dB on H3 T2V and Wan I2V over a uniform schedule at the same nominal budget, without executing more pairs (16.5 versus 16.8\% on H3 T2V). It assigns more to earlier windows: sparsifying the first window after the dense prefix distorts the output 7--8$\times$ more than sparsifying the last. At the tested 2.5\% budget, non-uniform schedules from other calibration inputs retain similar fidelity. Uniform outcomes lose about 0.8\,dB on H3 T2V and Wan I2V, but only 0.18\,dB on the less time-sensitive H3 I2V (\cref{fig:budget-alloc-main,sec:app-budget}).

%% file: tables/main_results.tex
\begin{table*}[t]
\centering
\caption{\textbf{Generation fidelity and speed.} Individual VBench dimensions (\%) use the same videos as the fidelity metrics. Shading marks \method; bold marks the best fidelity.}
\label{tab:main}
\footnotesize
\setlength{\tabcolsep}{3.3pt}
\renewcommand{\arraystretch}{1.08}
\begin{tabularx}{\textwidth}{@{}clYYYYYYYYY@{}}
\toprule
Task & Method & \multicolumn{3}{c}{Similarity to dense} & \multicolumn{4}{c}{VBench dimensions (\%)\up} & \multicolumn{2}{c}{Efficiency (8 GPUs)}\\
\cmidrule(lr){3-5}\cmidrule(lr){6-9}\cmidrule(lr){10-11}
& & PSNR\up & SSIM\up & LPIPS\down & Subject & Backgr. & Motion & Aesthetic & Time (s)\down & Speedup\up\\
\midrule
\multicolumn{11}{c}{\textit{MiniMax-H3}}\\
\midrule
\textbf{T2V} & Dense (FA3) & -- & -- & -- & 88.24 & 93.19 & 99.25 & 57.16 & 109.0 & $1.000\times$\\
 & SOL & 21.00 & 0.722 & 0.206 & 88.25 & 93.37 & 99.21 & 57.32 & 73.5 & 1.483$\times$\\
 & SubBlock 0.75 & 21.55 & 0.741 & 0.190 & 88.26 & 93.31 & 99.26 & 57.20 & 74.1 & 1.471$\times$\\
\rowcolor{oursrow}
 & \method 1.5\% & 27.83 & 0.875 & 0.085 & 88.10 & 93.18 & 99.20 & 56.96 & 73.0 & 1.493$\times$\\
\rowcolor{oursrow}
 & \method 5\% & \textbf{30.94} & \textbf{0.918} & \textbf{0.055} & 88.20 & 93.08 & 99.23 & 57.19 & 84.1 & 1.296$\times$\\
\midrule
\textbf{I2V} & Dense (FA3) & -- & -- & -- & 91.23 & 93.84 & 99.27 & 60.93 & 114.5 & $1.000\times$\\
 & SOL & 24.03 & 0.803 & 0.120 & 91.08 & 93.78 & 99.25 & 60.50 & 77.0 & 1.486$\times$\\
 & SubBlock 0.75 & 24.88 & 0.825 & 0.106 & 91.18 & 93.90 & 99.28 & 60.97 & 77.5 & 1.477$\times$\\
\rowcolor{oursrow}
 & \method 1.25\% & 30.37 & 0.914 & 0.052 & 91.14 & 93.84 & 99.24 & 60.44 & 77.5 & 1.476$\times$\\
\rowcolor{oursrow}
 & \method 5\% & \textbf{32.73} & \textbf{0.934} & \textbf{0.039} & 91.18 & 93.79 & 99.25 & 60.53 & 91.6 & 1.249$\times$\\
\midrule
\multicolumn{11}{c}{\textit{Wan2.1-14B}}\\
\midrule
\textbf{T2V} & Dense (FA3) & -- & -- & -- & 97.52 & 96.92 & 99.07 & 59.04 & 136.0 & $1.000\times$\\
 & SOL & 22.70 & 0.800 & 0.137 & 97.46 & 96.86 & 99.10 & 58.99 & 96.0 & 1.417$\times$\\
 & SubBlock 0.75 & 18.04 & 0.641 & 0.328 & 89.71 & 92.01 & 98.95 & 56.89 & 94.9 & 1.433$\times$\\
 & SVG2 & 24.69 & 0.849 & 0.099 & 97.39 & 96.71 & 99.04 & 59.13 & 127.5 & 1.067$\times$\\
 & SVG-EAR & 26.78 & 0.886 & 0.071 & 97.46 & 96.76 & 99.06 & 59.08 & 142.2 & 0.957$\times$\\
\rowcolor{oursrow}
 & \method 2.5\% & 27.48 & 0.899 & 0.064 & 97.52 & 96.89 & 99.07 & 58.94 & 88.4 & 1.538$\times$\\
\rowcolor{oursrow}
 & \method 5\% & \textbf{28.57} & \textbf{0.913} & \textbf{0.053} & 97.51 & 96.89 & 99.07 & 58.96 & 96.9 & 1.404$\times$\\
\midrule
\textbf{I2V} & Dense (FA3) & -- & -- & -- & 94.41 & 95.59 & 98.53 & 59.98 & 137.9 & $1.000\times$\\
 & SOL & 25.63 & 0.836 & 0.100 & 94.29 & 95.37 & 98.56 & 59.80 & 97.7 & 1.411$\times$\\
 & SubBlock 0.75 & 24.11 & 0.810 & 0.122 & 93.77 & 95.14 & 98.41 & 59.28 & 96.7 & 1.426$\times$\\
 & SVG2 & 28.64 & 0.884 & 0.069 & 94.28 & 95.41 & 98.50 & 59.55 & 129.9 & 1.061$\times$\\
 & SVG-EAR & 30.76 & 0.912 & 0.051 & 94.34 & 95.44 & 98.52 & 59.79 & 142.0 & 0.971$\times$\\
\rowcolor{oursrow}
 & \method 2.5\% & 30.26 & 0.910 & 0.054 & 94.39 & 95.48 & 98.54 & 59.84 & 90.9 & 1.516$\times$\\
\rowcolor{oursrow}
 & \method 5\% & \textbf{30.94} & \textbf{0.918} & \textbf{0.049} & 94.40 & 95.49 & 98.54 & 59.93 & 99.8 & 1.382$\times$\\
\bottomrule
\end{tabularx}
\end{table*}

%% file: tables/distilled_best.tex
\begin{table}[t]
\centering
\caption{\textbf{Eight-step distilled H3 models.} Four GPUs; 5\% budget, three dense calls, $R=1$. PSNR is relative to each model\textquotesingle s dense video; DiT/E2E denote denoiser/full-request speedups on a VBench subset.}
\label{tab:distilled-best}
\footnotesize
\setlength{\tabcolsep}{3.4pt}
\begin{tabular}{@{}lrrrrrr@{}}
\toprule
 & \multicolumn{3}{c}{T2V} & \multicolumn{3}{c}{I2V}\\
\cmidrule(lr){2-4}\cmidrule(lr){5-7}
Model & PSNR & DiT & E2E & PSNR & DiT & E2E\\
\midrule
LightX2V~\cite{lightx2v} & 35.57 & 1.17$\times$ & 1.11$\times$ & 35.02 & 1.14$\times$ & 1.09$\times$\\
HyperFlow~\cite{hyperflow} & 30.53 & 1.16$\times$ & 1.06$\times$ & 30.71 & 1.13$\times$ & 1.03$\times$\\
\bottomrule
\end{tabular}
\end{table}

%% file: tables/equal_density.tex
\begin{table}[t]
\centering
\caption{\textbf{Accuracy at equal executed density.} Attention SNR (dB); $b$: per-query budget, $\rho$: executed density (both \%). $^\dagger$SVG-EAR exceeds the target density because its minimum is about 10\%. On H3, pinned context is exact for every rule except the official SVG-EAR kernel. Full results: \cref{tab:operator-compare}.}
\label{tab:equal-density}
\scriptsize
\setlength{\tabcolsep}{2.2pt}
\begin{tabular}{@{}lrrrrrrrrr@{}}
\toprule
 & $b$ & $\rho$ & \method & EAR & SVG2 & XAttn & SubBl. & PISA & SOL\\
\midrule
H3 & 1.0 & 11.7 & \textbf{27.2} & 23.2 & 19.8 & 15.7 & 16.7 & 17.1 & 16.5\\
 & 3.5 & 25.7 & \textbf{34.6} & 27.7 & 27.1 & 21.3 & 21.6 & 21.1 & 20.6\\
\addlinespace[1pt]
Wan & 1.0 & 7.6 & \textbf{25.8} & 25.4$^\dagger$ & 15.1 & 10.7 & 11.2 & 13.6 & 12.8\\
 & 3.5 & 20.0 & \textbf{32.9} & 27.7 & 20.8 & 15.4 & 15.8 & 18.2 & 17.9\\
\addlinespace[1pt]
Hun. & 1.0 & 6.8 & \textbf{24.4} & 23.4$^\dagger$ & 15.4 & 9.3 & 9.9 & 10.4 & 10.4\\
 & 3.5 & 17.2 & \textbf{30.7} & 26.5 & 21.2 & 12.6 & 13.4 & 13.8 & 13.8\\
\addlinespace[1pt]
LTX & 1.0 & 6.4 & \textbf{29.6} & 29.6$^\dagger$ & 4.1 & 1.4 & 1.9 & 15.6 & 10.2\\
 & 3.5 & 14.8 & \textbf{34.5} & 30.7 & 7.4 & 4.6 & 4.9 & 18.6 & 14.8\\
\bottomrule
\end{tabular}
\end{table}

%% file: figures/kernel.tex
\begin{figure}[t]
\centering
\includegraphics[width=\columnwidth]{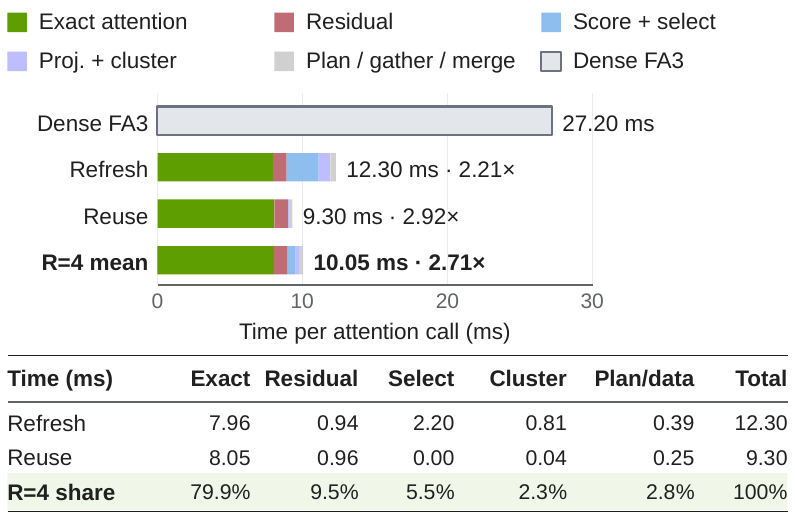}
\caption{\textbf{Time breakdown of an H3 attention call} (3.5\% budget, 23.4\% executed density). Bars show dense FA3, refresh, reuse, and their $R=4$ mean; the table lists stage times and mean shares.}
\label{fig:kernel}
\end{figure}

%% file: figures/speed_accuracy_fig.tex
\begin{figure}[t]
\centering
\includegraphics[width=\columnwidth]{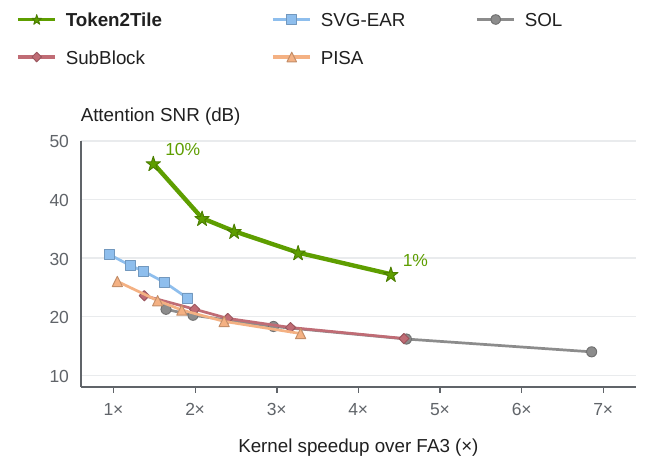}
\caption{\textbf{H3 kernel speed--accuracy trade-off.} Points average captured inputs. \method sweeps 10--1\% budgets; SOL sweeps $\tau=0.25$--2. Other kernels match \method{}\textquotesingle s executed density; hollow markers denote SVG-EAR at higher density. Four-model results: \cref{fig:speed-accuracy-all,tab:kernel-compare}.}
\label{fig:speed-accuracy}
\end{figure}

%% file: figures/tile_execution_fig.tex
\begin{figure}[t]
\centering
\includegraphics[width=\columnwidth]{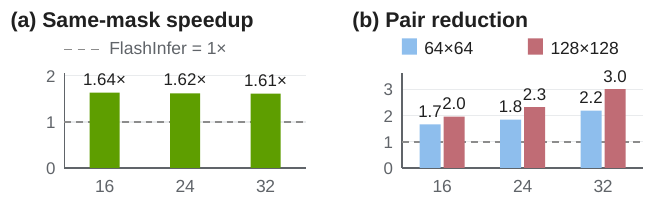}
\caption{\textbf{Tile execution} (H3, steps 16/24/32). (a) Same-mask exact-kernel speedup over FlashInfer. (b) Pair reduction against fixed-block covers. Details: \cref{tab:backends}.}
\label{fig:tile-execution}
\end{figure}

%% file: tables/ablation_main.tex
\begin{table}[t]
\centering
\caption{\textbf{Component ablations at matched executed density.} Changes from \method (positive = better). (a) H3 video PSNR at 2.5\%. (b,c) Attention SNR at 3.5\%: (b) ranges across generators; (c) the same summed-mass decisions and per-query residuals with two cluster budget policies. Shared variants use the query-cluster mean; PCA is principal component analysis. Full controls: \cref{sec:ablation-details}.}
\label{tab:ablation-main}
\footnotesize
\setlength{\tabcolsep}{4pt}
\renewcommand{\arraystretch}{1.08}
\begin{tabularx}{\columnwidth}{@{}Xrr@{}}
\toprule
\multicolumn{3}{@{}l}{\textbf{(a) Video fidelity: $\Delta$PSNR (dB)}}\\
Variant & T2V & I2V\\
\midrule
No per-query residual & $-$1.80 & $-$3.25\\
Shared selection & +0.04 & $-$0.99\\
Shared selection, no residual & $-$1.43 & $-$2.76\\
\addlinespace[2pt]
Per-tile decision, union budget & +0.16 & $-$0.08\\
\addlinespace[2pt]
Original 128-D space & $-$0.04 & $-$0.66\\
Shared 32-D PCA projection & $-$0.09 & $-$0.57\\
\bottomrule
\end{tabularx}
\par\vspace{4pt}
\begin{tabularx}{\columnwidth}{@{}Xr@{}}
\toprule
\multicolumn{2}{@{}l}{\textbf{(b) Attention accuracy: $\Delta$SNR (dB)}}\\
Variant & Range\\
\midrule
No per-query residual & $-$26.5 to $-$1.6\\
Residual shared within query cluster & $-$9.9 to $-$1.0\\
Shared selection & $-$2.3 to $-$0.9\\
\addlinespace[2pt]
Union selection, uniform budgets & $-$0.9 to $-$0.3\\
\bottomrule
\end{tabularx}
\par\vspace{4pt}
\begin{tabularx}{\columnwidth}{@{}Xrrrr@{}}
\toprule
\multicolumn{5}{@{}l}{\textbf{(c) Cluster budgets: $\Delta$SNR (dB)}}\\
Allocation & H3 & Wan & Hun. & LTX\\
\midrule
Union-derived & +0.15 & $-$0.36 & +0.03 & $-$0.19\\
Uniform & $-$0.80 & $-$0.77 & $-$0.32 & $-$0.66\\
\bottomrule
\end{tabularx}
\end{table}

%% file: tables/union_growth_main.tex
\begin{table}[t]
\centering
\caption{\textbf{Predicting union growth.} After/before ratio of union inflation when query-cluster size doubles (3.5\% budget). Model: \cref{sec:union-model}.}
\label{tab:union-growth}
\footnotesize
\setlength{\tabcolsep}{4pt}
\begin{tabularx}{\columnwidth}{@{}Xrrrr@{}}
\toprule
 & H3 & Wan & Hun. & LTX\\
\midrule
Measured & 1.33 & 1.25 & 1.26 & 1.27\\
Predicted & 1.32 & 1.26 & 1.28 & 1.25\\
\bottomrule
\end{tabularx}
\end{table}

%% file: figures/budget_alloc_main_fig.tex
\begin{figure}[t]
\centering
\includegraphics[width=\columnwidth]{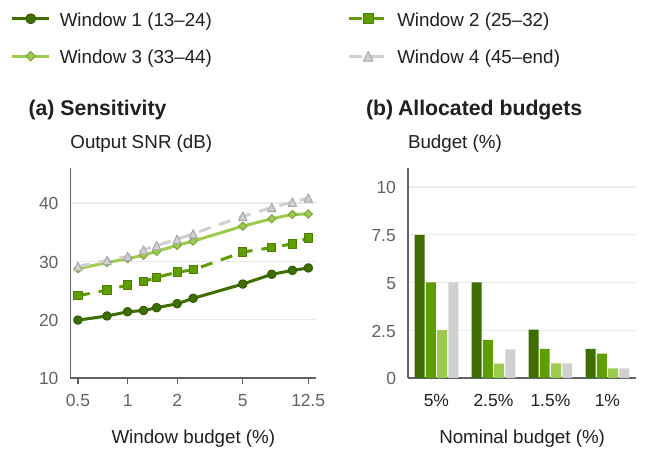}
\caption{\textbf{Denoising-time budget allocation} (H3 T2V). (a) Denoiser output SNR when one window is sparsified (higher: less sensitive). (b) Calibrated window budgets. Wan and gains over uniform schedules: \cref{fig:budget-alloc}.}
\label{fig:budget-alloc-main}
\end{figure}

%% file: sections/related_work.tex
\section{Related Work}
\ifsubmission
\paragraph{Sparse video attention.}
Exact and quantized kernels~\cite{flashattention,flashattention2,flashattention3,flashattention4,sageattention,sageattention2,flashinfer} are complementary to sparsity. Video generators~\cite{cogvideox,hunyuanvideo,wan,minimaxh3} motivate static patterns~\cite{svg,sta,radial} and dynamic block routing by learned tiles~\cite{vsa,vsa2}, low-resolution drafts~\cite{draftattention,draftattention2}, antidiagonal scores~\cite{xattention}, online filtering~\cite{spargeattn}, pooled-score thresholds~\cite{solattn}, or sub-block scores~\cite{subblock}; SVG2 and SVOO cluster queries and keys~\cite{svg2,svoo}. Earlier, clustered attention selected keys per query group with a centroid approximation~\cite{clusteredattention}, Reformer and Routing Transformer routed by hashing or $k$-means~\cite{reformer,routingtransformer}, and LLM decoding retrieves key clusters~\cite{clusterkv,retrievalattention,magicpig} or scores keys in a reduced space~\cite{loki,sparq}. These methods score from a group's or block's representative, or reduce $K$ alone; \method scores every query in a space fitted to $Q$ and $K$; its unions allocate exact work and form execution tiles.

\paragraph{Recovering omitted interactions.}
SLA adds a learned linear branch~\cite{sla}, PISA Taylor terms for skipped blocks~\cite{pisa}, SOL a pooled zeroth-order correction~\cite{solattn}, SVG-EAR query-specific centroid compensation~\cite{svgear}, and SparsePR an affine residual from probe queries~\cite{sparsepr}. \method takes its residual logits from the same closed-form factors that drive selection, over exactly the clusters outside the executed union. The closest designs keep less per query: clustered attention and SVG2 decide from a group centroid, SVG-EAR routes per group from raw-space clusters rebuilt every call, and Loki and SparQ select per token without tiles.
\fi

%% file: sections/conclusion.tex
\section{Conclusion}
\ifsubmission
\method combines per-query scores and residuals in one online low-rank space with exact attention over query-cluster unions. These unions organize tiles and adapt the allocation of exact work to the members' selections. With the online structure at about a tenth of a call, it reaches $1.48$--$1.54\times$ end to end on MiniMax-H3 and Wan2.1-14B at 4.6--6.8\,dB higher fidelity than SOL. The implementation targets noncausal attention on Hopper with $d=128$, and its cluster-logit workspace grows with $N\times C_K$.
\fi

%% file: sections/appendix_arxiv.tex
\raggedbottom
\makeatletter
\newenvironment{suppfigure}[1][]{\par\addvspace{2pt}\noindent\begin{minipage}{\linewidth}\def\@captype{figure}}{\end{minipage}\par\addvspace{2pt}}
\newenvironment{supptable}[1][]{\par\addvspace{2pt}\noindent\begin{minipage}{\linewidth}\def\@captype{table}}{\end{minipage}\par\addvspace{2pt}}
\let\figure\suppfigure\let\endfigure\endsuppfigure
\let\table\supptable\let\endtable\endsupptable
\expandafter\let\csname figure*\endcsname\suppfigure
\expandafter\let\csname endfigure*\endcsname\endsuppfigure
\expandafter\let\csname table*\endcsname\supptable
\expandafter\let\csname endtable*\endcsname\endsupptable
\makeatother
\renewcommand{\topfraction}{0.95}\renewcommand{\dbltopfraction}{0.95}
\renewcommand{\bottomfraction}{0.9}\renewcommand{\textfraction}{0.05}
\renewcommand{\floatpagefraction}{0.45}\renewcommand{\dblfloatpagefraction}{0.45}
\setcounter{topnumber}{4}\setcounter{bottomnumber}{3}\setcounter{totalnumber}{6}\setcounter{dbltopnumber}{4}
\makeatletter
\let\appendix@xfloat\@xfloat
\def\@xfloat#1[#2]{\appendix@xfloat{#1}[tbp]}
\setlength{\@fptop}{0pt}\setlength{\@dblfptop}{0pt}
\setlength{\@fpsep}{12pt plus 2pt}\setlength{\@dblfpsep}{12pt plus 2pt}
\makeatother
\section{Derivations}
\label{sec:full-derivations}
\subsection{Optimal logit-preserving projection}
For one head, let $\Theta=P_qP_k^\top$, $E=I-\Theta$, and let $M_q=Q^\top Q/N$, $M_k=K^\top K/N$ be the uncentered second moments. Expanding the Frobenius norm gives
\begin{align}
 J(\Theta)&=\frac{\|QEK^\top\|_F^2}{N^2}\nonumber\\
 &=\operatorname{tr}(EM_kE^\top M_q)\nonumber\\
 &=\|M_q^{1/2}M_k^{1/2}-\Xi\|_F^2,
\end{align}
where $\Xi=M_q^{1/2}\Theta M_k^{1/2}$. For positive-definite moments this change of variables preserves rank. Writing $M_q^{1/2}M_k^{1/2}=U\Sigma Z^\top$, Eckart--Young~\cite{eckartyoung} yields
\begin{equation}
 \Xi^\star=U_r\Sigma_rZ_r^\top,\qquad
 \min J=\sum_{j>r}\sigma_j^2.
\end{equation}
Undoing the change of variables and splitting $\Sigma_r$ equally gives the factors in \cref{eq:projection}. They balance the projected moments,
\begin{equation}
 P_q^\top M_qP_q=P_k^\top M_kP_k=\Sigma_r,
\end{equation}
and recover $P_qP_k^\top=I$ at full rank. In implementation, eigenvalues are floored before inversion; the same derivation applies to the regularized moments.

The geometry also explains the clustering objective. For original-space key centroids $\bar k_c$, the average within-cluster logit distortion is
\begin{equation}
 \frac{1}{N}\sum_i\sum_c\sum_{j\in c}
 \frac{[q_i(k_j-\bar k_c)^\top]^2}{d}
 =\frac{1}{d}\sum_c\sum_{j\in c}\|k_j-\bar k_c\|_{M_q}^2.
\end{equation}
Thus keys should be grouped by their effects on the queries. In the low-rank space the corresponding metric is $\Sigma_r$; Euclidean $k$-means approximates this metric rather than the original $M_q$.

\subsection{Per-query residual}
\label{sec:operator-details}
For key $j$ in cluster $c$, proxy error separates into projection and within-cluster terms:
\begin{equation}
 \ell_{ij}-g_{ic}=\frac{q_i(I-P_qP_k^\top)k_j^\top}{\sqrt d}
 +\frac{\widetilde q_i(\widetilde k_j-\mu_c)^\top}{\sqrt d}.
\end{equation}
Let $w_j=e^{\ell_{ij}}$ and let $\widehat w_j$ equal $w_j$ on $\mathcal E_i$ and $e^{g_{i,z_j}}$ elsewhere. Since both outputs are weighted averages of values with $V_{\max}=\max_j\|v_j\|$,
\begin{equation}
 \|O_i-\widehat O_i\|\le
 \frac{2V_{\max}\sum_j|w_j-\widehat w_j|}{\sum_jw_j}.
\end{equation}
With $\epsilon_i=\max_{j\notin\mathcal E_i}|\ell_{ij}-g_{i,z_j}|$, the total proxy error on omitted keys satisfies $|w_j-\widehat w_j|\le(e^{\epsilon_i}-1)w_j$, giving
\begin{equation}
 \|O_i-\widehat O_i\|\le 2V_{\max}(e^{\epsilon_i}-1)\alpha_i.
\end{equation}
When $\epsilon_i<\log 2$, this is a tighter worst-case bound than the truncation bound $2V_{\max}\alpha_i$; it does not order the actual errors for every input.

\section{Evaluation details}
\label{sec:protocol}
\paragraph{Baseline controls.} SOL uses the SGLang defaults ($\tau=1$, diagonal thresholding, dense layers 0--1); SubBlock uses sparsity 0.75 and $n_q=n_k=4$ with no dense layers, through an SM90 port. Both keep the first ten denoising steps dense, versus twelve for \method; matching their prefix to twelve steps raises H3 T2V PSNR by 1.2\,dB for both. SVG2 and SVG-EAR keep their official 300/1000 query/key clusters, $\mathrm{top}_p=0.9$, and dense fractions 0.20/0.35 for T2V/I2V; their dense calls use FA3. The H3 video ablations use 124 frames and two seeds per prompt.

\paragraph{Metrics and uncertainty.} PSNR uses the mean frame MSE, SSIM is averaged over frames, and LPIPS uses AlexNet. Video intervals use 5{,}000 prompt-bootstrap replicates.

\paragraph{Blind study.}\label{sec:human-study}
Raters compare randomly ordered, unnamed accelerated videos against a synchronized dense reference. Two rounds share this procedure; the second adds Wan trials against SVG-EAR. Participants include an author. Failed attention checks and answers before one playback are excluded. \Cref{tab:blind} pools both rounds.
\input{tables/blind_study}

\paragraph{Captured-attention controls.} Inputs pair steps/layers 16/8, 24/16, and 32/24; LTX uses the first pass at steps 10/15/20. SOL uses pooled $64\times64$ scores and row thresholds, XAttention its official stride-16 antidiagonal estimate on $128\times128$ blocks with the cumulative threshold matched to $\rho$, and SVG2 clusters in the original space. SVG-EAR uses its official processor and kernels, and PISA its official covariance-aware kernel. Kernel times are CUPTI duration sums measured alternately with FA3 on identical inputs; they include all GPU routing and structure kernels and exclude host-side time for every method. On 15 of the 21 captures, $k$-means maintenance runs on a non-fused path and is counted under ``Gather, merge, misc.'' in \cref{fig:kernel-overhead}; on the others it takes 1.6--1.9\% of an $R=4$ call at 3.5\%.

\section{Supplementary results}
\label{sec:operator-expanded}
\subsection{Accuracy at equal density}
\label{sec:app-equal}
\Cref{fig:methods-equal-all,tab:operator-compare} give three budgets and the component controls. \Cref{tab:same-granularity} gives the block rules as many key blocks as \method has key clusters: their SNR changes by $-0.4$ to $+1.4$\,dB, so the gap is not a matter of finer blocks.
\input{figures/methods_equal_all}
\input{tables/operator_compare}
\input{tables/same_granularity}

\subsection{Kernel performance}
\label{sec:app-kernel}
\Cref{fig:speed-accuracy-all,tab:kernel-compare,fig:kernel-overhead} extend the kernel comparison across generators, and \cref{tab:backends,tab:attn-memory} isolate exact-attention execution and memory. \Cref{tab:same-density-kernel} times \method, the fixed-block kernel, and SOL at the same executed density in one process: \method is faster than SOL in every case, and the routed fixed-block kernel is faster at 1\% and on LTX-2.3. End to end on Wan T2V, SVG-EAR reaches $1.11\times$ on one GPU in SGLang; on the per-rank inputs of an Ulysses-8 run, its clustering and error-map scoring take about 6\,ms on every call.
\input{figures/speed_accuracy_all}
\input{tables/kernel_compare}
\input{tables/attn_memory}
\input{tables/same_density_kernel}
\input{figures/kernel_overhead_fig}
\input{tables/backend_controls}

\subsection{The low-rank space}
\label{sec:app-lowrank}
\Cref{fig:dimred,tab:dimred} compare low-rank spaces, and \cref{tab:rank-split} varies the dimension separately for selection and the per-query residual. The full-call estimates in \cref{sec:dimred} substitute the 128-dimensional residual time, and a quarter of the assignment time (one refresh in four calls), of \cref{tab:dimred-fused} into the measured $R=4$ call and keep the other stages fixed; the estimate is the time saved relative to the 128-dimensional call.

\noindent\begin{minipage}[t]{0.49\textwidth}\input{figures/dimred_fig}\end{minipage}\hfill
\begin{minipage}[t]{0.49\textwidth}\input{figures/ablation_summary_fig}\end{minipage}\par
\input{tables/dimred}
\input{tables/dimred_fused}
\input{tables/rank_split}

\subsection{Query clusters, unions, and reuse}
\label{sec:ablation-details}
\Cref{tab:overlap,tab:union,tab:group-size} report selection overlap, union size, and query-cluster size.
\input{tables/selection_overlap}
\input{tables/union_inflation}
\input{tables/group_size}

\paragraph{Predicting union density.}\label{sec:union-model}
Let a query cluster hold $m$ queries $\widetilde q_i=\bar q+\xi_i$ with covariance $\Gamma$, and let $u_c=\mu_c/\sqrt d$ with size-weighted mean $\bar u$. Removing the score shift $\xi_i\bar u^\top$ common to all key clusters leaves the perturbation that reorders them,
\begin{equation}
 \Delta_{ic}=\xi_i(u_c-\bar u)^\top,\qquad
 \sigma_c^2=(u_c-\bar u)\Gamma(u_c-\bar u)^\top.
\end{equation}
We model the perturbations as Gaussian, the member selections as independent, and the budget as a common threshold $\theta$. With $\bar g_c=\bar q u_c^\top$ and the normal tail $\bar\Phi=1-\Phi$, a member selects cluster $c$ with probability $p_c$, the cluster enters the union with probability $1-(1-p_c)^m$, and the expected key count and pair density follow:
\begin{equation}
 p_c=\bar\Phi\!\left(\frac{\theta-\bar g_c}{\sigma_c}\right),\quad \sum_c n_cp_c\approx bN,\qquad
 \widehat L_a=\sum_c n_c[1-(1-p_c)^m],\qquad
 \widehat\rho=\frac{1}{N^2}\sum_a |\mathcal Q_a|\widehat L_a.
\end{equation}
These use only query-cluster covariances and key centroids. If $\bar g_c$ is Gaussian across clusters with standard deviation $s$ and $\sigma_c\approx\sigma$, the largest member perturbation shifts the threshold by about $\kappa_m\sigma$ with $\kappa_m\sim\sqrt{2\ln m}$:
\begin{equation}
 \frac{\widehat L_a}{bN}\approx
 \frac{\bar\Phi(z_b-\kappa_m\eta)}{b},\qquad
 z_b=\Phi^{-1}(1-b),\qquad \eta=\frac{\sigma}{s}.
\end{equation}
More disagreement among members thus gives a larger union; this closed form is qualitative and overestimates the measured inflation by 1.2--2.3$\times$. \Cref{tab:union-model} tests the full expression: rank correlation is 0.95--0.99, with overestimation at small budgets, consistent with positively correlated members.
\input{tables/union_model}

\paragraph{Scores and allocation.}
\Cref{tab:e1-aggregation,tab:tile-video} separate scoring, decisions, and allocation. Per-tile decisions keep each cluster\textquotesingle s union budget; the uniform-budget and global-greedy rows match only the total pair count, and the dispersion row scales the union budgets by the $\eta$ of \cref{sec:union-model}. \Cref{fig:ablation-summary} adds the video control with a shared residual. These are accuracy controls; the video variants were not timed.
\input{tables/e1_aggregation}
\input{tables/tile_video}

\paragraph{Clustering variability.}\label{sec:noise}
Two more $k$-means initializations change operator SNR by 0.03--0.53\,dB, and the effect of the residual dimension by less than 0.1\,dB. Shared selection and uniform cluster budgets stay less accurate, and per-tile decisions stay within $-0.52$ to $+0.41$\,dB of the union. Global greedy allocation improves only H3 at 3.5\%, by 0.07--0.28\,dB. H3 selection results above 16 dimensions vary by up to 2.4\,dB; on the other generators, 24--128 dimensions stay within 0.65\,dB of 32.

\paragraph{Reuse.} \Cref{tab:reuse} separates stale centroids from stale selections across refresh intervals.
\input{tables/reuse_interval}

\subsection{Denoising-time budgets}
\label{sec:app-budget}
The generation results use reference schedules, each calibrated on one input (\cref{fig:budget-alloc}). At 2.5\%, the non-uniform schedules from other inputs differ from the reference by at most 0.32\,dB on average (\cref{tab:sensitivity_test025}).
\input{figures/budget_alloc_fig}
\input{tables/sensitivity_test025}

\subsection{Qualitative examples}
\Cref{fig:qual-frames} shows four cases.
\input{figures/qual_frames}

%% file: tables/blind_study.tex
\begin{table}[t]
\centering
\caption{\textbf{Blind study.} Win/tie/loss percentages and score $=$ win $+\frac12$ tie; 95\% intervals bootstrap raters. Block-sparse combines SOL and SubBlock. Speedups span T2V/I2V at the rated settings: main-table budgets on H3, 5\% on Wan.}
\label{tab:blind}
\scriptsize
\setlength{\tabcolsep}{2.5pt}
\begin{tabular}{@{}lrrrrrr@{}}
\toprule
 & & & & & \multicolumn{2}{c}{Speedup ($\times$)}\\
\cmidrule(lr){6-7}
Comparison & Win & Tie & Loss & Score & Ours & Base\\
\midrule
All & 46 & 44 & 10 & 68 [62, 74] & -- & --\\
\midrule
Block-sparse & 66 & 29 & 5 & 81 [77, 85] & 1.38--1.49 & 1.41--1.49\\
SVG2 and SVG-EAR & 27 & 58 & 15 & 56 [52, 61] & 1.38--1.40 & 0.96--1.07\\
\midrule
H3, vs SOL & 71 & 24 & 5 & 83 [74, 92] & 1.48--1.49 & 1.48--1.49\\
H3, vs SubBlock & 58 & 32 & 11 & 74 [62, 87] & 1.48--1.49 & 1.47--1.48\\
Wan, vs SOL & 42 & 53 & 5 & 68 [53, 79] & 1.38--1.40 & 1.41--1.42\\
Wan, vs SubBlock & 94 & 6 & 0 & 97 [92, 100] & 1.38--1.40 & 1.43\\
Wan, vs SVG2 & 29 & 62 & 10 & 60 [48, 71] & 1.38--1.40 & 1.06--1.07\\
Wan, vs SVG-EAR & 27 & 57 & 17 & 55 [50, 57] & 1.38--1.40 & 0.96--0.97\\
\bottomrule
\end{tabular}
\end{table}

%% file: figures/methods_equal_all.tex
\begin{figure*}[t]
\centering
\includegraphics[width=\textwidth]{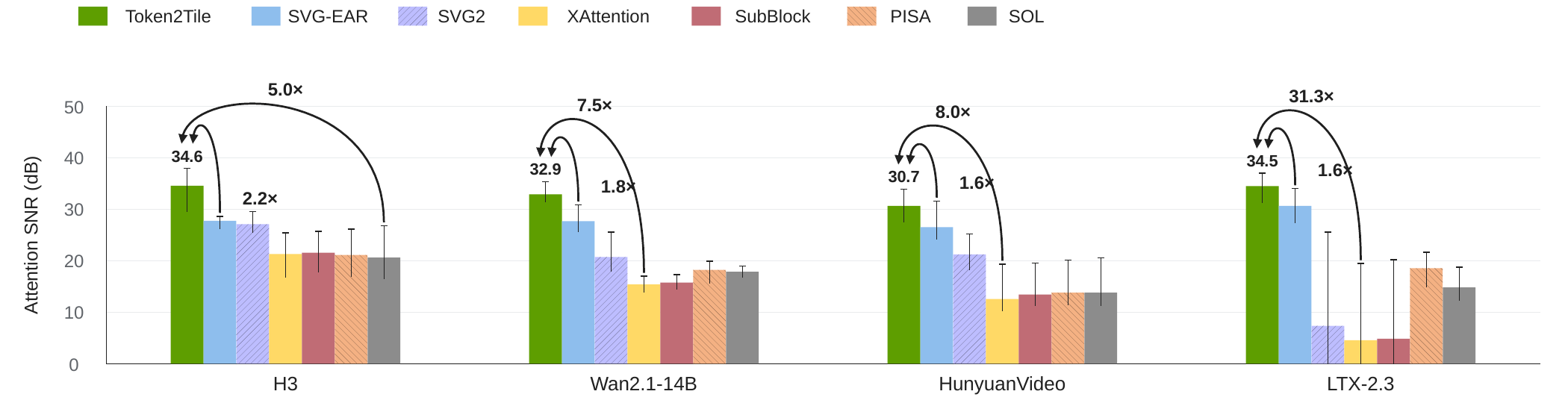}
\caption{\textbf{Equal-density accuracy across generators} (3.5\% budget). Whiskers span inputs; arrows give relative-error ratios of the nearest and farthest alternatives to \method. Full values: \cref{tab:operator-compare}.}
\label{fig:methods-equal-all}
\end{figure*}

%% file: tables/operator_compare.tex
\begin{table*}[t]
\centering
\caption{\textbf{Accuracy and component controls at matched density.} Attention SNR (dB), averaged over inputs. Controls give $\Delta$SNR against FP32 \method with the same state: no residual, shared residual, original-space centroid scores, or shared selection. Shared variants use the query-cluster mean. The \method column is the real kernel output; controls are FP32 differences. On H3, pinned context is exact for every rule except the official SVG-EAR kernel, unlike the real kernels of \cref{tab:kernel-compare}. $^\dagger$SVG-EAR exceeds the target density.}
\label{tab:operator-compare}
\footnotesize
\setlength{\tabcolsep}{4pt}
\begin{tabular}{llrrrrrrrrrrr}
\toprule
 & & & \multicolumn{4}{c}{Controls, $\Delta$ dB} & \multicolumn{6}{c}{Other rules at the same $\rho$}\\
\cmidrule(lr){4-7}\cmidrule(lr){8-13}
Budget (\%) & $\rho$ (\%) & \method & w/o res. & shared res. & 128-D res. & shared sel. & SVG-EAR & SVG2 & XAttn & SubBlock & PISA & SOL\\
\midrule
\multicolumn{13}{l}{\textit{H3}}\\
1.0 & 11.7 & \textbf{27.2} & $-$1.8 & $-$1.0 & +0.3 & $-$2.0 & 23.2 & 19.8 & 15.7 & 16.7 & 17.1 & 16.5\\
3.5 & 25.7 & \textbf{34.6} & $-$1.6 & $-$1.0 & +0.2 & $-$2.1 & 27.7 & 27.1 & 21.3 & 21.6 & 21.1 & 20.6\\
10.0 & 47.2 & \textbf{46.2} & $-$2.0 & $-$1.4 & +0.3 & $-$4.2 & 30.6 & 36.3 & 28.4 & 27.0 & 26.0 & 25.2\\
\midrule
\multicolumn{13}{l}{\textit{Wan2.1-14B}}\\
1.0 & 7.6 & \textbf{25.8} & $-$9.2 & $-$2.8 & +0.8 & $-$1.9 & 25.4$^\dagger$ & 15.1 & 10.7 & 11.2 & 13.6 & 12.8\\
3.5 & 20.0 & \textbf{32.9} & $-$10.6 & $-$3.3 & +0.7 & $-$2.3 & 27.7 & 20.8 & 15.4 & 15.8 & 18.2 & 17.9\\
10.0 & 41.0 & \textbf{41.0} & $-$12.2 & $-$3.9 & +0.8 & $-$3.1 & 32.7 & 26.8 & 20.8 & 21.2 & 23.9 & 23.1\\
\midrule
\multicolumn{13}{l}{\textit{HunyuanVideo}}\\
1.0 & 6.8 & \textbf{24.4} & $-$6.2 & $-$2.9 & +0.6 & $-$0.7 & 23.4$^\dagger$ & 15.4 & 9.3 & 9.9 & 10.4 & 10.4\\
3.5 & 17.2 & \textbf{30.7} & $-$6.0 & $-$2.9 & +0.4 & $-$0.9 & 26.5 & 21.2 & 12.6 & 13.4 & 13.8 & 13.8\\
10.0 & 36.2 & \textbf{38.4} & $-$5.8 & $-$2.7 & +0.4 & $-$1.1 & 30.5 & 28.3 & 17.1 & 18.0 & 18.0 & 17.8\\
\midrule
\multicolumn{13}{l}{\textit{LTX-2.3}}\\
1.0 & 6.4 & \textbf{29.6} & $-$24.6 & $-$8.8 & +0.6 & $-$1.2 & 29.6$^\dagger$ & 4.1 & 1.4 & 1.9 & 15.6 & 10.2\\
3.5 & 14.8 & \textbf{34.5} & $-$26.5 & $-$9.9 & +0.5 & $-$1.0 & 30.7 & 7.4 & 4.6 & 4.9 & 18.6 & 14.8\\
10.0 & 30.0 & \textbf{40.5} & $-$28.7 & $-$11.3 & +0.6 & $-$0.9 & 33.9 & 11.3 & 8.3 & 8.4 & 22.1 & 19.1\\
\bottomrule
\end{tabular}
\end{table*}

%% file: tables/same_granularity.tex
\begin{table*}[t]
\centering
\caption{\textbf{Block rules at \method{}\textquotesingle s granularity.} Attention SNR (dB) at \method{}\textquotesingle s executed density. Each block rule runs with $64\times64$ blocks and with as many key blocks as \method has key clusters: 128-query blocks over 2048 contiguous key blocks (block shape after the model name). SOL: row threshold with zeroth-order correction; top-$k$: per-query-block top-$k$ key blocks, exact only or with the same correction.}
\label{tab:same-granularity}
\footnotesize
\setlength{\tabcolsep}{4pt}
\begin{tabular}{@{}lrrrrrrrrr@{}}
\toprule
 & & & & \multicolumn{2}{c}{SOL} & \multicolumn{2}{c}{Top-$k$} & \multicolumn{2}{c}{Top-$k$ + corr.}\\
\cmidrule(lr){5-6}\cmidrule(lr){7-8}\cmidrule(lr){9-10}
Model (fine block) & $b$ & $\rho$ & \method & 64 & fine & 64 & fine & 64 & fine\\
\midrule
H3 (128$\times$36) & 1 & 11.7 & \textbf{27.2} & 16.5 & 16.3 & 16.2 & 15.8 & 16.6 & 16.3\\
 & 3.5 & 25.7 & \textbf{34.6} & 20.6 & 21.0 & 20.3 & 20.6 & 20.6 & 20.9\\
 & 10 & 47.2 & \textbf{46.2} & 25.2 & 26.1 & 25.2 & 25.9 & 25.4 & 26.1\\
\addlinespace[1pt]
Wan2.1-14B (128$\times$37) & 1 & 7.6 & \textbf{25.8} & 12.8 & 13.0 & 11.0 & 11.0 & 13.5 & 13.5\\
 & 3.5 & 20.0 & \textbf{32.9} & 17.9 & 18.1 & 15.7 & 15.8 & 17.9 & 18.1\\
 & 10 & 41.0 & \textbf{41.0} & 23.1 & 23.6 & 21.0 & 21.3 & 23.3 & 23.7\\
\addlinespace[1pt]
HunyuanVideo (128$\times$71) & 1 & 6.8 & \textbf{24.4} & 10.4 & 10.2 & 9.7 & 9.5 & 10.5 & 10.3\\
 & 3.5 & 17.2 & \textbf{30.7} & 13.8 & 13.4 & 13.3 & 13.0 & 13.8 & 13.5\\
 & 10 & 36.2 & \textbf{38.4} & 17.8 & 17.5 & 17.5 & 17.2 & 17.9 & 17.5\\
\addlinespace[1pt]
LTX-2.3 (128$\times$32) & 1 & 6.4 & \textbf{29.6} & 10.2 & 11.7 & 1.5 & 1.8 & 12.5 & 13.2\\
 & 3.5 & 14.8 & \textbf{34.5} & 14.8 & 15.9 & 4.5 & 4.8 & 15.5 & 16.5\\
 & 10 & 30.0 & \textbf{40.5} & 19.1 & 20.5 & 8.1 & 8.5 & 19.1 & 20.5\\
\bottomrule
\end{tabular}
\end{table*}

%% file: figures/speed_accuracy_all.tex
\begin{figure*}[t]
\centering
\includegraphics[width=\textwidth]{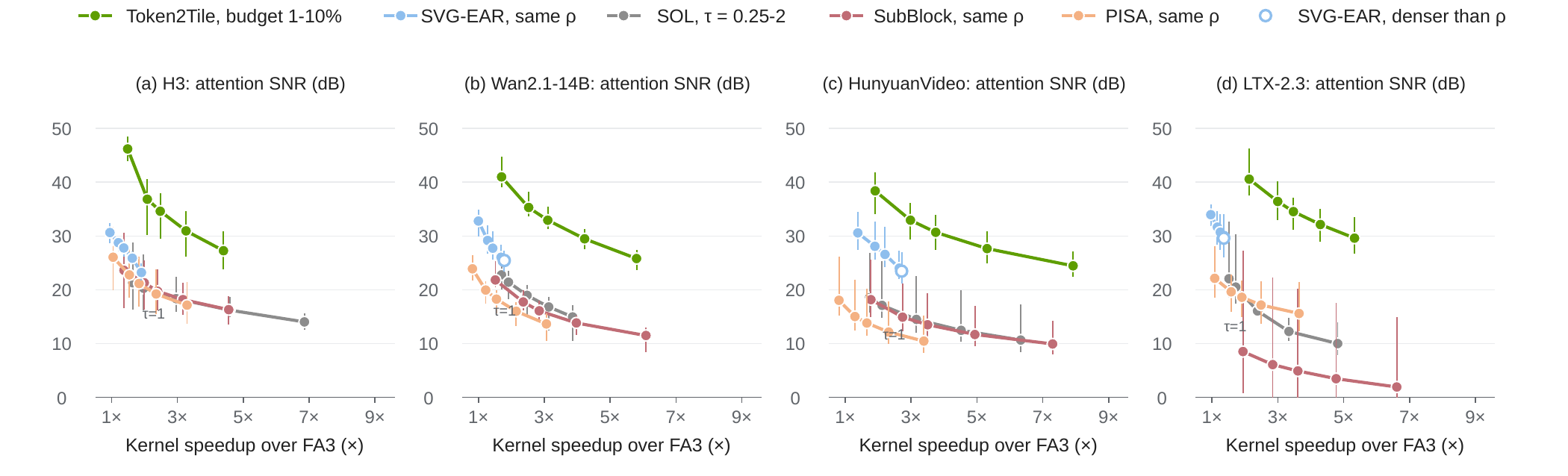}
\caption{\textbf{Kernel speed--accuracy curves across generators.} Points average inputs; bars span them. Competing kernels match \method{}\textquotesingle s executed density, except hollow SVG-EAR points at higher density. Values: \cref{tab:kernel-compare}.}
\label{fig:speed-accuracy-all}
\end{figure*}

%% file: tables/kernel_compare.tex
\begin{table*}[t]
\centering
\caption{\textbf{Kernel speed and accuracy.} Sp.: speedup over FA3 from summed times; SNR: mean real-kernel output accuracy (dB). \method rows average one refresh and three reuse calls ($R=4$). Densities match except below SVG-EAR\textquotesingle s minimum. Timing scope: \cref{sec:protocol}.}
\label{tab:kernel-compare}
\footnotesize
\setlength{\tabcolsep}{4pt}
\begin{tabular}{lrrrrrrrr}
\toprule
 & \multicolumn{2}{c}{H3} & \multicolumn{2}{c}{Wan2.1-14B} & \multicolumn{2}{c}{HunyuanVideo} & \multicolumn{2}{c}{LTX-2.3}\\
\cmidrule(lr){2-3}\cmidrule(lr){4-5}\cmidrule(lr){6-7}\cmidrule(lr){8-9}
Setting & Sp.$\uparrow$ & SNR$\uparrow$ & Sp.$\uparrow$ & SNR$\uparrow$ & Sp.$\uparrow$ & SNR$\uparrow$ & Sp.$\uparrow$ & SNR$\uparrow$\\
\midrule
\multicolumn{9}{l}{\textit{\method, per-query budget}}\\
1.0\% & 4.40 & 27.2 & 5.80 & 25.8 & 7.92 & 24.4 & 5.33 & 29.6\\
2.0\% & 3.26 & 30.9 & 4.23 & 29.4 & 5.31 & 27.6 & 4.29 & 32.1\\
3.5\% & 2.48 & 34.6 & 3.11 & 32.9 & 3.75 & 30.7 & 3.47 & 34.5\\
5.0\% & 2.08 & 36.8 & 2.53 & 35.3 & 2.98 & 32.9 & 3.00 & 36.4\\
10.0\% & 1.49 & 46.2 & 1.70 & 41.0 & 1.91 & 38.4 & 2.13 & 40.5\\
\midrule
\multicolumn{9}{l}{\textit{SOL kernel, threshold $\tau$}}\\
$\tau=0.25$ & 1.64 & 21.3 & 1.70 & 22.7 & 1.74 & 18.5 & 1.52 & 22.0\\
$\tau=0.5$ & 1.97 & 20.3 & 1.91 & 21.4 & 2.12 & 17.1 & 1.72 & 20.5\\
$\tau=1$ & 2.96 & 18.3 & 2.48 & 18.9 & 3.16 & 14.5 & 2.38 & 16.0\\
$\tau=1.5$ & 4.59 & 16.2 & 3.14 & 16.8 & 4.52 & 12.4 & 3.34 & 12.2\\
$\tau=2$ & 6.86 & 14.0 & 3.86 & 14.9 & 6.33 & 10.6 & 4.82 & 10.0\\
\midrule
\multicolumn{9}{l}{\textit{SubBlock kernel at \method{}\textquotesingle s density}}\\
1.0\% & 4.56 & 16.3 & 6.09 & 11.5 & 7.30 & 9.9 & 6.61 & 1.9\\
3.5\% & 2.40 & 19.7 & 2.85 & 16.1 & 3.50 & 13.5 & 3.61 & 4.9\\
10.0\% & 1.38 & 23.6 & 1.52 & 21.8 & 1.78 & 18.1 & 1.94 & 8.5\\
\midrule
\multicolumn{9}{l}{\textit{SVG-EAR kernels at \method{}\textquotesingle s density}}\\
1.0\% & 1.90 & 23.2 & 1.79 & 25.4 & 2.71 & 23.4 & 1.36 & 29.6\\
3.5\% & 1.37 & 27.7 & 1.44 & 27.7 & 2.21 & 26.5 & 1.25 & 30.7\\
10.0\% & 0.95 & 30.6 & 1.00 & 32.7 & 1.38 & 30.5 & 0.96 & 33.9\\
\midrule
\multicolumn{9}{l}{\textit{PISA kernel at \method{}\textquotesingle s density}}\\
1.0\% & 3.29 & 17.1 & 3.07 & 13.6 & 3.39 & 10.4 & 3.64 & 15.6\\
3.5\% & 1.84 & 21.1 & 1.55 & 18.2 & 1.66 & 13.8 & 1.91 & 18.6\\
10.0\% & 1.05 & 26.0 & 0.82 & 23.9 & 0.81 & 18.0 & 1.08 & 22.1\\
\midrule
FA3 (ms) & \multicolumn{2}{c}{27.98} & \multicolumn{2}{c}{21.20} & \multicolumn{2}{c}{46.98} & \multicolumn{2}{c}{12.46}\\
\bottomrule
\end{tabular}
\end{table*}

%% file: tables/attn_memory.tex
\begin{table*}[t]
\centering
\caption{\textbf{Allocated attention memory per rank} (GiB unless marked). Workspace is shared across layers; $G$ is its cluster-score buffer. Peak/call is transient allocation beyond the workspace. State stores each layer\textquotesingle s labels, centroids, unions, and plans; resident memory sums workspace and all-layer state. Budget: 3.5\%.}
\label{tab:attn-memory}
\footnotesize
\setlength{\tabcolsep}{4pt}
\begin{tabular}{lrrrrrrrrr}
\toprule
Input & $QKV$ & FA3 peak & Workspace & of which $G$ & Peak/call & State/layer (MiB) & Layers & State, all & Resident\\
\midrule
H3 & 0.37 & 0.12 & 2.82 & 1.96 & 0.13 & 11.2 & 50 & 0.55 & 3.37\\
Wan2.1-14B & 0.27 & 0.09 & 2.07 & 1.44 & 0.09 & 8.1 & 40 & 0.31 & 2.39\\
HunyuanVideo & 0.31 & 0.10 & 2.36 & 1.65 & 0.12 & 8.6 & 60 & 0.50 & 2.87\\
LTX-2.3 & 0.19 & 0.06 & 1.43 & 1.00 & 0.08 & 5.8 & 48 & 0.27 & 1.70\\
\bottomrule
\end{tabular}
\end{table*}

%% file: tables/same_density_kernel.tex
\begin{table}[t]
\centering
\caption{\textbf{Kernels at equal executed density.} Speedup over dense FA3 per attention call (GPU kernel time), all kernels interleaved in one process. \method: full call averaged over $R=4$. Fixed-block: the $64\times64$ block-sparse FlashAttention kernel with SubBlock routing, and the kernel alone on a precomputed plan. SOL: its kernel with $\tau$ set to keep the same density. $\rho$: density of the timed calls. Bold: fastest complete method.}
\label{tab:same-density-kernel}
\footnotesize
\setlength{\tabcolsep}{3pt}
\begin{tabular}{@{}lrrrrrrr@{}}
\toprule
 & & & FA3 & & \multicolumn{2}{c}{Fixed-block} & \\
\cmidrule(lr){6-7}
Model & $b$ & $\rho$ & (ms) & \method & routed & kernel & SOL\\
\midrule
H3 & 1 & 12.1 & 28.8 & 4.52 & \textbf{4.90} & 5.64 & 3.67\\
 & 3.5 & 26.0 & 28.7 & \textbf{2.55} & 2.43 & 2.57 & 2.14\\
 & 10 & 47.5 & 29.3 & \textbf{1.54} & 1.39 & 1.43 & 1.40\\
\addlinespace[1pt]
Wan2.1-14B & 1 & 7.9 & 21.5 & 5.76 & \textbf{6.55} & 8.08 & 3.48\\
 & 3.5 & 21.0 & 21.9 & \textbf{3.13} & 3.03 & 3.28 & 2.37\\
 & 10 & 42.5 & 21.8 & \textbf{1.75} & 1.54 & 1.60 & 1.55\\
\addlinespace[1pt]
HunyuanVideo & 1 & 6.8 & 48.2 & 7.91 & \textbf{7.99} & 10.08 & 5.59\\
 & 3.5 & 16.8 & 48.8 & \textbf{3.85} & 3.75 & 3.96 & 3.15\\
 & 10 & 35.2 & 50.2 & \textbf{2.05} & 1.89 & 1.95 & 1.89\\
\addlinespace[1pt]
LTX-2.3 & 1 & 6.4 & 12.5 & 5.13 & \textbf{7.80} & 9.84 & 3.53\\
 & 3.5 & 14.8 & 12.5 & 3.33 & \textbf{4.05} & 4.60 & 2.73\\
 & 10 & 29.9 & 12.7 & 2.14 & \textbf{2.19} & 2.30 & 1.96\\
\bottomrule
\end{tabular}
\end{table}

%% file: figures/kernel_overhead_fig.tex
\begin{figure*}[t]
\centering
\includegraphics[width=\textwidth]{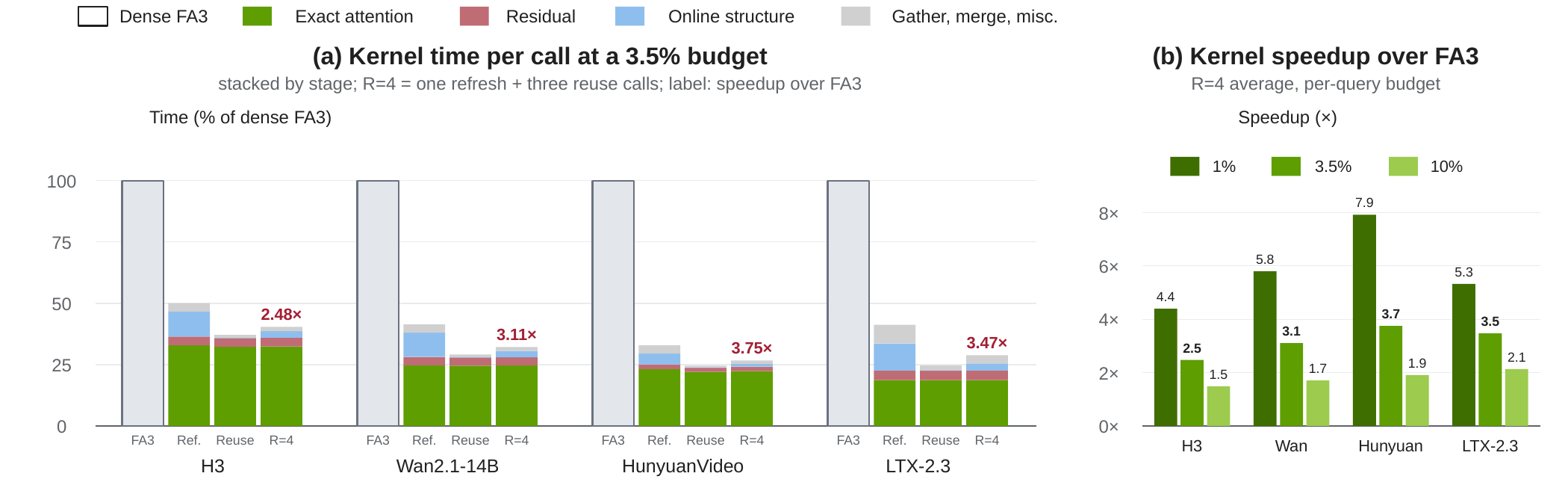}
\caption{\textbf{Kernel cost across generators.} (a) Stage times relative to FA3 at 3.5\%; $R=4$ averages one refresh and three reuse calls. (b) Speedup at three budgets. Labels and colors identify the stages and budgets.}
\label{fig:kernel-overhead}
\end{figure*}

%% file: tables/backend_controls.tex
\begin{table*}[t]
\centering
\caption{\textbf{Exact-attention backend controls.} H3 captures at a 3.5\% budget. Cells: pair density / time (ms). Variable-block attention uses the same mask; fixed-block covers add pairs. These standalone runs exclude planning and the residual; \cref{fig:kernel} instead attributes stages within full calls. Library kernels use default settings, including cuDNN SDPA~\cite{cudnn}.}
\label{tab:backends}
\footnotesize
\setlength{\tabcolsep}{4pt}
\begin{tabular}{lrrr}
\toprule
Backend & Step 16 & Step 24 & Step 32\\
\midrule
\rowcolor{oursrow}
Token2Tile exact & 0.468 / 16.51 & 0.358 / 12.68 & 0.236 / 8.39\\
FlashInfer var.-block & 0.468 / 27.00 & 0.358 / 20.51 & 0.236 / 13.52\\
FA4 128 x 128 cover & 0.913 / 28.61 & 0.828 / 25.68 & 0.708 / 21.80\\
FA4 64 x 64 cover & 0.776 / 34.97 & 0.660 / 29.55 & 0.515 / 23.36\\
Dense FA3 & 1.000 / 29.33 & 1.000 / 28.71 & 1.000 / 28.01\\
Dense FA4 & 1.000 / 28.78 & 1.000 / 28.88 & 1.000 / 28.02\\
FlashInfer dense / FA3 & 1.000 / 28.24 & 1.000 / 28.39 & 1.000 / 28.20\\
FlashInfer dense / FA2 & 1.000 / 48.48 & 1.000 / 48.59 & 1.000 / 48.90\\
cuDNN SDPA & 1.000 / 33.91 & 1.000 / 33.92 & 1.000 / 33.20\\
\bottomrule
\end{tabular}
\end{table*}

%% file: figures/dimred_fig.tex
\begin{figure}[t]
\centering
\includegraphics[width=\columnwidth]{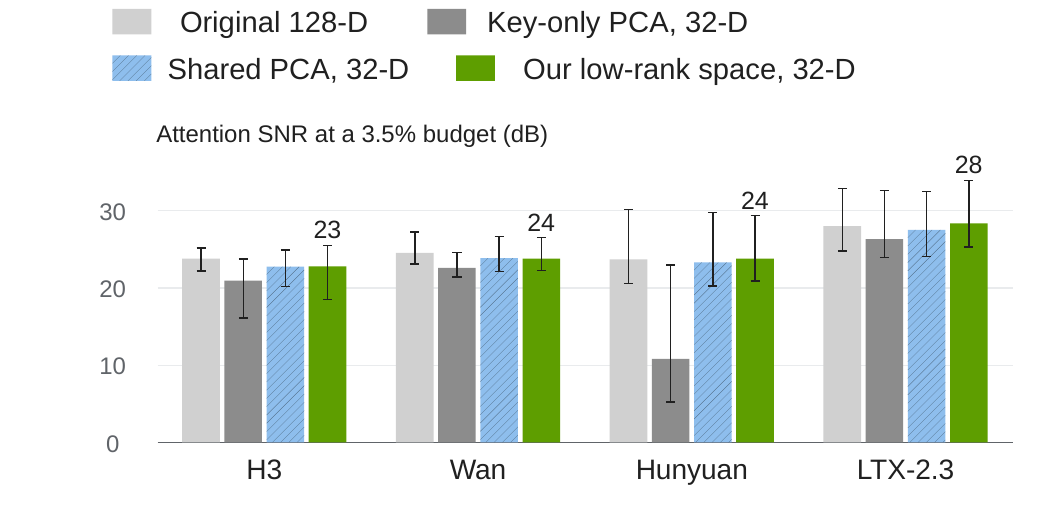}
\caption{\textbf{Accuracy of low-rank spaces} at 3.5\%. Compare the original space, key-only PCA, shared PCA, and our 32-dimensional low-rank space. Each query executes its own prefix without a union, with a per-query residual. Statistics and times: \cref{tab:dimred,tab:dimred-fused}.}
\label{fig:dimred}
\end{figure}

%% file: figures/ablation_summary_fig.tex
\begin{figure}[t]
\centering
\includegraphics[width=\columnwidth]{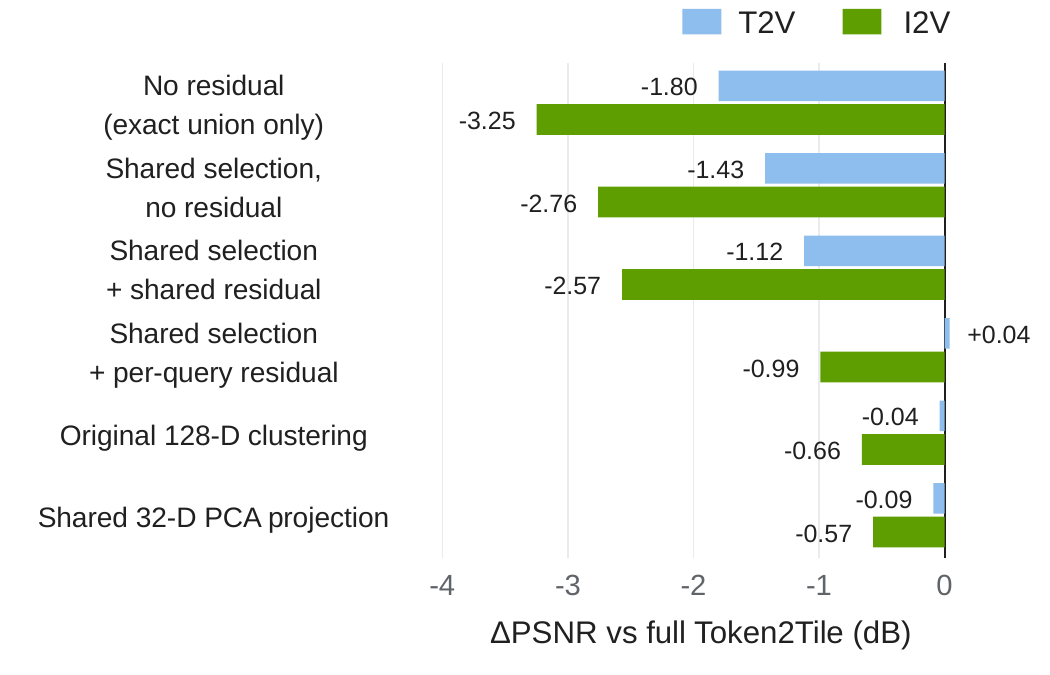}
\caption{\textbf{Video component ablations.} PSNR change from full \method at matched executed density, interpolated over 15--22\%. Changes within 0.1\,dB are treated as ties. The 128-D variant runs on the FlashInfer backend and is compared with a rank-32 control on that backend. Captured-attention controls: \cref{tab:operator-compare}.}
\label{fig:ablation-summary}
\end{figure}

%% file: tables/dimred.tex
\begin{table*}[t]
\centering
\caption{\textbf{Comparing low-rank spaces.} Means over captured inputs. Logit error and Spearman measure agreement with $QK^\top$; within-cluster variance is relative to total logit variance (\%). Mass and SNR use each query\textquotesingle s own cluster prefix without a union, with a per-query residual. SNR at 3.5\% is in \cref{fig:dimred}.}
\label{tab:dimred}
\footnotesize
\setlength{\tabcolsep}{4pt}
\begin{tabular}{lrrrrrr}
\toprule
Space & Logit err.$\downarrow$ & Spearman$\uparrow$ & Within-cl. var.$\downarrow$ & Mass 1\%$\uparrow$ & SNR 1\%$\uparrow$ & Mass 3.5\%$\uparrow$\\
\midrule
\multicolumn{7}{l}{\textit{H3}}\\
Original 128-D & 0.000 & 1.000 & 26.7 & 68.0 & 18.1 & 82.7\\
Key-only PCA, 32-D & 0.227 & 0.944 & 26.4 & 64.9 & 15.9 & 80.9\\
Shared PCA, 32-D & 0.186 & 0.963 & 26.1 & 66.7 & 17.1 & 81.8\\
Our low-rank space, 32-D & 0.181 & 0.965 & 24.3 & 68.0 & 17.4 & 82.9\\
\midrule
\multicolumn{7}{l}{\textit{Wan2.1-14B}}\\
Original 128-D & 0.000 & 1.000 & 22.5 & 61.2 & 20.0 & 76.4\\
Key-only PCA, 32-D & 0.268 & 0.950 & 21.7 & 59.6 & 17.7 & 75.5\\
Shared PCA, 32-D & 0.247 & 0.957 & 21.6 & 59.9 & 18.7 & 75.8\\
Our low-rank space, 32-D & 0.246 & 0.957 & 21.0 & 59.8 & 18.5 & 75.9\\
\midrule
\multicolumn{7}{l}{\textit{HunyuanVideo}}\\
Original 128-D & 0.000 & 1.000 & 26.1 & 62.4 & 19.0 & 78.9\\
Key-only PCA, 32-D & 0.240 & 0.919 & 25.6 & 48.1 & 9.2 & 66.2\\
Shared PCA, 32-D & 0.190 & 0.951 & 25.0 & 60.7 & 18.2 & 78.1\\
Our low-rank space, 32-D & 0.174 & 0.958 & 23.3 & 61.4 & 18.5 & 78.8\\
\midrule
\multicolumn{7}{l}{\textit{LTX-2.3}}\\
Original 128-D & 0.000 & 1.000 & 11.2 & 32.0 & 24.1 & 49.0\\
Key-only PCA, 32-D & 0.149 & 0.981 & 10.7 & 31.0 & 22.5 & 48.2\\
Shared PCA, 32-D & 0.126 & 0.986 & 10.5 & 31.5 & 23.3 & 48.7\\
Our low-rank space, 32-D & 0.116 & 0.988 & 9.9 & 31.7 & 24.0 & 49.0\\
\bottomrule
\end{tabular}
\end{table*}

%% file: tables/dimred_fused.tex
\begin{table*}[t]
\centering
\caption{\textbf{Stage time versus low-rank dimension} (ms). Assignment produces key/query labels; the per-query residual includes centroid scoring and value aggregation. Both use fused kernels with on-chip scores.}
\label{tab:dimred-fused}
\footnotesize
\setlength{\tabcolsep}{4pt}
\begin{tabular}{lrrrrrrrr}
\toprule
Model & $H\times N$ & \multicolumn{3}{c}{Assignment, $r=32/64/128$} & \multicolumn{3}{c}{Residual, $r=32/64/128$} & 128 vs 32\\
\midrule
H3 & 7$\times$73{,}393 & 0.57 & 0.64 & 1.02 & 0.84 & 1.02 & 1.63 & 1.88$\times$\\
Wan2.1-14B & 5$\times$75{,}600 & 0.44 & 0.49 & 0.77 & 0.63 & 0.77 & 0.99 & 1.65$\times$\\
HunyuanVideo & 3$\times$144{,}014 & 0.61 & 0.70 & 1.05 & 0.72 & 0.88 & 1.39 & 1.84$\times$\\
LTX-2.3 & 4$\times$65{,}280 & 0.30 & 0.33 & 0.53 & 0.44 & 0.54 & 0.71 & 1.68$\times$\\
\bottomrule
\end{tabular}
\end{table*}

%% file: tables/rank_split.tex
\begin{table*}[t]
\centering
\caption{\textbf{Dimension of the low-rank space.} First row: attention SNR (dB) with clusters rebuilt in FP32 for every dimension, so it differs from \cref{tab:operator-compare}; other rows: $\Delta$SNR. Selection changes recluster and rescore at dimension $r$, matching the 32-dimensional density; residual changes keep selection fixed. A 16-dimensional residual takes 0.80\,ms on H3; \cref{tab:dimred-fused} gives 32--128.}
\label{tab:rank-split}
\footnotesize
\setlength{\tabcolsep}{3.5pt}
\begin{tabular}{@{}lrrrrrrrr@{}}
\toprule
Dimension & \multicolumn{2}{c}{H3} & \multicolumn{2}{c}{Wan} & \multicolumn{2}{c}{Hun.} & \multicolumn{2}{c}{LTX}\\
\cmidrule(lr){2-3}\cmidrule(lr){4-5}\cmidrule(lr){6-7}\cmidrule(lr){8-9}
 & 1\% & 3.5\% & 1\% & 3.5\% & 1\% & 3.5\% & 1\% & 3.5\%\\
\midrule
32 (both) & 27.7 & 36.2 & 26.4 & 33.5 & 24.7 & 31.0 & 30.0 & 35.0\\
\midrule
\multicolumn{9}{@{}l}{\emph{Selection dimension $r$ (residual 32)}}\\
8 & $-$6.05 & $-$5.38 & $-$10.81 & $-$9.70 & $-$6.70 & $-$5.91 & $-$4.40 & $-$4.27\\
16 & $-$0.57 & $-$0.88 & $-$3.74 & $-$2.62 & $-$1.55 & $-$1.03 & +0.07 & +0.35\\
24 & $-$0.44 & +0.55 & $-$0.61 & $-$0.22 & $-$0.27 & +0.02 & +0.20 & +0.29\\
64 & +0.13 & $-$1.09 & +0.34 & +0.05 & $-$0.08 & $-$0.15 & $-$0.21 & $-$0.15\\
128 & +0.73 & +0.52 & +0.16 & $-$0.16 & $-$0.34 & $-$0.42 & $-$0.38 & $-$0.35\\
\midrule
\multicolumn{9}{@{}l}{\emph{Residual dimension $r$ (selection 32)}}\\
8 & $-$0.65 & $-$0.40 & $-$1.77 & $-$1.01 & $-$1.21 & $-$0.41 & $-$5.29 & $-$5.19\\
16 & $-$0.31 & $-$0.22 & $-$0.74 & $-$0.37 & $-$0.48 & $-$0.21 & $-$1.34 & $-$1.26\\
128 & +0.26 & +0.27 & +0.77 & +0.63 & +0.62 & +0.47 & +0.55 & +0.49\\
\bottomrule
\end{tabular}
\end{table*}

%% file: tables/selection_overlap.tex
\begin{table*}[t]
\centering
\caption{\textbf{Selection overlap within query clusters.} Token-weighted Jaccard, excluding context keys. Compare our low-rank space (ours), the original space (orig.), contiguous tiles, and random sets, with equal mean query counts. Values average captured inputs.}
\label{tab:overlap}
\footnotesize
\setlength{\tabcolsep}{4pt}
\begin{tabular}{lrrrrrrrrrrrrrrrr}
\toprule
 & \multicolumn{4}{c}{H3} & \multicolumn{4}{c}{Wan2.1-14B} & \multicolumn{4}{c}{HunyuanVideo} & \multicolumn{4}{c}{LTX-2.3}\\
\cmidrule(lr){2-5}\cmidrule(lr){6-9}\cmidrule(lr){10-13}\cmidrule(lr){14-17}
Budget (\%) & ours & orig. & tiles & rand. & ours & orig. & tiles & rand. & ours & orig. & tiles & rand. & ours & orig. & tiles & rand.\\
\midrule
1.0 & 0.24 & 0.23 & 0.14 & 0.04 & 0.36 & 0.35 & 0.14 & 0.08 & 0.35 & 0.32 & 0.12 & 0.05 & 0.36 & 0.34 & 0.13 & 0.03\\
3.5 & 0.30 & 0.29 & 0.18 & 0.05 & 0.40 & 0.39 & 0.17 & 0.06 & 0.42 & 0.38 & 0.16 & 0.05 & 0.45 & 0.44 & 0.20 & 0.06\\
\bottomrule
\end{tabular}
\end{table*}

%% file: tables/union_inflation.tex
\begin{table*}[t]
\centering
\caption{\textbf{Union size and accuracy.} Means over captured inputs. $\rho$: executed density (\%); infl.: ratio to each query\textquotesingle s own prefix; tiles: density under contiguous query tiles of the same size. $\Delta$SNR: gain from executing the union rather than the own prefix, both with a per-query residual.}
\label{tab:union}
\footnotesize
\setlength{\tabcolsep}{4pt}
\begin{tabular}{lrrrrrrrrrrrrrrrr}
\toprule
 & \multicolumn{4}{c}{H3} & \multicolumn{4}{c}{Wan2.1-14B} & \multicolumn{4}{c}{HunyuanVideo} & \multicolumn{4}{c}{LTX-2.3}\\
\cmidrule(lr){2-5}\cmidrule(lr){6-9}\cmidrule(lr){10-13}\cmidrule(lr){14-17}
Budget (\%) & $\rho$ & infl. & tiles & $\Delta$SNR & $\rho$ & infl. & tiles & $\Delta$SNR & $\rho$ & infl. & tiles & $\Delta$SNR & $\rho$ & infl. & tiles & $\Delta$SNR\\
\midrule
0.5 & 7.5 & 13.9$\times$ & 10.8 & +8.0 & 4.3 & 8.2$\times$ & 12.6 & +6.3 & 4.0 & 7.6$\times$ & 12.5 & +5.7 & 4.0 & 7.7$\times$ & 11.4 & +5.2\\
1.0 & 11.7 & 11.2$\times$ & 17.2 & +8.7 & 7.6 & 7.4$\times$ & 19.9 & +6.4 & 6.8 & 6.6$\times$ & 20.0 & +5.9 & 6.4 & 6.2$\times$ & 16.9 & +5.2\\
2.0 & 18.1 & 8.9$\times$ & 26.7 & +9.9 & 13.1 & 6.5$\times$ & 30.9 & +6.9 & 11.4 & 5.6$\times$ & 31.1 & +6.5 & 10.2 & 5.0$\times$ & 24.5 & +5.4\\
3.5 & 25.7 & 7.3$\times$ & 37.2 & +11.0 & 20.0 & 5.7$\times$ & 42.7 & +7.6 & 17.2 & 4.9$\times$ & 43.1 & +7.1 & 14.8 & 4.2$\times$ & 32.5 & +5.6\\
5.0 & 31.9 & 6.3$\times$ & 45.3 & +11.3 & 25.9 & 5.2$\times$ & 51.5 & +8.2 & 22.3 & 4.4$\times$ & 52.1 & +7.7 & 18.9 & 3.8$\times$ & 38.6 & +5.9\\
10.0 & 47.2 & 4.7$\times$ & 63.6 & +16.3 & 41.0 & 4.1$\times$ & 70.3 & +10.1 & 36.2 & 3.6$\times$ & 71.0 & +9.2 & 30.0 & 3.0$\times$ & 53.0 & +6.5\\
\bottomrule
\end{tabular}
\end{table*}

%% file: tables/group_size.tex
\begin{table*}[t]
\centering
\caption{\textbf{Query-cluster size.} Vary $C_Q$ at a 3.5\% budget on one capture per generator. Rows: queries per cluster; $\rho$: pair density (\%); SNR: attention accuracy (dB); Sp.: exact-branch speedup over FA3 using FlashInfer variable-block attention.}
\label{tab:group-size}
\footnotesize
\setlength{\tabcolsep}{4pt}
\begin{tabular}{lrrrrrrrrrrrrrrrr}
\toprule
 & \multicolumn{4}{c}{H3} & \multicolumn{4}{c}{Wan2.1-14B} & \multicolumn{4}{c}{HunyuanVideo} & \multicolumn{4}{c}{LTX-2.3}\\
\cmidrule(lr){2-5}\cmidrule(lr){6-9}\cmidrule(lr){10-13}\cmidrule(lr){14-17}
$C_Q$ & rows & $\rho$ & SNR & Sp. & rows & $\rho$ & SNR & Sp. & rows & $\rho$ & SNR & Sp. & rows & $\rho$ & SNR & Sp.\\
\midrule
$\times$0.25 & 513 & 44.5 & 39.4 & 1.64 & 591 & 33.6 & 39.9 & 2.18 & 511 & 27.1 & 37.1 & 2.60 & 510 & 27.7 & 33.8 & 2.50\\
$\times$0.5 & 257 & 35.3 & 36.8 & 1.76 & 295 & 25.8 & 37.5 & 2.52 & 256 & 20.9 & 35.2 & 3.04 & 256 & 21.5 & 32.4 & 2.99\\
deployed & 128 & 27.8 & 34.3 & 1.90 & 148 & 19.7 & 35.4 & 2.79 & 128 & 16.3 & 33.9 & 3.28 & 128 & 17.1 & 31.2 & 3.16\\
$\times$2 & 64 & 21.4 & 32.2 & 1.88 & 74 & 15.0 & 33.8 & 3.04 & 64 & 12.7 & 32.6 & 3.01 & 64 & 13.5 & 30.1 & 2.93\\
$\times$4 & 32 & 16.7 & 30.2 & 1.21 & 37 & 11.6 & 32.2 & 2.01 & 32 & 10.1 & 31.6 & 1.84 & 32 & 10.8 & 29.2 & 1.85\\
\bottomrule
\end{tabular}
\end{table*}

%% file: tables/union_model.tex
\begin{table}[t]
\centering
\caption{\textbf{Predicting union size from cluster statistics.} Measured/predicted inflation, median over query clusters at 1\% and 10\%, and Spearman correlation across query clusters between predicted and measured inflation and pairwise overlap (range over five budgets). Hun.: HunyuanVideo.}
\label{tab:union-model}
\footnotesize
\setlength{\tabcolsep}{2.5pt}
\begin{tabular}{@{}lrrrr@{}}
\toprule
 & \multicolumn{2}{c}{Inflation, meas.\ / pred.} & \multicolumn{2}{c}{Rank corr.}\\
\cmidrule(lr){2-3}\cmidrule(lr){4-5}
 & 1\% & 10\% & Inflation & Overlap\\
\midrule
H3 & 9.8 / 11.7 & 4.5 / 4.6 & 0.96--0.98 & 0.97--0.98\\
Wan & 7.2 / 8.4 & 4.1 / 4.2 & 0.97--0.99 & 0.97--0.99\\
Hun. & 6.4 / 7.6 & 3.5 / 3.6 & 0.95--0.97 & 0.92--0.98\\
LTX & 6.0 / 7.9 & 2.9 / 3.1 & 0.96--0.99 & 0.92--0.98\\
\bottomrule
\end{tabular}
\end{table}

%% file: tables/e1_aggregation.tex
\begin{table*}[t]
\centering
\caption{\textbf{Scores, decisions, and query-cluster budgets.} First row: attention SNR (dB); other rows: $\Delta$SNR from the union (positive = better). All variants use per-query residuals and match total pairs within 1\%. Rows 2--4 retain individual union budgets; later rows redistribute the total. Errors are RMS over sampled query clusters and every variant keeps the deployed union\textquotesingle s token count per cluster, so values differ from \cref{tab:operator-compare}.}
\label{tab:e1-aggregation}
\footnotesize
\setlength{\tabcolsep}{3pt}
\begin{tabular}{@{}lrrrrrrrr@{}}
\toprule
Selection rule & \multicolumn{2}{c}{H3} & \multicolumn{2}{c}{Wan} & \multicolumn{2}{c}{Hun.} & \multicolumn{2}{c}{LTX}\\
\cmidrule(lr){2-3}\cmidrule(lr){4-5}\cmidrule(lr){6-7}\cmidrule(lr){8-9}
 & 1\% & 3.5\% & 1\% & 3.5\% & 1\% & 3.5\% & 1\% & 3.5\%\\
\midrule
Per-query prefixes, cluster union (deployed) & 25.4 & 33.1 & 22.2 & 28.6 & 23.7 & 29.9 & 28.0 & 32.7\\
\midrule
Shared selection, union budget & $-$1.38 & $-$1.47 & $-$1.18 & $-$0.88 & $-$0.57 & $-$0.43 & $-$0.79 & $-$0.72\\
Per-tile decision (summed mass), union budget & +0.01 & +0.15 & $-$0.52 & $-$0.36 & $-$0.08 & +0.03 & $-$0.28 & $-$0.19\\
Per-tile decision (squared mass), union budget & +0.03 & +0.26 & $-$0.33 & $-$0.21 & $-$0.04 & +0.00 & $-$0.24 & $-$0.13\\
\midrule
Union selection, uniform cluster budgets & $-$0.35 & $-$0.93 & $-$0.15 & $-$0.62 & $-$0.33 & $-$0.36 & $-$0.14 & $-$0.26\\
Per-tile decision (summed mass), uniform cluster budgets & $-$0.39 & $-$0.80 & $-$0.75 & $-$0.77 & $-$0.46 & $-$0.32 & $-$0.42 & $-$0.66\\
\midrule
Per-tile decision (summed mass), union budget $\times$ dispersion $\eta$ & $-$0.24 & $-$0.06 & $-$0.62 & $-$0.54 & $-$0.30 & $-$0.44 & $-$0.51 & $-$0.44\\
Per-tile decision (summed mass), global greedy allocation & $-$0.59 & +0.07 & $-$0.94 & $-$0.34 & $-$0.69 & $-$0.26 & $-$0.47 & $-$0.45\\
\bottomrule
\end{tabular}
\end{table*}

%% file: tables/tile_video.tex
\begin{table}[t]
\centering
\caption{\textbf{Per-tile decisions versus the union.} Both use the same exact-attention backend. Each query cluster retains its union budget but selects by summed member mass. H3 videos, 2.5\% budget; $\Delta$PSNR with standard error, $\Delta$SSIM/$\Delta$LPIPS $\times100$.}
\label{tab:tile-video}
\footnotesize
\setlength{\tabcolsep}{3pt}
\begin{tabular}{@{}lrrrrrr@{}}
\toprule
 & Union & Per tile & $\Delta$ (s.e.) & Better & $\Delta$SSIM & $\Delta$LPIPS\\
\midrule
T2V & 28.53 & 28.69 & +0.16 (0.14) & 23/40 & +0.20 & $-$0.20\\
I2V & 31.58 & 31.50 & $-$0.08 (0.14) & 22/40 & $-$0.01 & +0.04\\
\bottomrule
\end{tabular}
\end{table}

%% file: tables/reuse_interval.tex
\begin{table}[t]
\centering
\caption{\textbf{Structure reuse.} Left: H3 video PSNR at 2.5\%, $\Delta$ against $R=4$. Right: kernel speedup at 3.5\%, computed from the refresh and reuse times of the $R=4$ profile. Fresh $\mu_c$ recomputes centroids; reselect also rescores, reselects, and replans. More $b$ uses 3.66/3.50\% on T2V/I2V, matching $R=2$ at 2.5\% in the H3 ablation control.}
\label{tab:reuse}
\footnotesize
\setlength{\tabcolsep}{2.6pt}
\begin{tabular}{@{}lrrrrrrrr@{}}
\toprule
 & \multicolumn{2}{c}{H3 T2V} & \multicolumn{2}{c}{H3 I2V} & \multicolumn{4}{c}{Kernel speedup}\\
\cmidrule(lr){2-3}\cmidrule(lr){4-5}\cmidrule(lr){6-9}
 & PSNR & $\Delta$ & PSNR & $\Delta$ & H3 & Wan & Hun. & LTX\\
\midrule
$R=1$ & 29.04 & +0.42 & 31.92 & +0.27 & 2.00 & 2.42 & 3.04 & 2.42\\
$R=2$ & 28.96 & +0.34 & 31.81 & +0.16 & 2.29 & 2.84 & 3.48 & 3.03\\
$R=4$ & 28.62 & +0.00 & 31.65 & +0.00 & 2.48 & 3.11 & 3.75 & 3.47\\
$R=8$ & 28.07 & $-$0.55 & 30.62 & $-$1.03 & 2.58 & 3.26 & 3.90 & 3.74\\
$R=4$, fresh $\mu_c$ & 28.62 & $-$0.01 & 31.68 & +0.03 & 2.48 & -- & -- & --\\
$R=4$, reselect & 28.79 & +0.17 & 31.84 & +0.19 & 2.15 & 2.63 & 3.35 & 2.72\\
$R=4$, more $b$ & 29.70 & +1.07 & 32.11 & +0.46 & -- & -- & -- & --\\
\bottomrule
\end{tabular}
\end{table}

%% file: figures/budget_alloc_fig.tex
\begin{figure*}[t]
\centering
\includegraphics[width=0.5\textwidth]{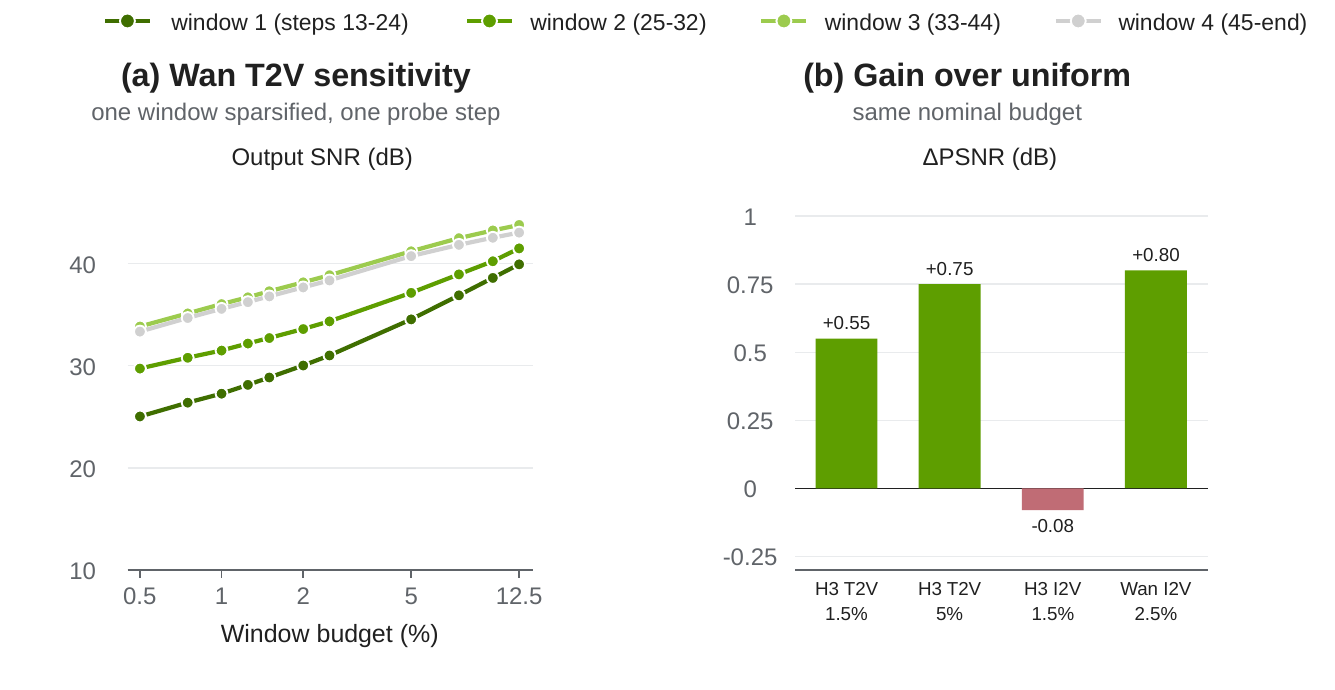}
\caption{\textbf{Denoising-time budget allocation.} (a) Wan T2V denoiser output SNR when one window is sparsified (higher: less sensitive). (b) Video PSNR gain over uniform allocation at the same nominal budget.}
\label{fig:budget-alloc}
\end{figure*}

%% file: tables/sensitivity_test025.tex
\begin{table*}[t]
\centering
\caption{\textbf{Calibration-input sensitivity.} Distinct window budgets from 12 calibrations per task at 2.5\%, their frequencies, and video-fidelity changes relative to the reference schedule. Window budgets are normalized to the target at deployment.}
\label{tab:sensitivity_test025}
\small
\setlength{\tabcolsep}{5pt}
\begin{tabular}{llrrrr}
\toprule
Model/task & Window allocation (\%) & Frequency & $\Delta$PSNR [95\% CI] & $\Delta$SSIM & $\Delta$LPIPS\\
\midrule
\rowcolor{oursrow}
H3 T2V & 5/2/0.75/1.5 (reference) & 5/12 & -- & -- & --\\
H3 T2V & 2.5/2.5/2.5/2.5 & 1/12 & -0.76 [-1.10, -0.44] & -0.0089 & +0.0066\\
H3 T2V & 5/2/1/0.75 & 4/12 & +0.17 [-0.01, +0.41] & -0.0008 & +0.0006\\
H3 T2V & 5/2.5/0.75/0.5 & 1/12 & +0.15 [-0.01, +0.31] & -0.0015 & +0.0012\\
H3 T2V & 5/1.25/1.25/1.5 & 1/12 & -0.03 [-0.08, +0.03] & -0.0003 & +0.0001\\
\rowcolor{oursrow}
H3 I2V & 2.5/1.5/2/5 (reference) & 1/12 & -- & -- & --\\
H3 I2V & 2.5/2.5/2.5/2.5 & 9/12 & -0.18 [-0.38, +0.02] & -0.0035 & +0.0017\\
H3 I2V & 5/2.5/0.5/1.25 & 1/12 & -0.32 [-0.80, +0.17] & -0.0050 & +0.0033\\
H3 I2V & 5/1.25/1.25/1.5 & 1/12 & -0.10 [-0.56, +0.36] & -0.0031 & +0.0021\\
\rowcolor{oursrow}
Wan T2V & 5/2/1/1 (reference) & 7/12 & -- & -- & --\\
Wan T2V & 5/1.5/1.25/1.25 & 5/12 & +0.01 [-0.08, +0.11] & -0.0003 & +0.0002\\
\rowcolor{oursrow}
Wan I2V & 5/2.5/0.75/1 (reference) & 1/12 & -- & -- & --\\
Wan I2V & 2.5/2.5/2.5/2.5 & 6/12 & -0.80 [-1.01, -0.61] & -0.0068 & +0.0041\\
Wan I2V & 5/1.5/1.25/1.25 & 2/12 & -0.08 [-0.28, +0.11] & +0.0006 & -0.0001\\
Wan I2V & 5/2/1/1 & 3/12 & -0.14 [-0.34, +0.04] & -0.0013 & +0.0007\\
\bottomrule
\end{tabular}
\end{table*}

%% file: figures/qual_frames.tex
\begin{figure*}[p]
\centering
\setlength{\tabcolsep}{1pt}\renewcommand{\arraystretch}{0.6}
\begin{tabular}{@{}c@{\hspace{3pt}}cccc@{}}
\multicolumn{5}{@{}l}{\footnotesize (a) H3 T2V: ``a bottle on the left of a wine glass, front view'' \quad video PSNR 36.9\,dB}\\[2pt]
\rotatebox{90}{\scriptsize\hspace{4pt}Dense} & \includegraphics[width=0.212\textwidth]{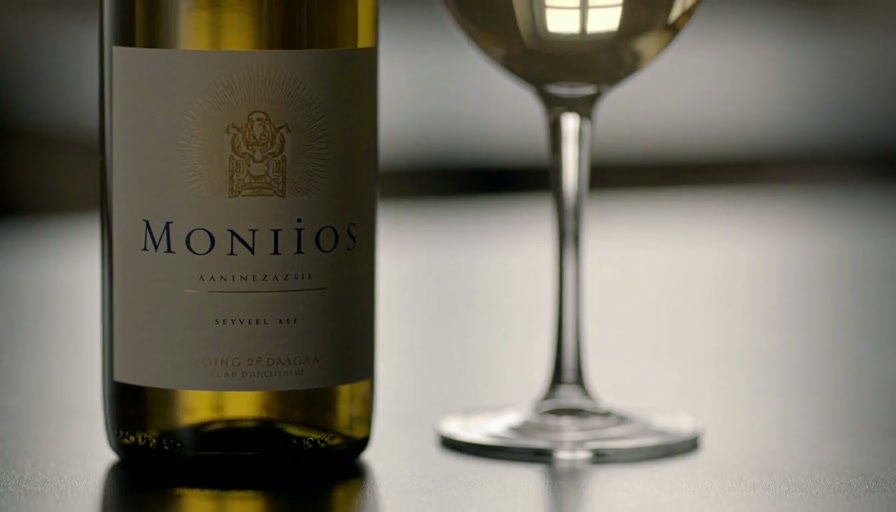} & \includegraphics[width=0.212\textwidth]{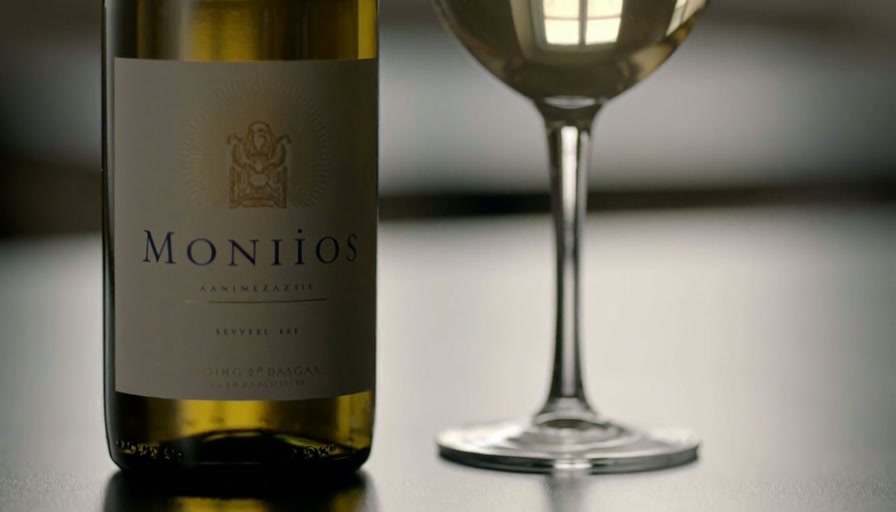} & \includegraphics[width=0.212\textwidth]{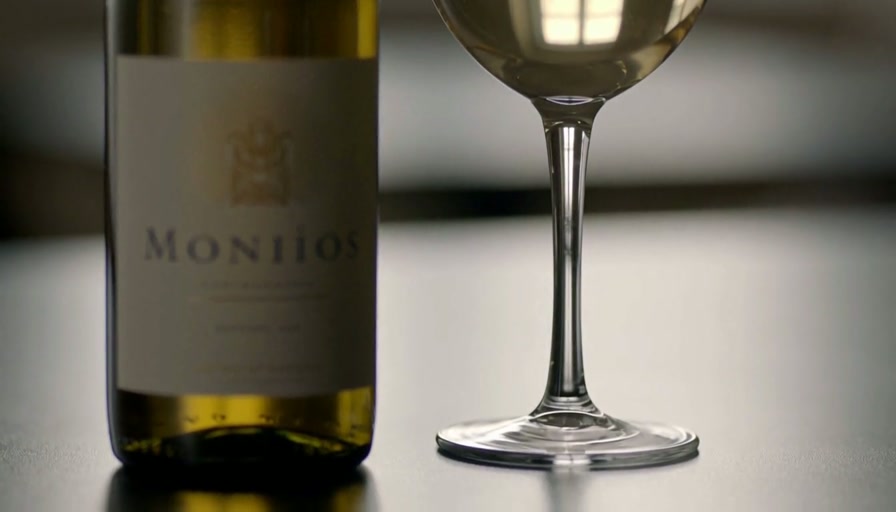} & \includegraphics[width=0.212\textwidth]{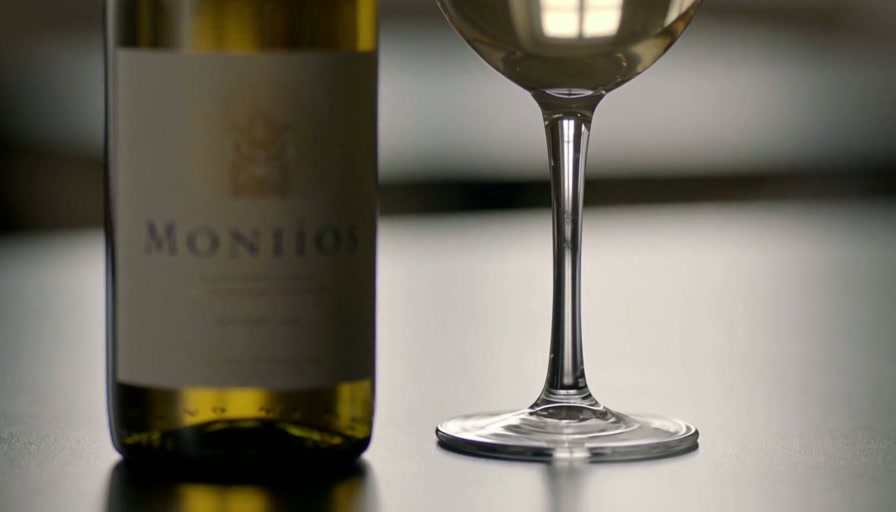}\\
\rotatebox{90}{\scriptsize\hspace{4pt}\method{} 1.5\%} & \includegraphics[width=0.212\textwidth]{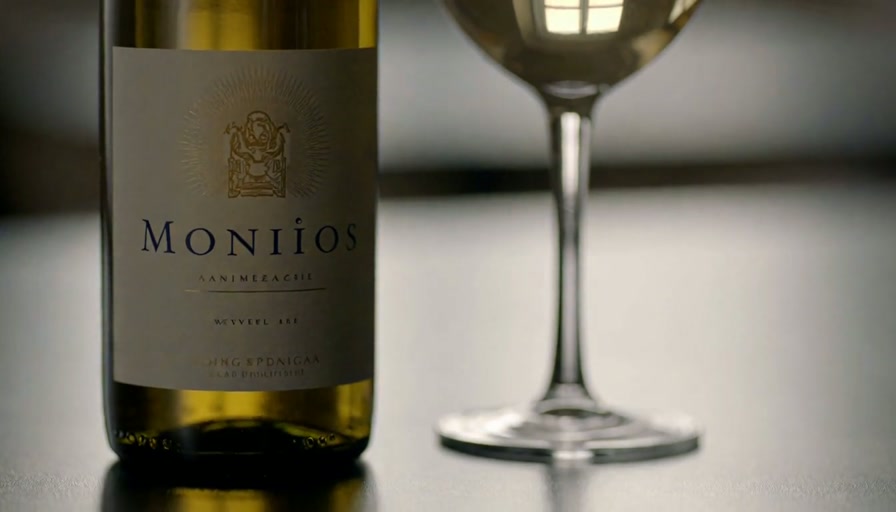} & \includegraphics[width=0.212\textwidth]{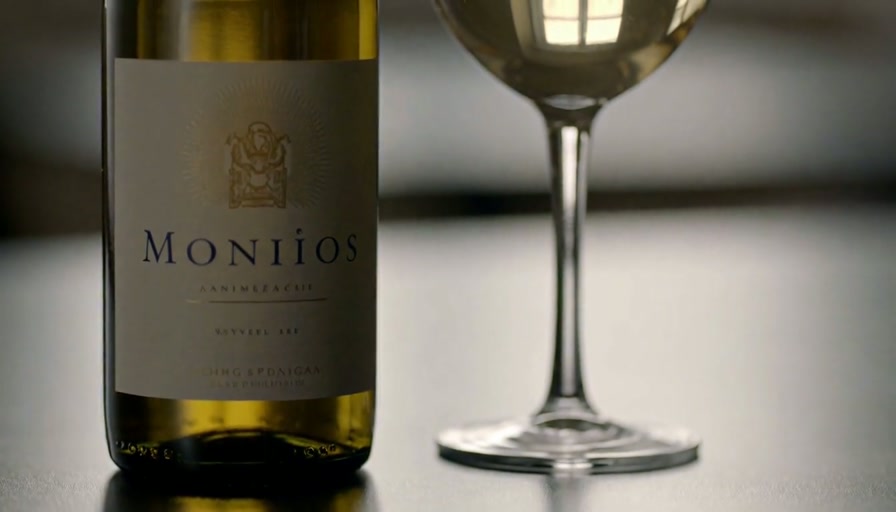} & \includegraphics[width=0.212\textwidth]{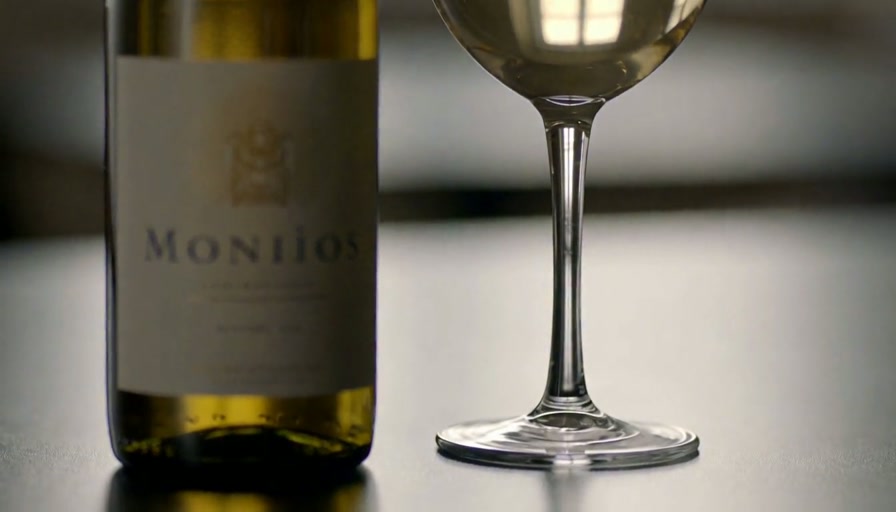} & \includegraphics[width=0.212\textwidth]{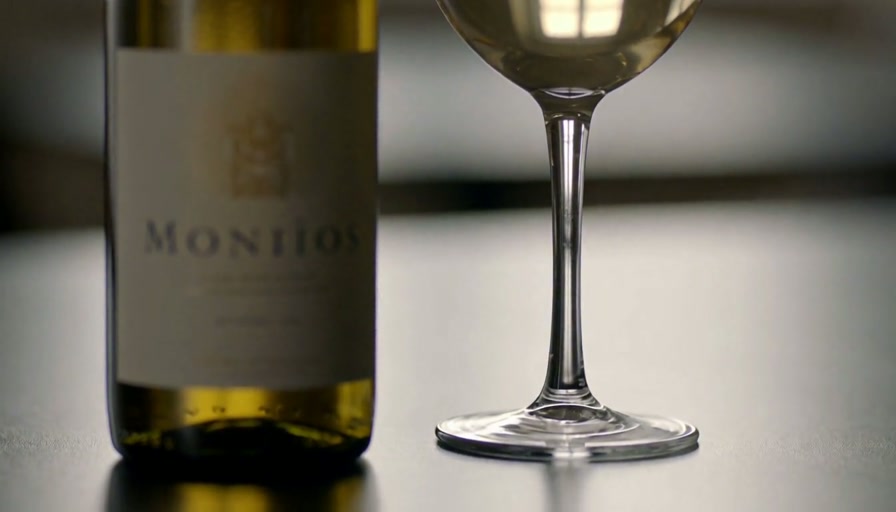}\\
\noalign{\vspace{5pt}}
\multicolumn{5}{@{}l}{\footnotesize (b) H3 T2V: ``Gwen Stacy reading a book by Hokusai, in the style of Ukiyo'' \quad video PSNR 25.8\,dB}\\[2pt]
\rotatebox{90}{\scriptsize\hspace{4pt}Dense} & \includegraphics[width=0.212\textwidth]{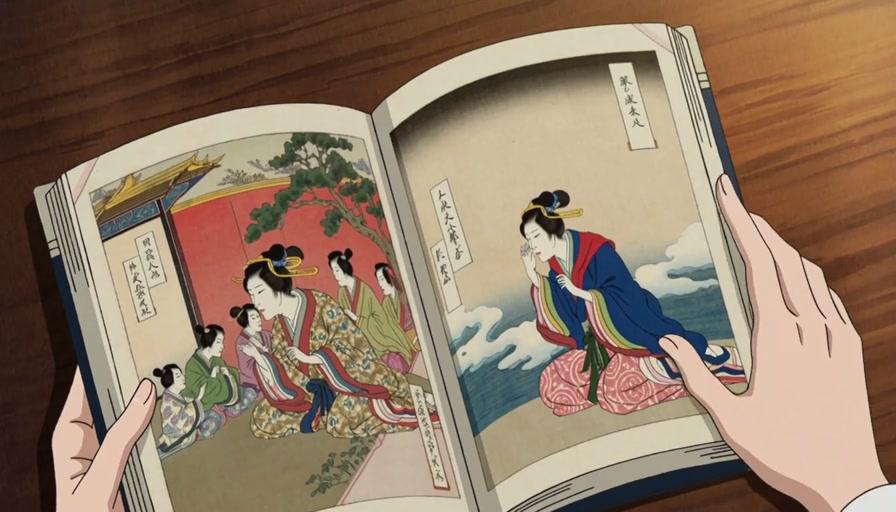} & \includegraphics[width=0.212\textwidth]{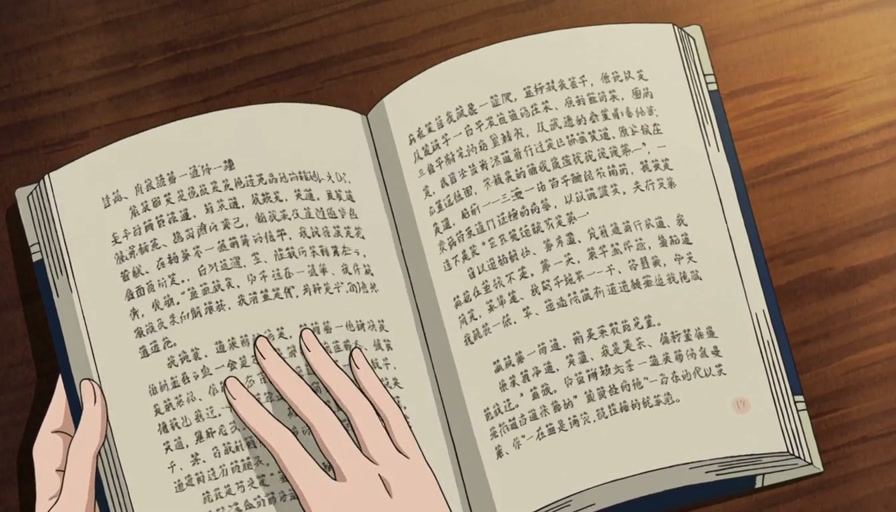} & \includegraphics[width=0.212\textwidth]{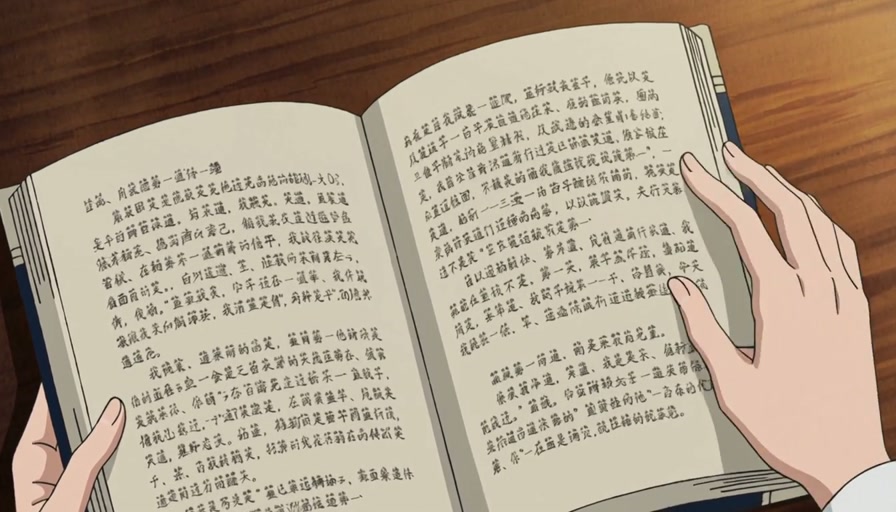} & \includegraphics[width=0.212\textwidth]{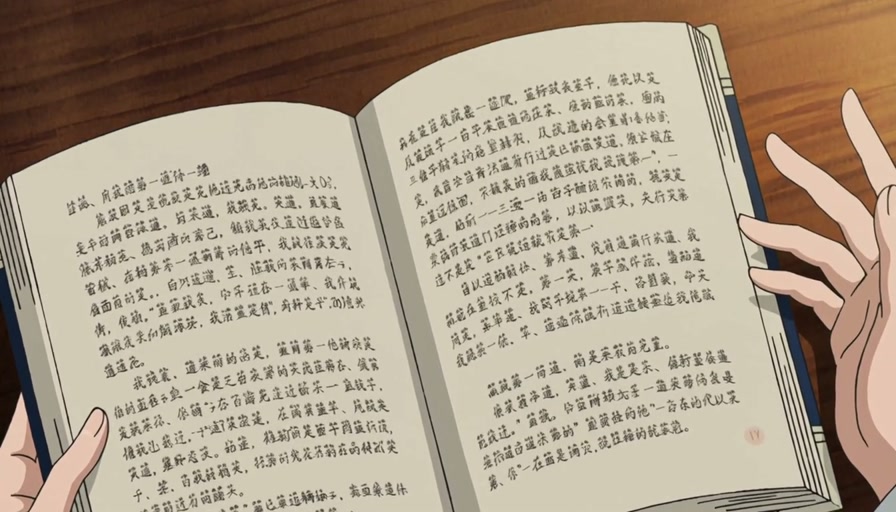}\\
\rotatebox{90}{\scriptsize\hspace{4pt}\method{} 1.5\%} & \includegraphics[width=0.212\textwidth]{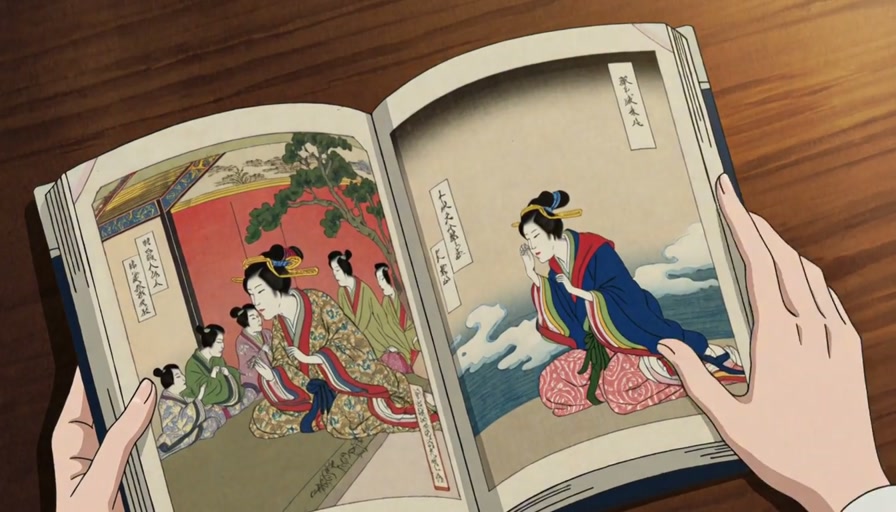} & \includegraphics[width=0.212\textwidth]{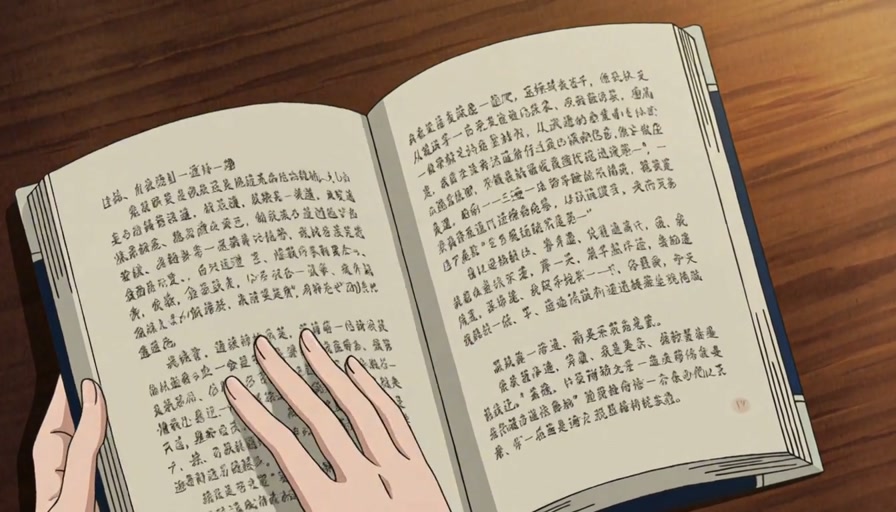} & \includegraphics[width=0.212\textwidth]{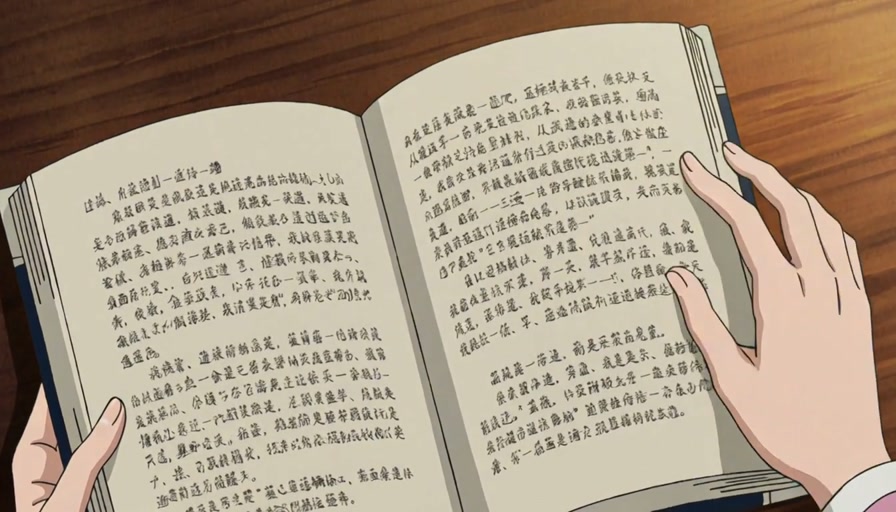} & \includegraphics[width=0.212\textwidth]{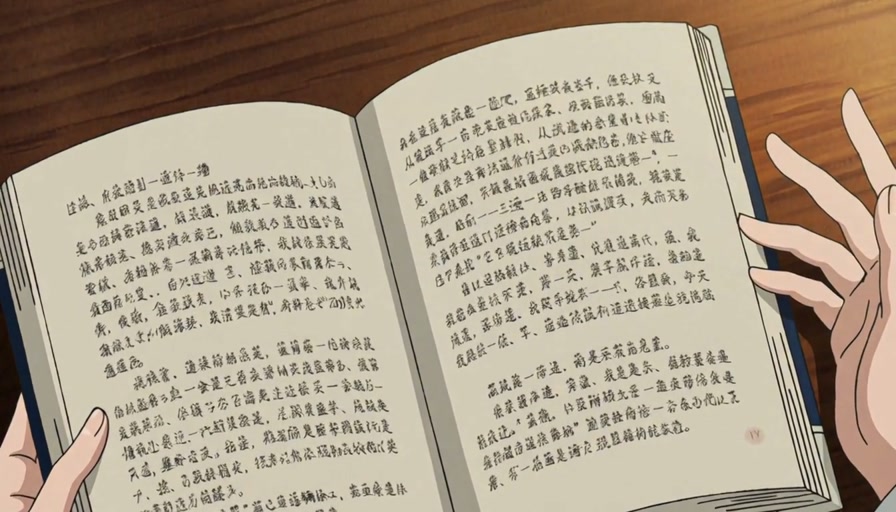}\\
\noalign{\vspace{5pt}}
\multicolumn{5}{@{}l}{\footnotesize (c) H3 I2V: ``an older woman and a young girl are knitting together'' \quad video PSNR 24.5\,dB}\\[2pt]
\rotatebox{90}{\scriptsize\hspace{4pt}Dense} & \includegraphics[width=0.212\textwidth]{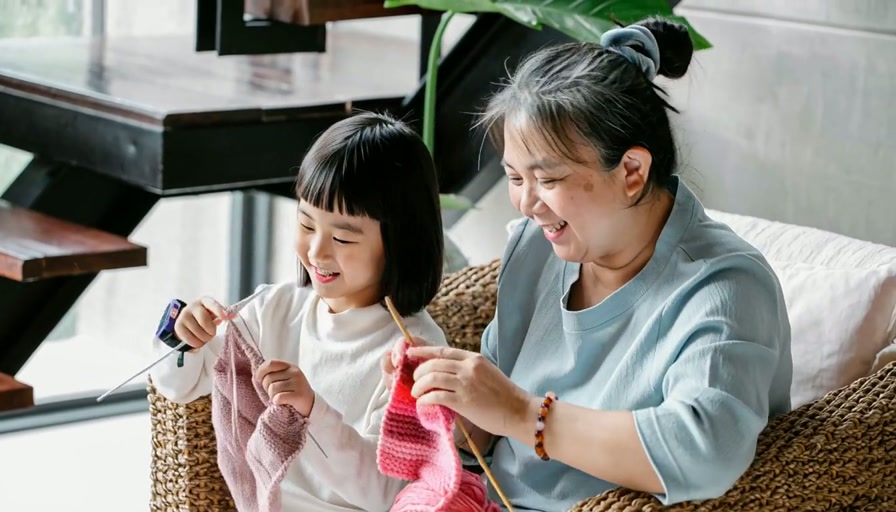} & \includegraphics[width=0.212\textwidth]{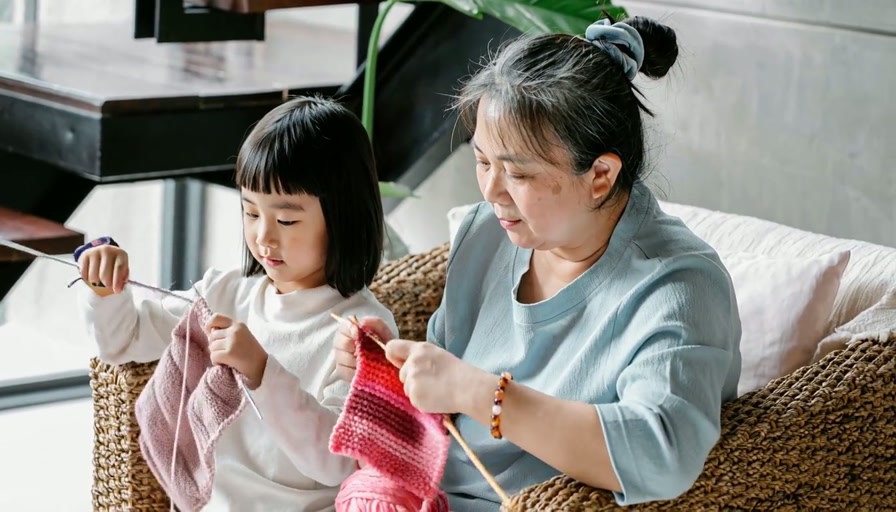} & \includegraphics[width=0.212\textwidth]{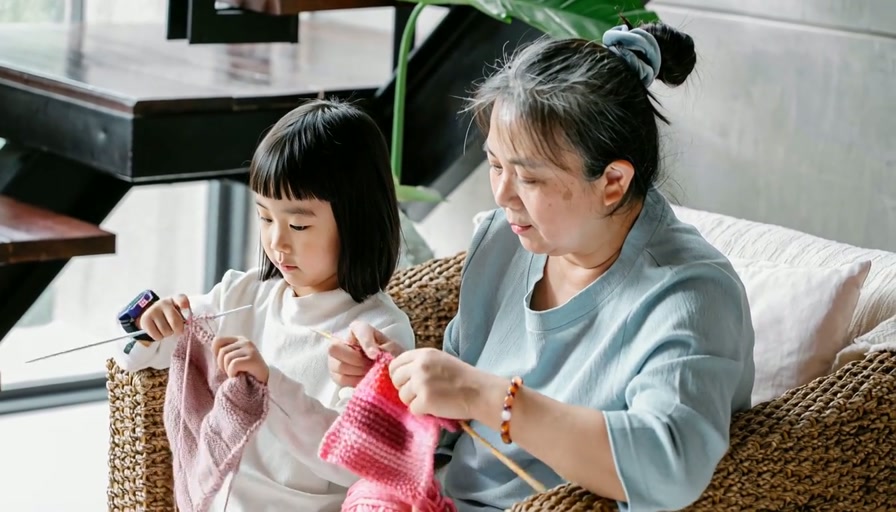} & \includegraphics[width=0.212\textwidth]{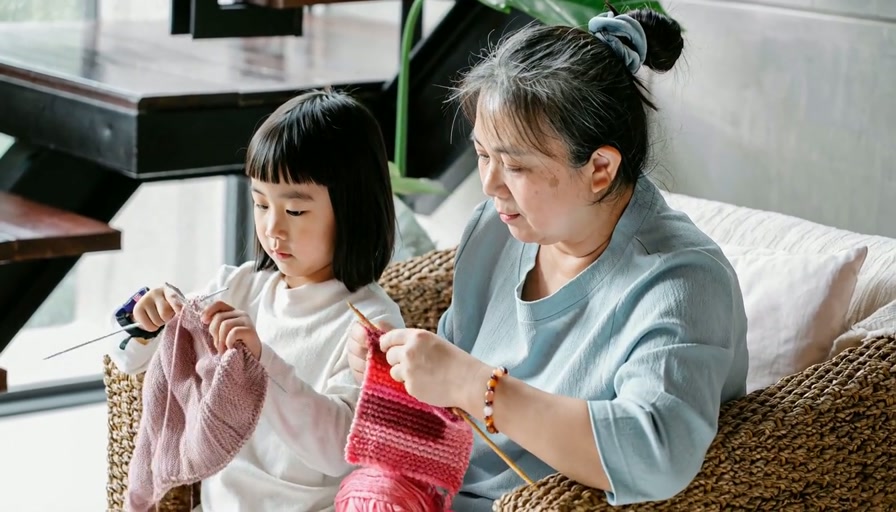}\\
\rotatebox{90}{\scriptsize\hspace{4pt}\method{} 1.25\%} & \includegraphics[width=0.212\textwidth]{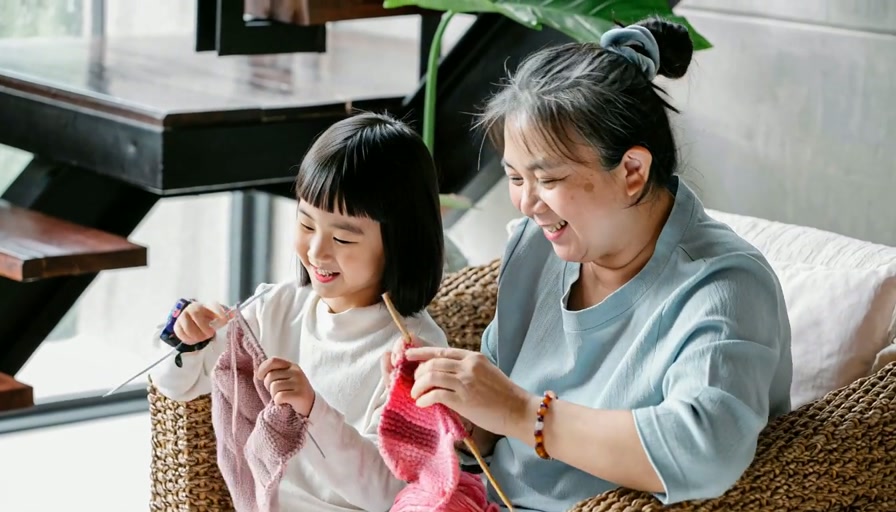} & \includegraphics[width=0.212\textwidth]{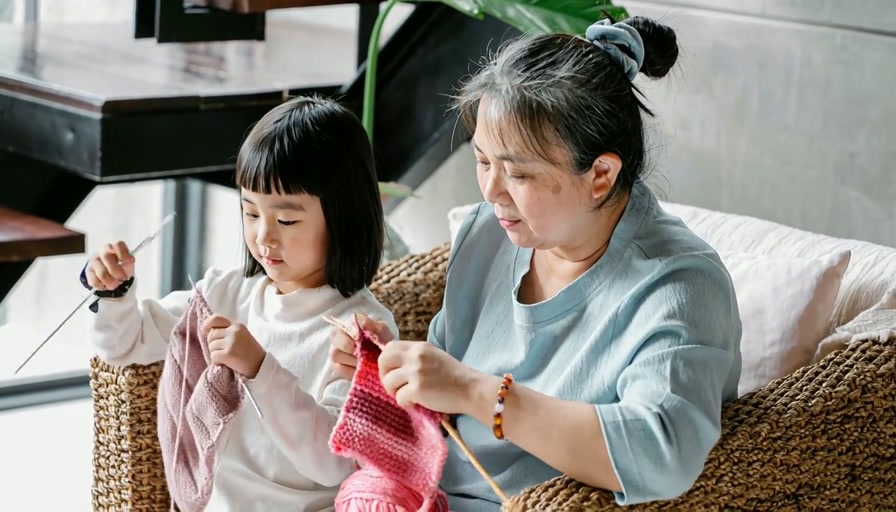} & \includegraphics[width=0.212\textwidth]{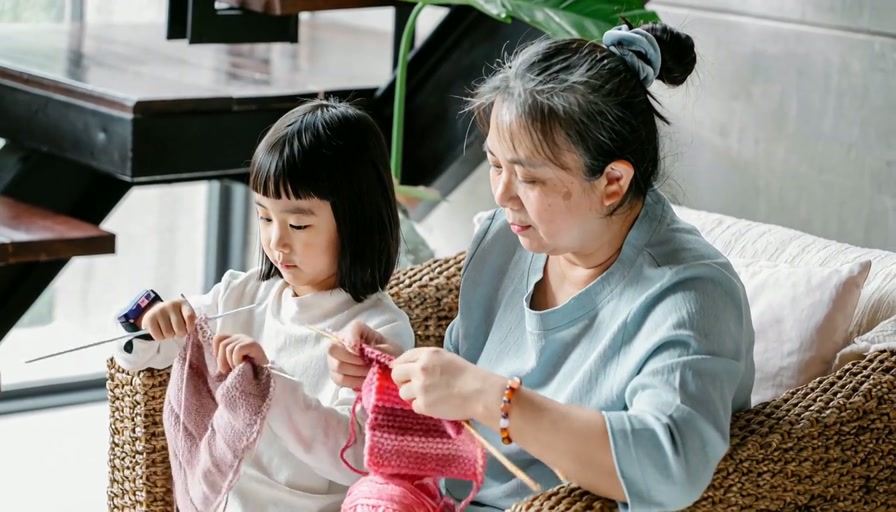} & \includegraphics[width=0.212\textwidth]{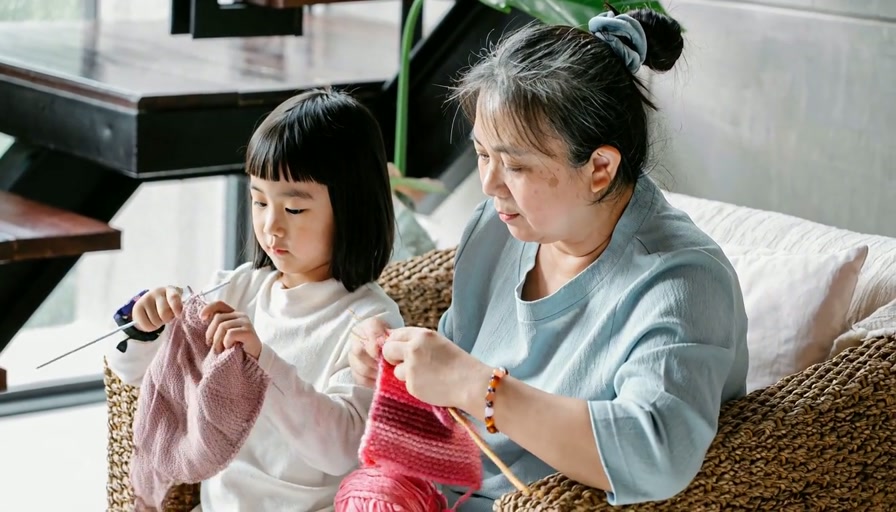}\\
\noalign{\vspace{5pt}}
\multicolumn{5}{@{}l}{\footnotesize (d) H3 I2V: ``a woman with afro hair is smiling while wearing earphones'' \quad video PSNR 30.6\,dB}\\[2pt]
\rotatebox{90}{\scriptsize\hspace{4pt}Dense} & \includegraphics[width=0.212\textwidth]{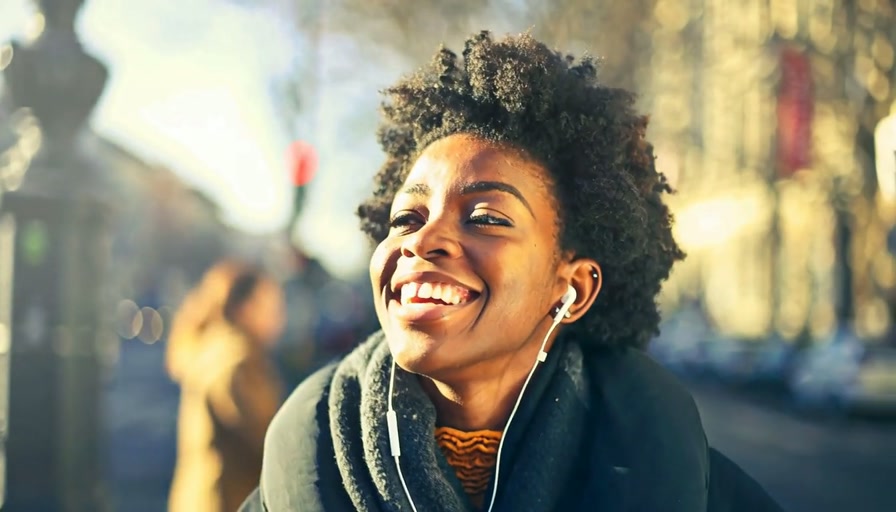} & \includegraphics[width=0.212\textwidth]{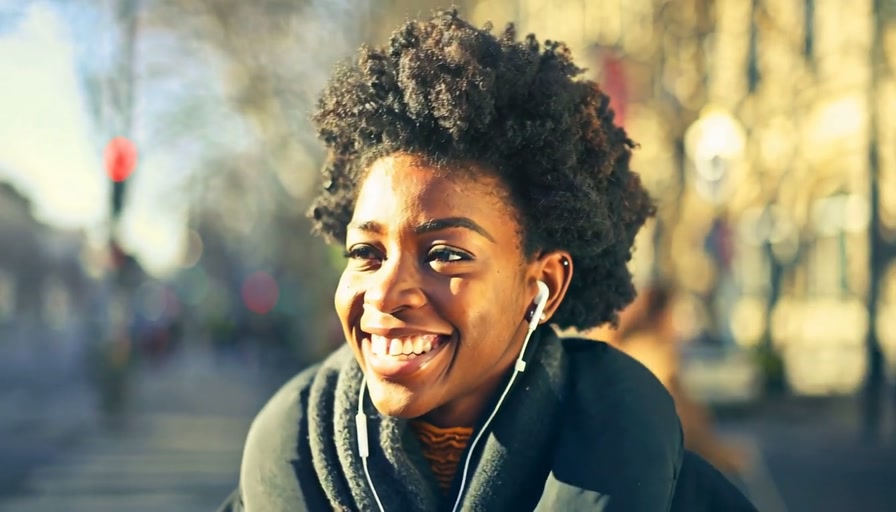} & \includegraphics[width=0.212\textwidth]{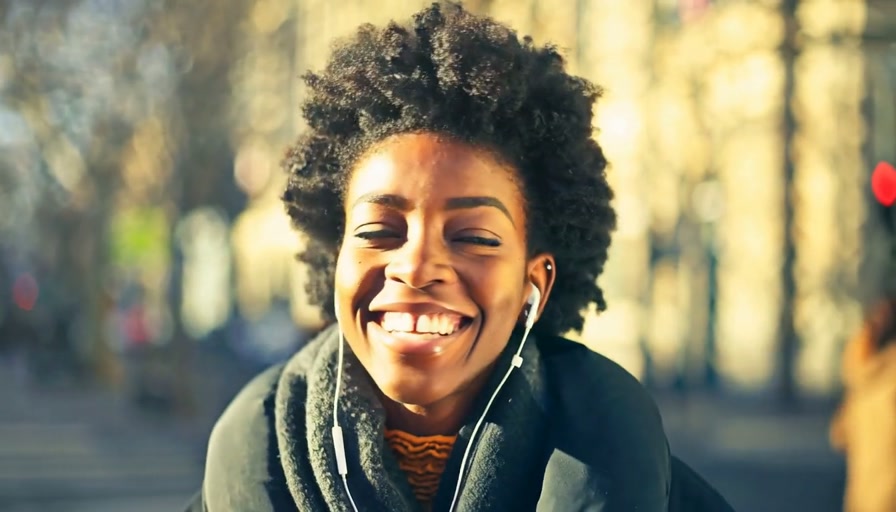} & \includegraphics[width=0.212\textwidth]{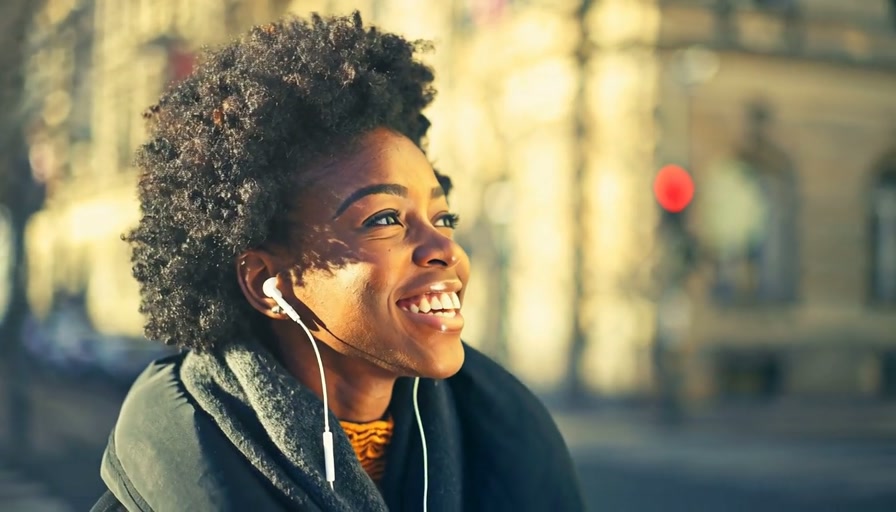}\\
\rotatebox{90}{\scriptsize\hspace{4pt}\method{} 1.25\%} & \includegraphics[width=0.212\textwidth]{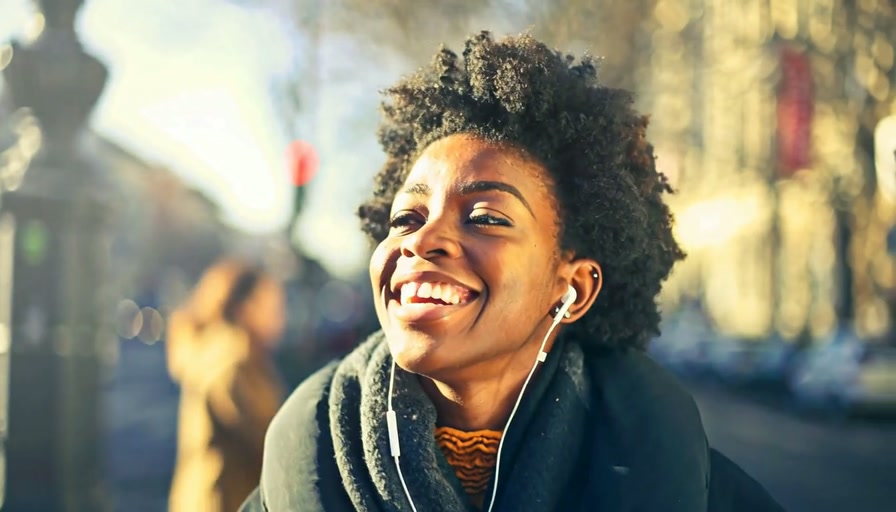} & \includegraphics[width=0.212\textwidth]{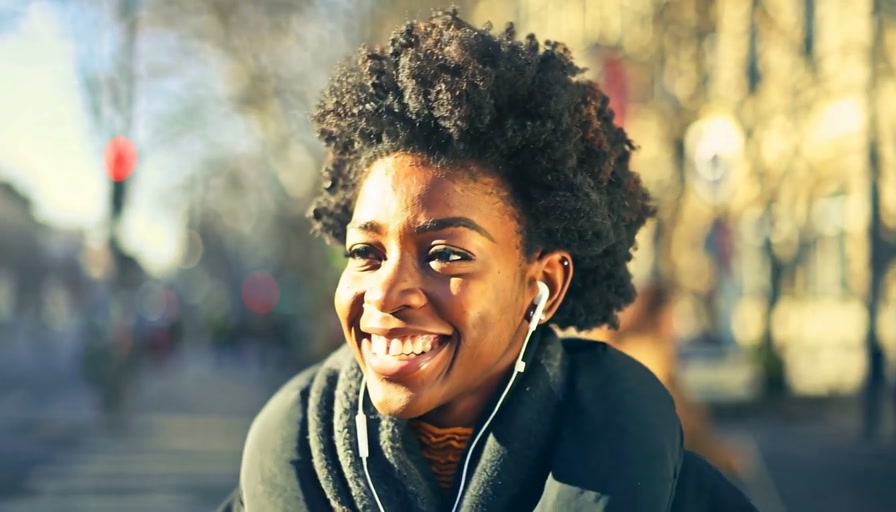} & \includegraphics[width=0.212\textwidth]{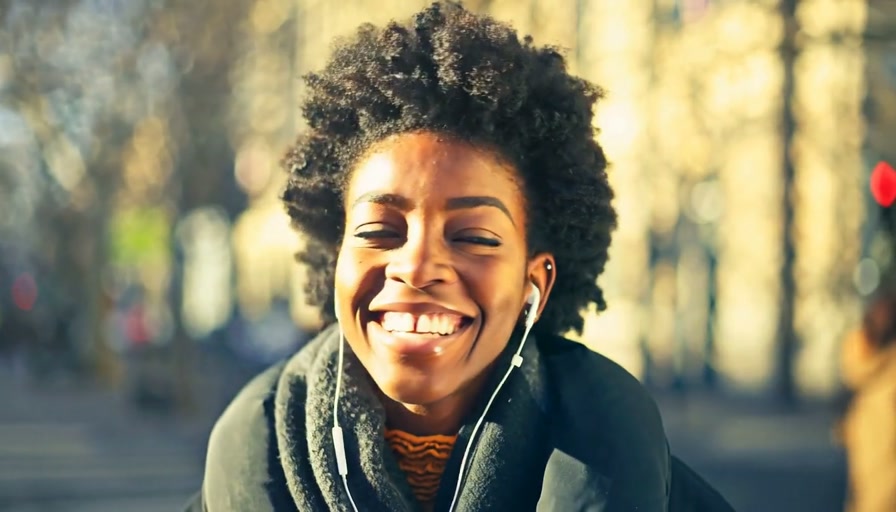} & \includegraphics[width=0.212\textwidth]{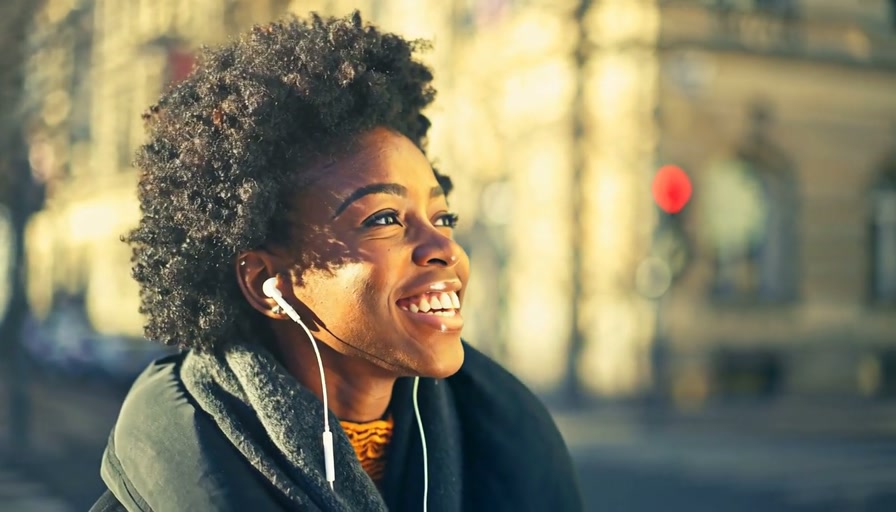}\\
 & \scriptsize frame 36 (1.5\,s) & \scriptsize frame 97 (4.0\,s) & \scriptsize frame 157 (6.5\,s) & \scriptsize frame 218 (9.1\,s)\\
\end{tabular}
\caption{\textbf{Dense and \method frames at matched timestamps.} Cases show lettering (a, b), faces, and hands (c, d). Rows use the main-table settings; video PSNR is relative to dense.}
\label{fig:qual-frames}
\end{figure*}

%% file: main.bbl
\begin{thebibliography}{49}
\providecommand{\natexlab}[1]{#1}
\providecommand{\url}[1]{\texttt{#1}}
\expandafter\ifx\csname urlstyle\endcsname\relax
  \providecommand{\doi}[1]{doi: #1}\else
  \providecommand{\doi}{doi: \begingroup \urlstyle{rm}\Url}\fi

\bibitem[Chen et~al.(2024)Chen, Sadhukhan, Ye, Zhou, Zhang, Nolte, Tian, Douze,
  Bottou, Jia, and Chen]{magicpig}
Zhuoming Chen, Ranajoy Sadhukhan, Zihao Ye, Yang Zhou, Jianyu Zhang, Niklas
  Nolte, Yuandong Tian, Matthijs Douze, Leon Bottou, Zhihao Jia, and Beidi
  Chen.
\newblock {MagicPIG}: {LSH} sampling for efficient {LLM} generation.
\newblock \emph{arXiv preprint arXiv:2410.16179}, 2024.

\bibitem[Dao(2023)]{flashattention2}
Tri Dao.
\newblock {FlashAttention-2: Faster Attention with Better Parallelism and Work
  Partitioning}.
\newblock \emph{arXiv preprint arXiv:2307.08691}, 2023.

\bibitem[Dao et~al.(2022)Dao, Fu, Ermon, Rudra, and Ré]{flashattention}
Tri Dao, Daniel~Y. Fu, Stefano Ermon, Atri Rudra, and Christopher Ré.
\newblock {FlashAttention: Fast and Memory-Efficient Exact Attention with
  IO-Awareness}.
\newblock In \emph{Advances in Neural Information Processing Systems}, 2022.

\bibitem[Ding et~al.(2026)Ding, Li, Ma, Zhou, Li, Zhou, Cao, Li, Geng, Gu, Lin,
  Xie, and Shen]{draftattention2}
Rui Ding, Haopeng Li, Weize Ma, Yufa Zhou, Yitong Li, Xiaoling Zhou, Jiashuo
  Cao, Liyang Li, Hua Geng, Jiuxiang Gu, Jun Lin, Enze Xie, and Xuan Shen.
\newblock {DraftAttention2: Fast Video Diffusion with Low-Resolution-Guided
  Mixed-Precision Attention}.
\newblock \emph{arXiv preprint arXiv:2609.32628}, 2026.

\bibitem[Eckart and Young(1936)]{eckartyoung}
Carl Eckart and Gale Young.
\newblock {The Approximation of One Matrix by Another of Lower Rank}.
\newblock \emph{Psychometrika}, 1\penalty0 (3):\penalty0 211--218, 1936.

\bibitem[HaCohen et~al.(2024)HaCohen, Chiprut, Brazowski, Shalem, Moshe,
  Richardson, Levin, Shiran, Zabari, Gordon, Panet, Weissbuch, Kulikov,
  Bitterman, Melumian, and Bibi]{ltxvideo}
Yoav HaCohen, Nisan Chiprut, Benny Brazowski, Daniel Shalem, Dudu Moshe, Eitan
  Richardson, Eran Levin, Guy Shiran, Nir Zabari, Ori Gordon, Poriya Panet,
  Sapir Weissbuch, Victor Kulikov, Yaki Bitterman, Zeev Melumian, and Ofir
  Bibi.
\newblock {LTX-Video}: Realtime video latent diffusion.
\newblock \emph{arXiv preprint arXiv:2501.00103}, 2024.

\bibitem[Huang et~al.(2024)Huang, He, Yu, Zhang, Si, Jiang, Zhang, Wu, Jin,
  Chanpaisit, Wang, Chen, Wang, Lin, Qiao, and Liu]{vbench}
Ziqi Huang, Yinan He, Jiashuo Yu, Fan Zhang, Chenyang Si, Yuming Jiang, Yuanhan
  Zhang, Tianxing Wu, Qingyang Jin, Nattapol Chanpaisit, Yaohui Wang, Xinyuan
  Chen, Limin Wang, Dahua Lin, Yu Qiao, and Ziwei Liu.
\newblock {VBench: Comprehensive Benchmark Suite for Video Generative Models}.
\newblock In \emph{IEEE/CVF Conference on Computer Vision and Pattern
  Recognition}, 2024.

\bibitem[Huang et~al.(2026)Huang, Zhang, Xu, He, Yu, Dong, Ma, Chanpaisit, Si,
  Jiang, Wang, Chen, Chen, Wang, Lin, Qiao, and Liu]{vbenchpp}
Ziqi Huang, Fan Zhang, Xiaojie Xu, Yinan He, Jiashuo Yu, Ziyue Dong, Qianli Ma,
  Nattapol Chanpaisit, Chenyang Si, Yuming Jiang, Yaohui Wang, Xinyuan Chen,
  Ying-Cong Chen, Limin Wang, Dahua Lin, Yu Qiao, and Ziwei Liu.
\newblock {VBench++: Comprehensive and Versatile Benchmark Suite for Video
  Generative Models}.
\newblock \emph{IEEE Transactions on Pattern Analysis and Machine
  Intelligence}, 2026.

\bibitem[Jacobs et~al.(2023)Jacobs, Tanaka, Zhang, Zhang, Song, Rajbhandari,
  and He]{ulysses}
Sam~Ade Jacobs, Masahiro Tanaka, Chengming Zhang, Minjia Zhang, Shuaiwen~Leon
  Song, Samyam Rajbhandari, and Yuxiong He.
\newblock {DeepSpeed Ulysses: System Optimizations for Enabling Training of
  Extreme Long Sequence Transformer Models}.
\newblock \emph{arXiv preprint arXiv:2309.14509}, 2023.

\bibitem[Kitaev et~al.(2020)Kitaev, Kaiser, and Levskaya]{reformer}
Nikita Kitaev, {\L}ukasz Kaiser, and Anselm Levskaya.
\newblock Reformer: The efficient transformer.
\newblock In \emph{International Conference on Learning Representations}, 2020.

\bibitem[Kong et~al.(2024)Kong, Tian, Zhang, Min, Dai, Zhou, Xiong, Li, Wu,
  Zhang, Wu, Lin, Yuan, Long, Wang, Wang, Li, Huang, Yang, Tan, Wang, Song,
  Bai, Wu, Xue, Wang, Wang, Liu, Li, Li, Wang, Yu, Deng, Li, Chen, Cui, Peng,
  Yu, He, Xu, Zhou, Xu, Tao, Lu, Liu, Zhou, Wang, Yang, Wang, Liu, Jiang, and
  Zhong]{hunyuanvideo}
Weijie Kong, Qi Tian, Zijian Zhang, Rox Min, Zuozhuo Dai, Jin Zhou, Jiangfeng
  Xiong, Xin Li, Bo Wu, Jianwei Zhang, Kathrina Wu, Qin Lin, Junkun Yuan,
  Yanxin Long, Aladdin Wang, Andong Wang, Changlin Li, Duojun Huang, Fang Yang,
  Hao Tan, Hongmei Wang, Jacob Song, Jiawang Bai, Jianbing Wu, Jinbao Xue, Joey
  Wang, Kai Wang, Mengyang Liu, Pengyu Li, Shuai Li, Weiyan Wang, Wenqing Yu,
  Xinchi Deng, Yang Li, Yi Chen, Yutao Cui, Yuanbo Peng, Zhentao Yu, Zhiyu He,
  Zhiyong Xu, Zixiang Zhou, Zunnan Xu, Yangyu Tao, Qinglin Lu, Songtao Liu, Dax
  Zhou, Hongfa Wang, Yong Yang, Di Wang, Yuhong Liu, Jie Jiang, and Caesar
  Zhong.
\newblock {HunyuanVideo: A Systematic Framework For Large Video Generative
  Models}.
\newblock \emph{arXiv preprint arXiv:2412.03603}, 2024.

\bibitem[Li et~al.(2026{\natexlab{a}})Li, Li, Chen, Ye, Liu, Yu, Wang, Zhang,
  Xie, Xie, and Han]{solattn}
Haopeng Li, Yitong Li, Junsong Chen, Tian Ye, Haozhe Liu, Jincheng Yu, Duomin
  Wang, Ruihua Zhang, Zeke Xie, Enze Xie, and Song Han.
\newblock {Sol-Attn: Accelerating Video Generation Inference via On-the-Fly
  Attention Sparsification}.
\newblock \emph{arXiv preprint arXiv:2607.24027}, 2026{\natexlab{a}}.

\bibitem[Li et~al.(2026{\natexlab{b}})Li, Shao, Zhong, Zhou, Bai, Xiong, and
  Xie]{pisa}
Haopeng Li, Shitong Shao, Wenliang Zhong, Zikai Zhou, Lichen Bai, Hui Xiong,
  and Zeke Xie.
\newblock {PISA}: Piecewise sparse attention is wiser for efficient diffusion
  transformers.
\newblock In \emph{Advances in Neural Information Processing Systems},
  2026{\natexlab{b}}.

\bibitem[Li et~al.(2025)Li, Li, Cai, Xi, Yang, Lin, Zhang, Yang, Hu, Peng,
  Agrawala, Stoica, Keutzer, and Han]{radial}
Xingyang Li, Muyang Li, Tianle Cai, Haocheng Xi, Shuo Yang, Yujun Lin, Lvmin
  Zhang, Songlin Yang, Jinbo Hu, Kelly Peng, Maneesh Agrawala, Ion Stoica, Kurt
  Keutzer, and Song Han.
\newblock {Radial Attention: $O(n\log n)$ Sparse Attention with Energy Decay
  for Long Video Generation}.
\newblock In \emph{Advances in Neural Information Processing Systems}, 2025.

\bibitem[{LightX2V Team}(2026)]{lightx2v}
{LightX2V Team}.
\newblock {MiniMax-H3 Turbo}.
\newblock Distilled model, Hugging Face, 2026.

\bibitem[Liu et~al.(2024{\natexlab{a}})Liu, Chen, Lu, Jiang, Han, Zhang, Chen,
  Zhang, Ding, Zhang, Chen, Yang, Yang, and Qiu]{retrievalattention}
Di Liu, Meng Chen, Baotong Lu, Huiqiang Jiang, Zhenhua Han, Qianxi Zhang, Qi
  Chen, Chengruidong Zhang, Bailu Ding, Kai Zhang, Chen Chen, Fan Yang, Yuqing
  Yang, and Lili Qiu.
\newblock {RetrievalAttention}: Accelerating long-context {LLM} inference via
  vector retrieval.
\newblock \emph{arXiv preprint arXiv:2409.10516}, 2024{\natexlab{a}}.

\bibitem[Liu et~al.(2024{\natexlab{b}})Liu, Li, Zhao, Zhang, and
  Guo]{clusterkv}
Guangda Liu, Chengwei Li, Jieru Zhao, Chenqi Zhang, and Minyi Guo.
\newblock {ClusterKV}: Manipulating {LLM} {KV} cache in semantic space for
  recallable compression.
\newblock \emph{arXiv preprint arXiv:2412.03213}, 2024{\natexlab{b}}.

\bibitem[Lloyd(1982)]{lloyd}
S. Lloyd.
\newblock {Least squares quantization in PCM}.
\newblock \emph{IEEE Transactions on Information Theory}, 28\penalty0
  (2):\penalty0 129--137, 1982.

\bibitem[Luo et~al.(2026)Luo, Chen, Wang, Wang, Zhu, Sun, Gao, Chen, and
  Li]{svoo}
Jiayi Luo, Jiayu Chen, Jiankun Wang, Cong Wang, Hanxin Zhu, Qingyun Sun, Chen
  Gao, Zhibo Chen, and Jianxin Li.
\newblock {Attention Sparsity is Input-Stable: Training-Free Sparse Attention
  for Video Generation via Offline Sparsity Profiling and Online QK
  Co-Clustering}.
\newblock In \emph{International Conference on Machine Learning}, 2026.

\bibitem[{MiniMax}(2026)]{minimaxh3}
{MiniMax}.
\newblock {MiniMax H3}.
\newblock Model card, Hugging Face, 2026.

\bibitem[{NVIDIA}(2026)]{cudnn}
{NVIDIA}.
\newblock {NVIDIA cuDNN}.
\newblock Library documentation, 2026.

\bibitem[Peebles and Xie(2023)]{dit}
William Peebles and Saining Xie.
\newblock {Scalable Diffusion Models with Transformers}.
\newblock In \emph{IEEE/CVF International Conference on Computer Vision}, 2023.

\bibitem[Ribar et~al.(2024)Ribar, Chelombiev, Hudlass-Galley, Blake, Luschi,
  and Orr]{sparq}
Luka Ribar, Ivan Chelombiev, Luke Hudlass-Galley, Charlie Blake, Carlo Luschi,
  and Douglas Orr.
\newblock {SparQ} attention: Bandwidth-efficient {LLM} inference.
\newblock In \emph{International Conference on Machine Learning}, 2024.

\bibitem[Roy et~al.(2021)Roy, Saffar, Vaswani, and
  Grangier]{routingtransformer}
Aurko Roy, Mohammad Saffar, Ashish Vaswani, and David Grangier.
\newblock Efficient content-based sparse attention with routing transformers.
\newblock \emph{Transactions of the Association for Computational Linguistics},
  9:\penalty0 53--68, 2021.

\bibitem[{SGLang Contributors}(2026{\natexlab{a}})]{sglangdiffusion}
{SGLang Contributors}.
\newblock Attention backends.
\newblock {SGLang} Diffusion documentation, 2026{\natexlab{a}}.

\bibitem[{SGLang Contributors}(2026{\natexlab{b}})]{subblock}
{SGLang Contributors}.
\newblock {MiniMax-H3}: {SubBlock} sparse attention.
\newblock {SGLang} Diffusion cookbook, 2026{\natexlab{b}}.

\bibitem[Shah et~al.(2024)Shah, Bikshandi, Zhang, Thakkar, Ramani, and
  Dao]{flashattention3}
Jay Shah, Ganesh Bikshandi, Ying Zhang, Vijay Thakkar, Pradeep Ramani, and Tri
  Dao.
\newblock {FlashAttention-3: Fast and Accurate Attention with Asynchrony and
  Low-precision}.
\newblock \emph{arXiv preprint arXiv:2407.08608}, 2024.

\bibitem[Shen et~al.(2025)Shen, Han, Zhou, Xie, Gong, Wang, Wang, Wang, Zhao,
  and Gu]{draftattention}
Xuan Shen, Chenxia Han, Yufa Zhou, Yanyue Xie, Yifan Gong, Quanyi Wang, Yiwei
  Wang, Yanzhi Wang, Pu Zhao, and Jiuxiang Gu.
\newblock {DraftAttention: Fast Video Diffusion via Low-Resolution Attention
  Guidance}.
\newblock \emph{arXiv preprint arXiv:2505.14708}, 2025.

\bibitem[Singhania et~al.(2024)Singhania, Singh, He, Feizi, and Bhatele]{loki}
Prajwal Singhania, Siddharth Singh, Shwai He, Soheil Feizi, and Abhinav
  Bhatele.
\newblock {Loki}: Low-rank keys for efficient sparse attention.
\newblock In \emph{Advances in Neural Information Processing Systems}, 2024.

\bibitem[Taghavi et~al.(2026)Taghavi, Langari, and Pandey]{sparsepr}
Pardis Taghavi, Reza Langari, and Gaurav Pandey.
\newblock {Partition the Support, Reconstruct the Residual: Training-Free
  Sparse Attention for Video Generation and World Models}.
\newblock \emph{arXiv preprint arXiv:2608.18484}, 2026.

\bibitem[{Team Wan} et~al.(2025){Team Wan}, Wang, Ai, Wen, Mao, Xie, Chen, Yu,
  Zhao, Yang, Zeng, Wang, Zhang, Zhou, Wang, Chen, Zhu, Zhao, Yan, Huang, Feng,
  Zhang, Li, Wu, Chu, Feng, Zhang, Sun, Fang, Wang, Gui, Weng, Shen, Lin, Wang,
  Wang, Zhou, Wang, Shen, Yu, Shi, Huang, Xu, Kou, Lv, Li, Liu, Wang, Zhang,
  Huang, Li, Wu, Liu, Pan, Zheng, Hong, Shi, Feng, Jiang, Han, Wu, and
  Liu]{wan}
{Team Wan}, Ang Wang, Baole Ai, Bin Wen, Chaojie Mao, Chen-Wei Xie, Di Chen,
  Feiwu Yu, Haiming Zhao, Jianxiao Yang, Jianyuan Zeng, Jiayu Wang, Jingfeng
  Zhang, Jingren Zhou, Jinkai Wang, Jixuan Chen, Kai Zhu, Kang Zhao, Keyu Yan,
  Lianghua Huang, Mengyang Feng, Ningyi Zhang, Pandeng Li, Pingyu Wu, Ruihang
  Chu, Ruili Feng, Shiwei Zhang, Siyang Sun, Tao Fang, Tianxing Wang, Tianyi
  Gui, Tingyu Weng, Tong Shen, Wei Lin, Wei Wang, Wei Wang, Wenmeng Zhou, Wente
  Wang, Wenting Shen, Wenyuan Yu, Xianzhong Shi, Xiaoming Huang, Xin Xu, Yan
  Kou, Yangyu Lv, Yifei Li, Yijing Liu, Yiming Wang, Yingya Zhang, Yitong
  Huang, Yong Li, You Wu, Yu Liu, Yulin Pan, Yun Zheng, Yuntao Hong, Yupeng
  Shi, Yutong Feng, Zeyinzi Jiang, Zhen Han, Zhi-Fan Wu, and Ziyu Liu.
\newblock {Wan: Open and Advanced Large-Scale Video Generative Models}.
\newblock \emph{arXiv preprint arXiv:2503.20314}, 2025.

\bibitem[{Video Rebirth}(2026)]{hyperflow}
{Video Rebirth}.
\newblock {HyperFlow} for {MiniMax-H3}.
\newblock Model repository, Hugging Face, 2026.

\bibitem[Vyas et~al.(2020)Vyas, Katharopoulos, and Fleuret]{clusteredattention}
Apoorv Vyas, Angelos Katharopoulos, and Fran{\c{c}}ois Fleuret.
\newblock Fast transformers with clustered attention.
\newblock In \emph{Advances in Neural Information Processing Systems}, 2020.

\bibitem[Wang et~al.(2004)Wang, Bovik, Sheikh, and Simoncelli]{ssim}
Zhou Wang, Alan~C. Bovik, Hamid~R. Sheikh, and Eero~P. Simoncelli.
\newblock Image quality assessment: From error visibility to structural
  similarity.
\newblock \emph{IEEE Transactions on Image Processing}, 13\penalty0
  (4):\penalty0 600--612, 2004.

\bibitem[Xi et~al.(2025)Xi, Yang, Zhao, Xu, Li, Li, Lin, Cai, Zhang, Li, Chen,
  Stoica, Keutzer, and Han]{svg}
Haocheng Xi, Shuo Yang, Yilong Zhao, Chenfeng Xu, Muyang Li, Xiuyu Li, Yujun
  Lin, Han Cai, Jintao Zhang, Dacheng Li, Jianfei Chen, Ion Stoica, Kurt
  Keutzer, and Song Han.
\newblock {Sparse VideoGen: Accelerating Video Diffusion Transformers with
  Spatial-Temporal Sparsity}.
\newblock \emph{arXiv preprint arXiv:2502.01776}, 2025.

\bibitem[Xu et~al.(2025)Xu, Xiao, Huang, Guo, and Han]{xattention}
Ruyi Xu, Guangxuan Xiao, Haofeng Huang, Junxian Guo, and Song Han.
\newblock {XAttention}: Block sparse attention with antidiagonal scoring.
\newblock \emph{arXiv preprint arXiv:2503.16428}, 2025.

\bibitem[Yang et~al.(2025{\natexlab{a}})Yang, Xi, Zhao, Li, Zhang, Cai, Lin,
  Li, Xu, Chen, Han, Keutzer, and Stoica]{svg2}
Shuo Yang, Haocheng Xi, Yilong Zhao, Muyang Li, Jintao Zhang, Han Cai, Yujun
  Lin, Xiuyu Li, Chenfeng Xu, Jianfei Chen, Song Han, Kurt Keutzer, and Ion
  Stoica.
\newblock {Sparse VideoGen2: Accelerate Video Generation with Sparse Attention
  via Semantic-Aware Permutation}.
\newblock \emph{arXiv preprint arXiv:2505.18875}, 2025{\natexlab{a}}.

\bibitem[Yang et~al.(2025{\natexlab{b}})Yang, Teng, Zheng, Ding, Huang, Xu,
  Yang, Hong, Zhang, Feng, Yin, Zhang, Wang, Cheng, Xu, Gu, Dong, and
  Tang]{cogvideox}
Zhuoyi Yang, Jiayan Teng, Wendi Zheng, Ming Ding, Shiyu Huang, Jiazheng Xu,
  Yuanming Yang, Wenyi Hong, Xiaohan Zhang, Guanyu Feng, Da Yin, Yuxuan Zhang,
  Weihan Wang, Yean Cheng, Bin Xu, Xiaotao Gu, Yuxiao Dong, and Jie Tang.
\newblock {CogVideoX: Text-to-Video Diffusion Models with An Expert
  Transformer}.
\newblock In \emph{International Conference on Learning Representations},
  2025{\natexlab{b}}.

\bibitem[Ye et~al.(2025)Ye, Chen, Lai, Lin, Zhang, Wang, Chen, Kasikci, Grover,
  Krishnamurthy, and Ceze]{flashinfer}
Zihao Ye, Lequn Chen, Ruihang Lai, Wuwei Lin, Yineng Zhang, Stephanie Wang,
  Tianqi Chen, Baris Kasikci, Vinod Grover, Arvind Krishnamurthy, and Luis
  Ceze.
\newblock {FlashInfer: Efficient and Customizable Attention Engine for LLM
  Inference Serving}.
\newblock In \emph{Conference on Machine Learning and Systems}, 2025.

\bibitem[Zadouri et~al.(2026)Zadouri, Hoehnerbach, Shah, Liu, Thakkar, and
  Dao]{flashattention4}
Ted Zadouri, Markus Hoehnerbach, Jay Shah, Timmy Liu, Vijay Thakkar, and Tri
  Dao.
\newblock {FlashAttention-4: Algorithm and Kernel Pipelining Co-Design for
  Asymmetric Hardware Scaling}.
\newblock \emph{arXiv preprint arXiv:2603.05451}, 2026.

\bibitem[Zhang et~al.(2025{\natexlab{a}})Zhang, Huang, Zhang, Wei, Zhu, and
  Chen]{sageattention2}
Jintao Zhang, Haofeng Huang, Pengle Zhang, Jia Wei, Jun Zhu, and Jianfei Chen.
\newblock {SageAttention2}: Efficient attention with thorough outlier smoothing
  and per-thread {INT4} quantization.
\newblock In \emph{Proceedings of the 42nd International Conference on Machine
  Learning}, pages 75097--75119, 2025{\natexlab{a}}.

\bibitem[Zhang et~al.(2025{\natexlab{b}})Zhang, Wang, Jiang, Yang, Zheng, Xi,
  Wang, Zhu, Zhao, Stoica, Gonzalez, Zhu, and Chen]{sla}
Jintao Zhang, Haoxu Wang, Kai Jiang, Shuo Yang, Kaiwen Zheng, Haocheng Xi,
  Ziteng Wang, Hongzhou Zhu, Min Zhao, Ion Stoica, Joseph~E. Gonzalez, Jun Zhu,
  and Jianfei Chen.
\newblock {SLA: Beyond Sparsity in Diffusion Transformers via Fine-Tunable
  Sparse-Linear Attention}.
\newblock \emph{arXiv preprint arXiv:2509.24006}, 2025{\natexlab{b}}.

\bibitem[Zhang et~al.(2025{\natexlab{c}})Zhang, Wei, Huang, Zhang, Zhu, and
  Chen]{sageattention}
Jintao Zhang, Jia Wei, Haofeng Huang, Pengle Zhang, Jun Zhu, and Jianfei Chen.
\newblock {SageAttention: Accurate 8-Bit Attention for Plug-and-play Inference
  Acceleration}.
\newblock In \emph{International Conference on Learning Representations},
  2025{\natexlab{c}}.

\bibitem[Zhang et~al.(2025{\natexlab{d}})Zhang, Xiang, Huang, Wei, Xi, Zhu, and
  Chen]{spargeattn}
Jintao Zhang, Chendong Xiang, Haofeng Huang, Jia Wei, Haocheng Xi, Jun Zhu, and
  Jianfei Chen.
\newblock {SpargeAttention: Accurate and Training-free Sparse Attention
  Accelerating Any Model Inference}.
\newblock In \emph{International Conference on Machine Learning},
  2025{\natexlab{d}}.

\bibitem[Zhang et~al.(2025{\natexlab{e}})Zhang, Chen, Huang, Lin, Liu, Stoica,
  Xing, and Zhang]{vsa}
Peiyuan Zhang, Yongqi Chen, Haofeng Huang, Will Lin, Zhengzhong Liu, Ion
  Stoica, Eric Xing, and Hao Zhang.
\newblock {VSA: Faster Video Diffusion with Trainable Sparse Attention}.
\newblock In \emph{Advances in Neural Information Processing Systems},
  2025{\natexlab{e}}.

\bibitem[Zhang et~al.(2025{\natexlab{f}})Zhang, Chen, Su, Ding, Stoica, Liu,
  and Zhang]{sta}
Peiyuan Zhang, Yongqi Chen, Runlong Su, Hangliang Ding, Ion Stoica, Zhengzhong
  Liu, and Hao Zhang.
\newblock {Fast Video Generation with Sliding Tile Attention}.
\newblock In \emph{International Conference on Machine Learning},
  2025{\natexlab{f}}.

\bibitem[Zhang et~al.(2026)Zhang, Wei, Zhao, Zhang, Zhou, Lin, Zhang, Nie,
  Zeng, and Zhang]{vsa2}
Peiyuan Zhang, Guoqiang Wei, Yilong Zhao, Zixiang Zhang, Wei Zhou, Will Lin,
  Heng Zhang, Xiaonan Nie, Yan Zeng, and Hao Zhang.
\newblock {Improving Video Sparse Attention with Fine-grained Router and Sparse
  Rebasing}.
\newblock \emph{arXiv preprint arXiv:2609.32882}, 2026.

\bibitem[Zhang et~al.(2018)Zhang, Isola, Efros, Shechtman, and Wang]{lpips}
Richard Zhang, Phillip Isola, Alexei~A. Efros, Eli Shechtman, and Oliver Wang.
\newblock {The Unreasonable Effectiveness of Deep Features as a Perceptual
  Metric}.
\newblock In \emph{IEEE/CVF Conference on Computer Vision and Pattern
  Recognition}, 2018.

\bibitem[Zhou et~al.(2026)Zhou, Mang, Yang, Xi, Zhang, Mao, Gonzalez, Keutzer,
  Stoica, and Cheung]{svgear}
Xuanyi Zhou, Qiuyang Mang, Shuo Yang, Haocheng Xi, Jintao Zhang, Huanzhi Mao,
  Joseph~E. Gonzalez, Kurt Keutzer, Ion Stoica, and Alvin Cheung.
\newblock {SVG-EAR: Parameter-Free Linear Compensation for Sparse Video
  Generation via Error-aware Routing}.
\newblock \emph{arXiv preprint arXiv:2603.08982}, 2026.

\end{thebibliography}
